\documentclass[sigconf, nonacm]{acmart}

\AtBeginDocument{
  }

\usepackage{multirow}
\usepackage{subcaption}

\begin{document}

\title{Internally Referenced Low-Light Enhancement}

\author{Peiyuan He, Hainuo Wang, Hengxing Liu, Mingjia Li, Xiaojie Guo\textsuperscript{*}}
\thanks{\textsuperscript{*}Corresponding author.} 

\affiliation{%
  \institution{College of Intelligence and Computing, Tianjin University, Tianjin 300350, China \newline
  \text{peiyuan\_he@tju.edu.cn \quad hainuo@tju.edu.cn \quad chrisliu.jz@gmail.com} \newline
  \text{mingjiali@tju.edu.cn \quad xj.max.guo@gmail.com}}
  \country{} 
}

\renewcommand{\shortauthors}{He et al.}

\begin{abstract}
Self-supervised low-light image enhancement (LLIE) is highly appealing as it eliminates the reliance on external paired data. 
However, the lack of external references causes networks to struggle with decoupling entangled illumination, delicate textures, and amplified noise. To resolve this challenge, we propose an Internally Referenced LLIE framework that extracts reliable physical and structural references from the degraded input image itself. 
First, we introduce a local exposure-simulated scheme to extract a low-frequency pseudo ground-truth. This serves as an internal physical reference to guide global illumination estimation and correct color casts.
Second, we propose a dual-domain preservation strategy with spatial and spectral constraints to construct internal structural references. Specifically, an Illumination-Aligned Perceptual loss preserves global structures under illumination shifts, while a Shift-Invariant Spectral Correlation loss captures fine-grained local structures and suppresses high-frequency noise.
Finally, we propose a Gain-Adaptive Feature Modulation (GAFM) mechanism to address highly spatially-variant residual noise. By transforming the self-estimated illumination map into an internal spatial gain prior, GAFM dynamically guides a blind-spot network for spatially-aware denoising.
Extensive experiments demonstrate that our method achieves state-of-the-art performance, delivering superior noise suppression and textural fidelity. Code will be publicly released at https://visonj.github.io/IRLE/.
\end{abstract}


\begin{teaserfigure}
  \centering
  \includegraphics[width=\textwidth]{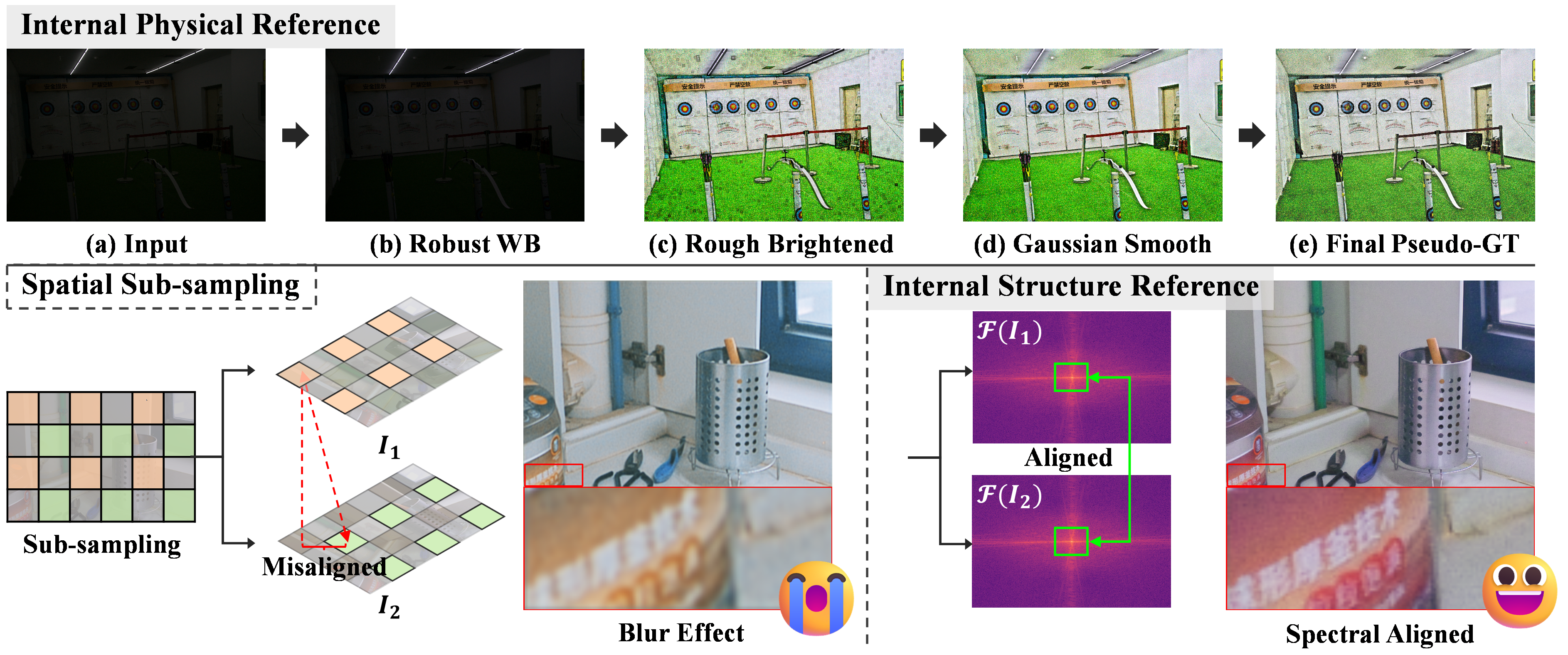}
  \vspace{-12pt}
  \caption{\textit{Top: Internal Physical Reference.} A local exposure-simulated pseudo-GT provides an internal physical reference for global color and brightness restoration. \textit{Bottom: Internal Structural References.} Our spectral-domain design provides internal structural references that preserve structures and suppress noise without the blur artifacts caused by spatial misalignment.}
  \Description{A two-part teaser figure. The top row illustrates the generation of an internal physical reference from a dark input image, ending in a pseudo-GT with improved global color and brightness. The bottom row contrasts spectral and spatial handling of sub-sampled images: aligned amplitude spectra in the spectral domain support structure preservation, whereas direct spatial reconstruction between misaligned sub-images leads to blur artifacts.}
  \label{fig:teaser}
\end{teaserfigure}

\maketitle

\section{Introduction}
Low-Light Image Enhancement (LLIE) aims to restore visual details and correct color casts in images captured under low-light conditions. Although supervised deep models have achieved impressive performance~\cite{zhang2019kindling, xu2022snr, cai2023retinexformer, zhou2024glare}, they rely heavily on large-scale paired references, which are difficult to acquire in real-world dynamic scenes due to camera shake, object motion, and lighting variations. To reduce this dependency, some methods learn from unpaired data via generative models such as GANs~\cite{jiang2021enlightengan}. However, they over-rely on the external target distribution, making them prone to hallucinations and distribution bias. As a result, zero-reference learning, which relies solely on low-light inputs without external normal-light targets, has emerged as a promising alternative for LLIE.

\begin{figure}[t]
  \centering
  \includegraphics[width=\linewidth]{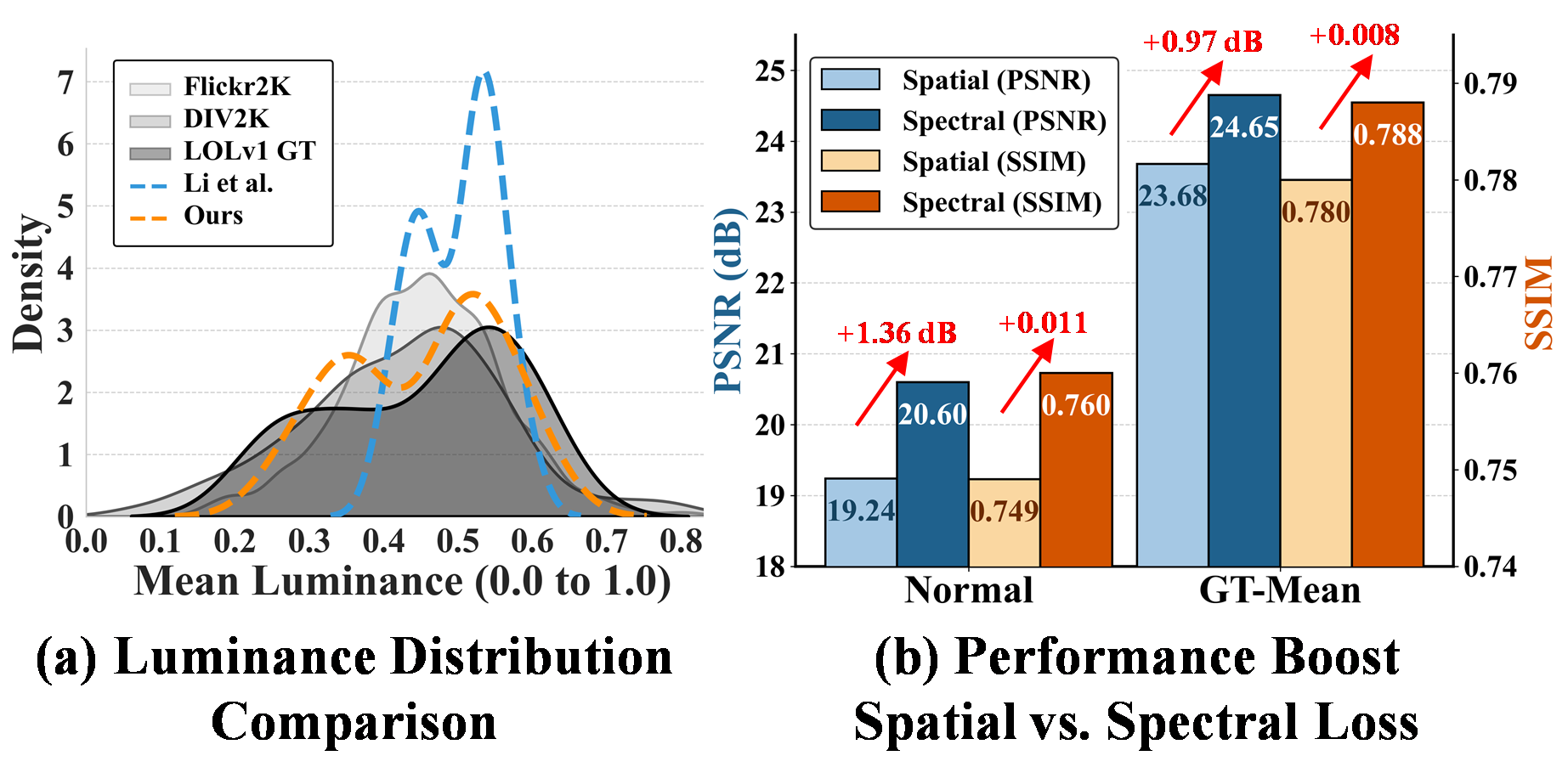}
  
  \Description{A figure with two side-by-side plots. The left is a density line graph comparing luminance distributions. The green line (Li \textit{et al.}) shows sharp, unnatural spikes. The red line (Ours) forms a smooth bell-like curve that closely overlaps with the shaded grey and black distributions of natural high-quality datasets (Flickr2K and DIV2K). The right is a bar chart comparing Spatial and Spectral losses. The bars for Spectral loss are consistently higher across both PSNR and SSIM metrics. Red arrows highlight the performance boosts, such as +1.36 dB for Normal PSNR and +0.97 dB for GT-Mean PSNR.}
  \vspace{-8pt}
  \caption{(a) Global luminance distribution comparison. Our method produces a luminance distribution closer to that of high-quality natural images. (b) Performance comparison between classical spatial loss and our spectral SISC loss.}
  \vspace{-12pt}
  \label{fig:illumination_analysis}
\end{figure}

Despite its theoretical appeal, existing zero-reference LLIE faces a critical issue: without external references, networks struggle to decouple entangled illumination, delicate textures, and amplified noise. Classical self-supervised methods (\emph{e.g.}, Zero-DCE~\cite{guo2020zero}, SCI~\cite{ma2022toward}) primarily treat LLIE as a pure illumination adjustment task. However, according to the Poisson-Gaussian physical model~\cite{foi2008practical}, photon noise is inherently coupled to the signal. When algorithms non-linearly boost dark regions, the hidden sensory noise is inevitably amplified.  As pointed out by prior studies~\cite{shi2024zero, yi2023diff}, this process alters noise characteristics and further entangles noise with the image content, making it difficult for the network to distinguish high-frequency pristine textures from random noise. Consequently, many methods rely on spatial sub-sampling (\emph{e.g.}, Neighbor2Neighbor~\cite{huang2021neighbor2neighbor}) to construct consistency constraints. However, enforcing consistency across spatially shifted pixels often introduces blurring effects due to inherent misalignment, as shown at the bottom-left of Fig.~\ref{fig:teaser}. Moreover, because the amplified noise is highly spatially-variant, standard self-supervised denoising networks (\emph{e.g.}, Blind-Spot Networks~\cite{lee2022ap}) that assume spatially invariant noise are inadequate for balancing heavy noise amplification in dark regions against the preservation of cleaner textures in bright areas.
In addition, previous unsupervised LLIE methods often exhibit a tendency to converge toward a narrow, averaged luminance range~\cite{han2026towards}. As shown in Fig.~\ref{fig:illumination_analysis}(a), this collapse significantly deviates from the natural distributions of well-illuminated and detail-preserved images found in high-quality datasets such as Flickr2K~\cite{lim2017enhanced} and DIV2K~\cite{Agustsson2017}.

Our philosophy is that, \textit{although a degraded input image lacks an external normal-light counterpart, it still contains sufficient physical and structural cues to guide the LLIE process.} 
This internally referenced perspective helps resolve the decoupling ambiguity among illumination, textures, and noise, while alleviating the tendency of previous methods to collapse toward an unnatural luminance range. 
Based on this insight, we propose an Internally Referenced Low-Light Image Enhancement (IRLE) framework, which extracts reliable physical and structural references from the input image, followed by a spatially-aware modulated denoising process.

We firstly propose a local exposure-simulated scheme featuring quantile-based robust white balancing and adaptive shadow desaturation to extract a pseudo ground-truth (pseudo-GT) from the input image, as shown at the top of Fig.~\ref{fig:teaser}. Serving as an \textit{internal physical reference}, this low-frequency reference mitigates color degradation and hue shifts in pitch-black areas, providing a reliable internal benchmark to guide global illumination estimation. 
Second, we further propose a dual-domain collaborative preservation strategy to construct \textit{internal structural references}. 
In the spatial domain, we introduce an Illumination-Aligned Perceptual (IAP) loss to preserve global structures against drastic illumination shifts.
Since comparing the enhanced output with the dark input in pixel space is unstable under severe brightness changes, IAP provides a semantic structural constraint that stabilizes the global structure during enhancement.
In the spectral domain, we introduce a Shift-Invariant Spectral Correlation (SISC) loss to extract fine-grained local structures without absorbing random noise. 
Previous self-supervised methods commonly employ spatial sub-sampling to obtain observations with independent noise realizations from a single image~\cite{huang2021neighbor2neighbor}. 
However, conventional spatial cross-reconstruction on these sub-sampled images inevitably introduces blurring artifacts due to inherent pixel misalignment, as shown at the bottom-left of Fig.~\ref{fig:teaser}. In contrast, by operating in the spectral domain, we exploit the fact that \textit{the amplitude spectra remain aligned despite spatial shifts~\cite{lv2024fourier, li2302embedding, wang2023fourllie, ryou2024robust}}, as shown in the bottom-right of Fig.~\ref{fig:teaser}. Based on this property, SISC aligns the amplitude spectrum to ensure blur-free structural consistency. This encourages the network to preserve fine-grained local structures while rejecting random noise. This advantage is also reflected quantitatively in Fig.~\ref{fig:illumination_analysis}(b), where replacing spatial consistency with our spectral design consistently improves performance under normal and GT-Mean evaluation settings.
Finally, we propose a Gain-Adaptive Feature Modulation (GAFM) mechanism for gain-guided blind-spot denoising. Since the residual noise after enhancement is highly correlated with the local illumination amplification, GAFM converts the self-estimated illumination map into an internal spatial gain prior. This prior enables a blind-spot network (\emph{e.g.}, PUCA~\cite{jang2023puca}) to perform spatially-aware denoising, applying stronger smoothing to severely dark regions while preserving delicate textures in well-illuminated areas.

Our main contributions can be summarized as follows:
\begin{itemize}
    \item We propose an Internally Referenced LLIE framework. By extracting physical and structural references directly from the degraded input, it alleviates the entanglement of illumination, textures, and noise in zero-reference learning.
    \item We propose a local exposure-simulated scheme to generate a pseudo-GT as an internal physical reference. In addition, we design a dual-domain collaborative strategy with the SISC and IAP losses to construct internal structural references for structure preservation and texture extraction.
    \item We propose a Gain-Adaptive Feature Modulation mechanism for blind-spot denoising. By converting the estimated illumination map into an internal spatial gain prior, it enables effective spatially-variant denoising without compromising delicate textures in well-illuminated areas.
\end{itemize}
Extensive experiments on multiple datasets demonstrate that our method achieves state-of-the-art performance for self-supervised LLIE, offering superior noise suppression and detail preservation.

\section{Related Work}
\label{sec:related}

\subsection{Low-Light Image Enhancement}
Deep learning has shifted low-light image enhancement (LLIE) from heuristic hand-crafted priors~\cite{guo2016lime} to data-driven mapping. Early supervised methods, such as LLNet~\cite{lore2017llnet}, MBLLEN~\cite{lv2018mbllen}, RetinexNet~\cite{wei2018deep}, and KinD~\cite{zhang2019kindling}, pioneered this transition. Subsequently, advanced architectures including URetinex-Net~\cite{wu2022uretinex}, normalizing flows such as LLFlow~\cite{wang2022low}, and transformer-based models including Restormer~\cite{zamir2022restormer} and Retinexformer~\cite{cai2023retinexformer} achieved remarkable success. However, their generalization ability is fundamentally constrained by the limited diversity of strictly paired training data, which still relies on hard-to-collect external references. To alleviate this dependency, zero-reference and self-supervised paradigms have evolved rapidly in recent years. Zero-DCE~\cite{guo2020zero} and Zero-DCE++~\cite{li2021learning} formulated LLIE as an image-specific curve estimation problem. Other studies explored semantic guidance (\emph{e.g.}, SGZ~\cite{zheng2022semantic}) and implicit neural representations (\emph{e.g.}, NeRCo~\cite{Yang_2023_ICCV}) to regularize the highly ill-posed enhancement process. SCI~\cite{ma2022toward} further improved computational efficiency by introducing a lightweight illumination learning framework. Later, methods such as PairLIE~\cite{fu2023learning} extracted self-consistent Retinex representations from unpaired data.

Despite these advances, most zero-reference methods still treat LLIE primarily as an illumination or brightness adjustment problem, often overlooking the fact that, under the Poisson-Gaussian model, photon noise is inherently coupled with the signal~\cite{xu2022snr}. Thus, aggressively boosting dark regions inevitably amplifies hidden sensory noise, entangling illumination, textures, and noise more severely. Different from these methods, our IRLE constructs a pseudo-GT from the degraded input as an internal physical reference. By using a local exposure-simulated scheme with quantile-based robust white balancing and adaptive shadow desaturation, we provide reliable low-frequency guidance for illumination estimation and color correction, thereby reducing the ambiguity between brightness enhancement and noise amplification.

\subsection{Self-Supervised Image Denoising}
Image denoising without clean ground truth has been extensively explored since Noise2Noise~\cite{lehtinen2018noise2noise}. To eliminate the need for multiple noisy captures, single-image self-supervised methods such as Noise2Void~\cite{krull2019noise2void}, Noise2Self~\cite{batson2019noise2self}, and Neighbor2Neighbor~\cite{huang2021neighbor2neighbor} were proposed. A prominent line of research within this domain is Blind-Spot Networks (BSNs)~\cite{laine2019high, wang2022blind2unblind}, which constrain the receptive field to exclude the central pixel. Advanced BSN architectures, such as AP-BSN~\cite{lee2022ap} and PUCA~\cite{jang2023puca}, employ operations like patch-unshuffle downsampling to break spatial noise correlation induced by image signal processors while enlarging the receptive field under the highly challenging zero-reference evaluation setting.

However, standard self-supervised denoisers generally assume a spatially invariant noise distribution. When applied after low-light enhancement, they face a mismatch: the amplified noise is highly spatially variant, being much stronger in severely dark regions and much weaker in already bright regions. Consequently, a BSN with uniform denoising strength must trade off between under-denoising dark regions and over-smoothing fine textures in bright regions. In contrast, our method introduces a gain-guided blind-spot network equipped with a Gain-Adaptive Feature Modulation (GAFM) mechanism. By translating the estimated illumination map into an internal spatial gain prior, the network becomes spatially aware and can adapt its denoising behavior according to the image's own illumination-dependent degradation.

\begin{figure*}[t]
    \centering
    \includegraphics[width=0.98\textwidth]{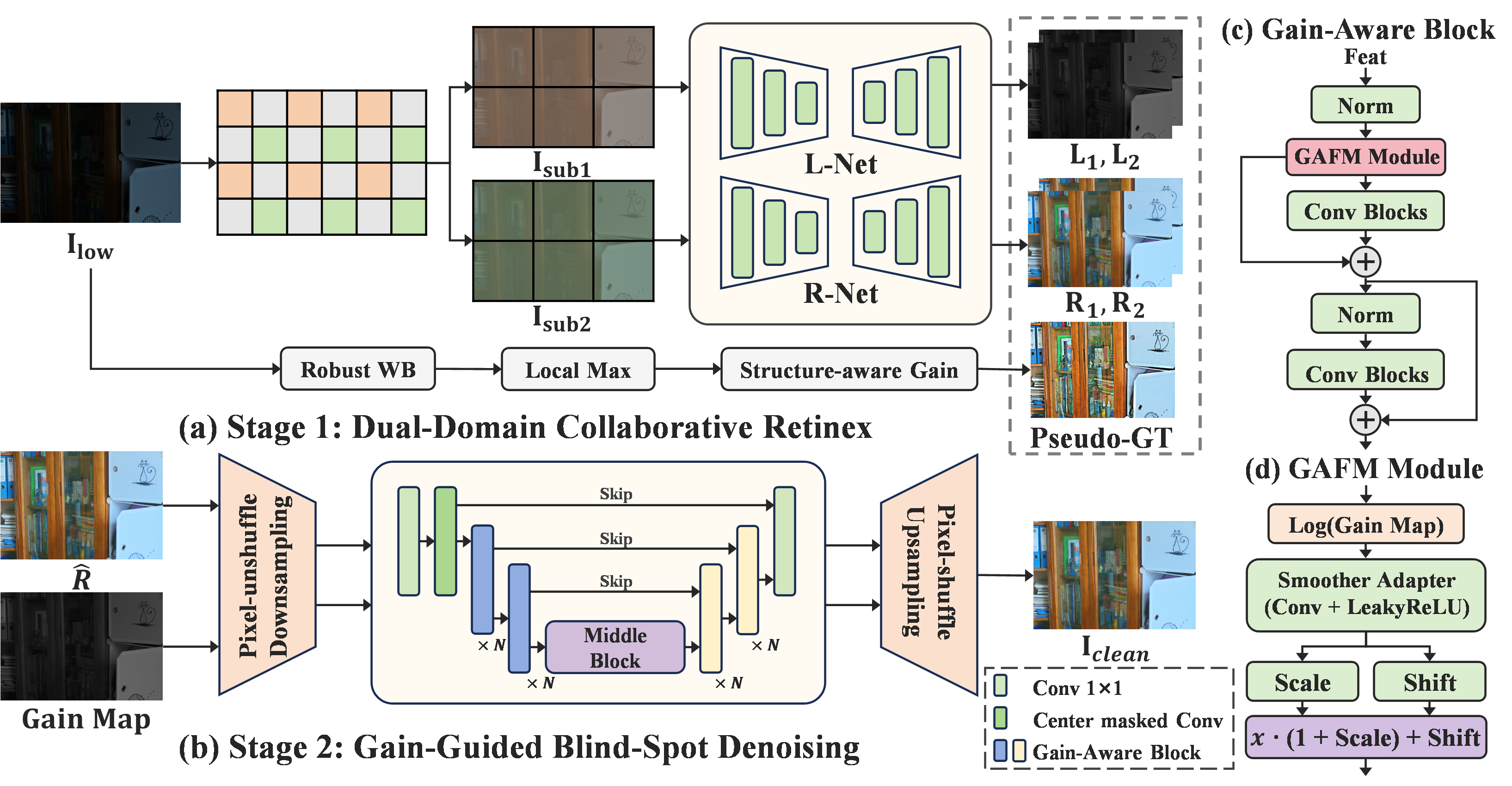}
    
    \Description{A comprehensive block diagram detailing the architecture of the IRLE framework, divided into four panels. Panel (a) shows Stage 1: a low-light input image splits into two paths. The upper path performs spatial sub-sampling into two sub-images, which are fed into a Dual-Domain Retinex module (comprising L-Net and R-Net) to output paired illumination and reflectance maps. The lower path processes the input through Robust White Balance, Local Max, and Structure-aware Gain blocks to generate a Pseudo Ground Truth. Panel (b) shows Stage 2: the estimated reflectance and a Gain Map undergo pixel-unshuffle downsampling and enter a U-Net-like Gain-Guided Blind-Spot Network (BSN) with skip connections and a middle block, followed by pixel-shuffle upsampling to output the final clean image. Panel (c) is a vertical flowchart of the Gain-Aware Block, consisting of normalization, a GAFM module, and convolutional blocks connected via residual additions. Panel (d) illustrates the GAFM Module's internal data flow, where the logarithm of the Gain Map passes through a Smoother Adapter to generate Scale and Shift parameters, which then linearly modulate the input features.}
    \vspace{-10pt}
    \caption{\textbf{Overview of our IRLE.} \textbf{(a) \textit{Stage 1}:} Illumination estimation and structure extraction via a Dual-Domain Collaborative Retinex network, guided by a local exposure-simulated pseudo-GT. \textbf{(b) \textit{Stage 2}:} Gain-guided blind-spot denoising, which handles spatially-variant noise. \textbf{(c)} The detailed architecture of the Gain-Aware Block. \textbf{(d)} The Gain-Adaptive Feature Modulation (GAFM) module, which translates the inverse illumination map into an internal spatial gain prior for dynamic denoising.}
    \vspace{-6pt}
    \label{fig:pipeline}
\end{figure*}

\subsection{Frequency Analysis in Image Restoration}
The frequency domain naturally separates low-frequency structural components from high-frequency noise patterns~\cite{huang2022deep, li2023embedding}. Recent image restoration methods have exploited Fourier-based amplitude and phase manipulation~\cite{yang2020fda, chen2021amplitude} to decouple style-related variations from structural semantics, while Focal Frequency Loss~\cite{jiang2021focal} directly emphasizes hard-to-synthesize frequency bands during optimization. Nevertheless, exploiting frequency-domain priors for zero-reference LLIE remains relatively under-explored, especially in the presence of spatial sub-sampling and severe noise amplification. First, traditional self-supervised sub-sampling methods such as Neighbor2Neighbor~\cite{huang2021neighbor2neighbor} enforce spatial consistency between shifted samples, which often introduces blurring artifacts due to inherent pixel misalignment. Second, directly matching frequency statistics of noisy observations in a zero-reference setting can cause the network to inherit undesirable high-frequency sensory noise rather than the underlying authentic physical structures.

To address these issues, our framework constructs global and local internal structural references through a dual-domain collaborative preservation strategy. In the spectral domain, we propose the Shift-Invariant Spectral Correlation (SISC) loss, which leverages the translation invariance of the Fourier transform to align amplitude spectra between spatially shifted sub-images without inducing spatial blur. In addition, SISC masks out noise-dominated high-frequency bands in the cross-frequency correlation matrix, encouraging the network to preserve shared pristine textures while rejecting unshared random noise. In the spatial domain, the IAP loss references global structures against drastic illumination shifts. Together, these constraints provide unified internal structural guidance for structure preservation and texture extraction.

\section{Methodology}
\label{sec:method}

\subsection{Overall Framework}
\label{subsec:overview}
Given a single low-light and noisy image $\mathbf{I}_{\text{low}} \in \mathbb{R}^{H \times W \times 3}$, our primary objective is to recover a normally-exposed, noise-suppressed, and detail-preserved image $\mathbf{I}_{\text{clean}}$ without relying on any paired normal-light ground truth. To effectively address the highly ill-posed nature of zero-reference LLIE, we formulate our framework based on a physically grounded noise-corrupted Retinex model:
\begin{equation}
\mathbf{I}_{\text{low}} = \mathbf{R}_{\text{clean}} \odot \mathbf{L} + \mathbf{N},
\end{equation}
where $\mathbf{R}_{\text{clean}}$ denotes the desired clean reflectance (i.e., $\mathbf{I}_{\text{clean}}$), $\mathbf{L}$ represents the illumination map, $\mathbf{N}$ is the inherent sensory noise, and $\odot$ denotes element-wise multiplication.

During illumination boosting, the recovered reflectance becomes approximately $\mathbf{R}_{\text{clean}} + \mathbf{N}/\mathbf{L}$. As the illumination $\mathbf{L}$ approaches zero in dark regions, the noise term $\mathbf{N}/\mathbf{L}$ is significantly amplified, resulting in highly spatially-variant noise. 
This makes zero-reference LLIE particularly challenging, since relying solely on spatial consistency or global frequency matching without external targets leads to optimization ambiguity. 
To address these issues, we propose an internally referenced LLIE framework, illustrated in Fig.~\ref{fig:pipeline}, which addresses the degradation from the perspectives of global illumination, structural preservation and spatially-variant denoising.

In Stage 1, we decouple illumination and structure estimation from noise fitting. We first generate a local exposure-simulated pseudo-GT to guide global color and brightness restoration. To maintain structural integrity, we apply a Dual-Domain Collaborative Preservation strategy: an Illumination-Aligned Perceptual (IAP) loss in the spatial domain preserves global topology against illumination shifts, while a Shift-Invariant Spectral Correlation (SISC) loss in the spectral domain aligns sub-sampled amplitude spectra and masked frequency correlations to extract fine-grained textures without blurring. This stage outputs the estimated illumination map $\hat{\mathbf{L}}$ alongside a well-illuminated, structurally preserved, yet residually noisy reflectance map $\hat{\mathbf{R}}$.

In Stage 2, we address the highly spatially-variant residual noise ($\mathbf{N}/\mathbf{L}$) in $\hat{\mathbf{R}}$. By converting the estimated illumination $\hat{\mathbf{L}}$ into an internal spatial gain prior, we inject it into a hierarchical blind-spot network via a Gain-Adaptive Feature Modulation (GAFM) mechanism. This dynamic guidance explicitly directs the network to apply heavy smoothing to dark, high-gain regions while preserving delicate textures in well-illuminated areas.

\subsection{Internal Physical Reference}
\label{subsec:stage1}

We derive a local exposure-simulated pseudo-GT to serve as an internal physical reference for global color and brightness restoration.
\label{subsubsec:pseudo_gt}
First, for robust white balancing, standard Gray-World assumptions often struggle in low-light conditions as heavy read noise can skew global estimations. Instead, we extract a high-confidence mask $\mathbf{M}_{\text{wb}}$ by isolating the top $q$-th quantile luminance $\mathcal{Q}_q(\cdot)$ of the input $\mathbf{I}_{\text{low}}$:
\begin{equation}
    \mathbf{M}_{\text{wb}} = \mathbb{I}(\mathbf{Y}_{\text{low}} \ge \mathcal{Q}_q(\mathbf{Y}_{\text{low}})),
\end{equation}
where $\mathbf{Y}_{\text{low}}$ denotes the extracted grayscale luminance of the input $\mathbf{I}_{\text{low}}$, and $\mathbb{I}(\cdot)$ is the standard indicator function. By computing the channel-wise mean intensities exclusively within the reliable region $\mathbf{M}_{\text{wb}}$, we obtain the robust white-balance gains $\mathbf{w} \in \mathbb{R}^3$. This produces the color-corrected image $\mathbf{I}_{\text{wb}} = \mathbf{I}_{\text{low}} \odot \mathbf{w}$.

Second, to derive a structure-aware physical gain that adapts to spatially non-uniform illumination, we first compute a patch-wise local maximum $\mathbf{I}_{\text{max}}$ to yield a rough amplification map $\mathbf{G}_{\text{rough}} = 1 / (\mathbf{I}_{\text{max}} + \epsilon)$, as visualized in Fig.~\ref{fig:teaser}(c). Unlike naive pixel-wise Max-RGB boosting, which inherently destroys local contrast and amplifies independent noise, our patch-wise formulation dynamically adapts to local darkness while preserving essential spatial coherence. Subsequently, to ensure natural illumination transitions and establish a smooth target gain, we apply a large-kernel Gaussian low-pass filter $\mathcal{G}_{\sigma}(\cdot)$ directly to the rough gain map, yielding $\mathbf{G}_{\text{target}} = \mathcal{G}_{\sigma}(\mathbf{G}_{\text{rough}})$. As shown in Fig.~\ref{fig:teaser}(d), this filtering operation effectively softens sharp gain boundaries, mitigating unnatural dark halos surrounding bright light tubes and the undesirable blocky patterns on the grass. Ultimately, this yields the safely brightened final output image $\mathbf{I}_{\text{bright}} = \mathbf{I}_{\text{wb}} \odot \mathbf{G}_{\text{target}}$.

Third, boosting pitch-black regions frequently amplifies hidden chroma noise into noticeable color blotches. To mitigate this undesirable color degradation issue, we introduce an adaptive shadow desaturation mechanism controlled by a smooth luminance-dependent spatial weight map $\mathbf{W}_{\text{c}} \in [0, 1]^{H \times W}$:
\begin{equation}
    \mathbf{W}_{\text{c}} = \mathrm{clip}\left( \frac{\mathbf{Y}_{\text{bright}} - \theta_{\text{min}}}{\theta_{\text{max}} - \theta_{\text{min}}}, 0, 1 \right),
\end{equation}
where $[\theta_{\text{min}}, \theta_{\text{max}}]$ defines the shadow luminance transition band, and $\mathbf{Y}_{\text{bright}}$ is the grayscale luminance of $\mathbf{I}_{\text{bright}}$. The final pseudo-GT smoothly blends the color image with its grayscale counterpart:
\begin{equation}
    \mathbf{I}_{\text{pseudo}} = \mathbf{Y}_{\text{bright}} + \mathbf{W}_{\text{c}} \odot (\mathbf{I}_{\text{bright}} - \mathbf{Y}_{\text{bright}}).
\end{equation}
As shown in Fig.~\ref{fig:teaser}(e), by decaying $\mathbf{W}_{\text{c}}$ to zero in extreme shadows, this mechanism smoothly forces irrecoverable noisy regions to clean grayscale while preserving vivid colors in well-lit areas. Since $\mathbf{I}_{\text{pseudo}}$ unavoidably retains amplified high-frequency sensory noise, we utilize it primarily as a reliable low-frequency internal physical reference. The task of high-frequency noise rejection is then delegated to our subsequent dual-domain constraints.

\begin{table*}[t]
  \centering
  \caption{Quantitative comparison on LOL~\cite{wei2018deep,yang2021sparse} datasets. Both standard and GT-Mean metrics are reported. The best and second-best results among unsupervised methods are highlighted in \textbf{bold} and \underline{underline}, respectively.}
  \label{tab:quantitative_main}
  \setlength{\tabcolsep}{6pt}
  \begin{tabular}{@{}l cc cc cc cc cc cc@{}}
    \toprule
    \multicolumn{1}{c}{\multirow{3}{*}{Method}} & \multicolumn{4}{c}{LOLv1} & \multicolumn{4}{c}{LOLv2-Real} & \multicolumn{4}{c}{LOLv2-Synthetic} \\
    \cmidrule(lr){2-5} \cmidrule(lr){6-9} \cmidrule(lr){10-13}
    & \multicolumn{2}{c}{Normal} & \multicolumn{2}{c}{GT-Mean} & \multicolumn{2}{c}{Normal} & \multicolumn{2}{c}{GT-Mean} & \multicolumn{2}{c}{Normal} & \multicolumn{2}{c}{GT-Mean} \\
    \cmidrule(lr){2-3} \cmidrule(lr){4-5} \cmidrule(lr){6-7} \cmidrule(lr){8-9} \cmidrule(lr){10-11} \cmidrule(lr){12-13}
    & PSNR & SSIM & PSNR & SSIM & PSNR & SSIM & PSNR & SSIM & PSNR & SSIM & PSNR & SSIM \\
    \midrule
    Zero-DCE~\cite{guo2020zero} (CVPR'20) & 14.86 & 0.559 & 21.06 & 0.535 & 18.06 & 0.574 & 20.78 & 0.542 & 17.76 & \underline{0.813} & 21.50 & \textbf{0.849} \\
    RRDNet~\cite{zhu2020zero} (ICME'20) & 11.46 & 0.460 & 18.96 & 0.484 & 13.96 & 0.483 & 19.15 & 0.495 & 14.87 & 0.657 & 18.39 & 0.758 \\
    RUAS~\cite{liu2021retinex} (CVPR'21) & 16.41 & 0.500 & 18.65 & 0.518 & 15.33 & 0.488 & 19.06 & 0.510 & 13.40 & 0.644 & 17.79 & 0.695 \\
    EnlightenGAN~\cite{jiang2021enlightengan} (TIP'21) & 17.56 & 0.665 & 21.33 & 0.649 & 18.68 & 0.673 & 21.04 & 0.663 & 16.49 & 0.775 & 19.32 & 0.823 \\
    RetinexDIP~\cite{zhao2021retinexdip} (TCSVT'21) & 11.67 & 0.484 & 19.74 & 0.471 & 14.51 & 0.521 & 19.48 & 0.487 & 16.01 & 0.733 & 20.12 & 0.800 \\
    SCI~\cite{ma2022toward} (CVPR'22) & 14.78 & 0.522 & 18.97 & 0.501 & 17.30 & 0.534 & 19.47 & 0.509 & 15.43 & 0.748 & 18.64 & 0.788 \\
    PSENet~\cite{nguyen2023psenet} (WACV'23) & 17.50 & 0.543 & 20.93 & 0.546 & 17.63 & 0.531 & 20.64 & 0.550 & 16.62 & 0.777 & 20.67 & 0.824 \\
    PairLIE~\cite{fu2023learning} (CVPR'23) & 19.51 & 0.736 & 23.17 & 0.753 & 19.89 & 0.778 & 24.03 & 0.803 & \underline{19.07} & 0.797 & \underline{21.68} & 0.820 \\
    GDP~\cite{fei2023generative} (CVPR'23) & 15.82 & 0.541 & 19.09 & 0.578 & 14.40 & 0.494 & 19.32 & 0.559 & 12.12 & 0.497 & 15.83 & 0.667 \\
    NeRCo~\cite{Yang_2023_ICCV} (ICCV'23) & 19.74 & 0.743 & 22.41 & 0.755 & 19.66 & 0.717 & 23.63 & 0.750 & 17.59 & 0.734 & 19.66 & 0.752 \\
    CLIP-LIT~\cite{liang2023iterative} (ICCV'23) & 12.39 & 0.493 & 20.03 & 0.442 & 15.18 & 0.529 & 19.45 & 0.468 & 16.19 & 0.775 & 20.75 & 0.817 \\
    CoLIE~\cite{Chobola2024} (ECCV'24) & 13.76 & 0.481 & 20.37 & 0.479 & 15.08 & 0.501 & 20.22 & 0.496 & 14.30 & 0.654 & 19.04 & 0.786 \\
    CLODE~\cite{jung2025continuous} (ICLR'25) & 19.60 & 0.718 & 22.59 & 0.736 & 17.87 & 0.681 & 22.57 & 0.703 & 17.21 & 0.783 & 20.63 & 0.797 \\
    Li \textit{et al.}~\cite{li2025interpretable} (ICLR'25) & \underline{19.82} & \underline{0.751} & \underline{23.97} & \underline{0.779} & \underline{20.35} & \textbf{0.795} & \underline{26.14} & \textbf{0.828} & 17.82 & 0.802 & 20.78 & 0.820 \\
    \midrule
    Ours & \textbf{20.60} & \textbf{0.760} & \textbf{24.65} & \textbf{0.788} & \textbf{20.72} & \underline{0.792} & \textbf{26.24} & \underline{0.826} & \textbf{19.62} & \textbf{0.815} & \textbf{22.82} & \underline{0.833} \\
  \bottomrule
  \end{tabular}
\end{table*}

\begin{figure}[t]
  \centering
  \includegraphics[width=\linewidth]{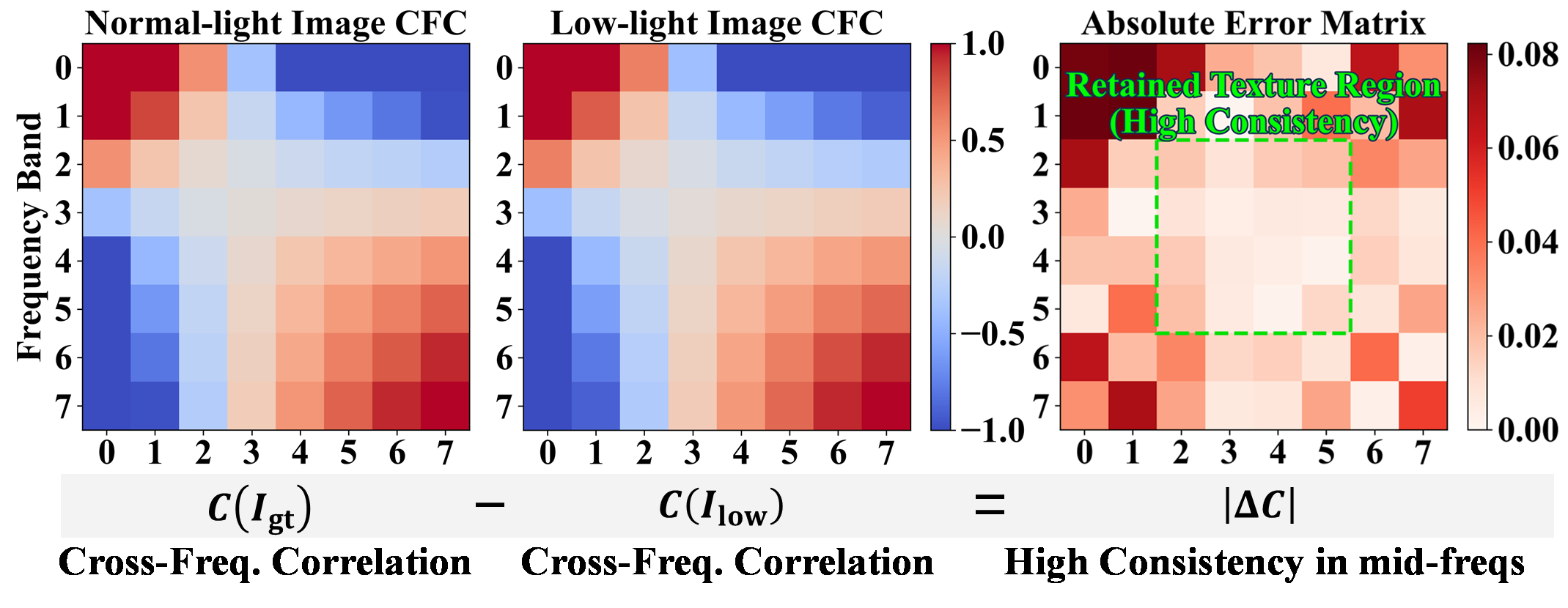}
  
  \Description{Three heatmaps arranged horizontally showing cross-frequency correlation (CFC). The left and middle heatmaps represent the CFC of normal-light and low-light images, respectively. The right heatmap shows their absolute error matrix. The outer edges (representing extreme low and high frequencies) exhibit higher errors in dark red. A green dashed box highlights the center region (bands 2 to 5) with very light colors, labeled 'Retained Texture Region (High Consistency)', indicating minimal error.}
  \vspace{-10pt}
  \caption{\textbf{Cross-Frequency Correlation (CFC) analysis.} $C(\cdot)$ denotes the CFC matrix calculated across $K$ frequency bands.}
  \label{fig:cfc_matrix}
  \vspace{-6pt}
\end{figure}

\subsection{Internal Structural Reference}
\label{subsubsec:dual_domain}

To extract the internal structural reference without absorbing the amplified signal-dependent noise, we propose a Dual-Domain Collaborative Preservation strategy. Our motivation is to construct internal structural references by separating shared physical textures from independent random noise. While spatial-domain constraints are intuitive, imposing consistency on spatially shifted samples risks introducing blur to compensate for pixel misalignment. To bypass this dilemma, we shift our perspective to the spectral domain to exploit its shift-invariant properties, while using a spatial-domain perceptual constraint to preserve global structures.

In the spectral domain, we introduce the Shift-Invariant Spectral Correlation (SISC) loss to preserve fine-grained local structures and suppress noise without compromising sharpness. This loss operates on network predictions $\hat{\mathbf{R}}_1$ and $\hat{\mathbf{R}}_2$, derived from spatially sub-sampled images. While sensory noise in these sub-images is uncorrelated, their underlying physical structures are identical. However, directly penalizing the absolute differences between their frequency spectra is problematic: it forces the network to fit specific high-frequency noise patterns. To separate deterministic textures from random noise, we shift our focus from absolute frequency magnitudes to the relative dependencies between different frequency bands. As supported by the visual evidence in Fig.~\ref{fig:cfc_matrix}, natural textures exhibit strong structural coupling across multiple frequency bands, whereas random noise lacks this organized cross-band co-occurrence. Specifically, the absolute error matrix reveals distinct degradation behaviors: extreme low-frequency bands (e.g., bands 0-1) exhibit significant deviations due to severe illumination degradation, while extreme high-frequency bands (e.g., bands 6-7) are heavily corrupted by amplified sensory noise. In contrast, the mid-frequency region (highlighted by the green dashed box covering bands 2 to 5) demonstrates robust stability, maintaining high structural consistency regardless of the lighting conditions.

To explicitly model these underlying shared structural relationships and effectively filter out the degraded extreme bands, we first compute the normalized cross-frequency correlation (CFC) matrix $\mathbf{C}(\hat{\mathbf{R}}) \in \mathbb{R}^{K \times K}$ using the band-wise log-energy vector $\mathbf{E}(\hat{\mathbf{R}}) \in \mathbb{R}^K$ across $K$ disjoint radial frequency bands:
\begin{equation}
    \mathbf{C}(\hat{\mathbf{R}}) = \frac{\big(\mathbf{E}(\hat{\mathbf{R}}) - \mu_E\big)\big(\mathbf{E}(\hat{\mathbf{R}}) - \mu_E\big)^\top}{\sigma_E \sigma_E^\top + \epsilon},
\end{equation}
where $\mu_E$ and $\sigma_E$ are the scalar mean and standard deviation of $\mathbf{E}(\hat{\mathbf{R}})$. Based on the critical observation from Fig.~\ref{fig:cfc_matrix}, we design a binary symmetric mask $\mathbf{M}_{\text{freq}} \in \{0, 1\}^{K \times K}$ to zero out interactions involving the highly corrupted frequency extremes. This strategy ensures that the network enforces correlation alignment exclusively within the reliable mid-frequency textural regions.

Additionally, directly aligning spatially shifted sub-images in the pixel domain introduces blurring artifacts due to inherent spatial misalignment. We resolve this by leveraging the translation invariance of the Fourier transform, where spatial shifts alter the phase but leave the underlying amplitude spectrum invariant. By combining the masked frequency correlation with an amplitude consistency penalty, the overall SISC loss is defined as:
\begin{equation}
    \mathcal{L}_{\text{sisc}} = \big\| \mathbf{M}_{\text{freq}} \odot \big( \mathbf{C}(\hat{\mathbf{R}}_1) - \mathbf{C}(\hat{\mathbf{R}}_2) \big) \big\|_F + \gamma \big\| |\mathcal{F}(\hat{\mathbf{R}}_1)| - |\mathcal{F}(\hat{\mathbf{R}}_2)| \big\|_1,
\end{equation}
where $\mathcal{F}(\cdot)$ denotes the 2D Fast Fourier Transform, $|\cdot|$ extracts the amplitude spectrum, and $\gamma$ balances the amplitude penalty. By evaluating both correlation and amplitude consistency in the spectral domain, SISC encourages the network to converge on shared physical structures and discard the independent random noise, thereby avoiding the spatial blurring penalties typical of conventional spatial sub-sampling methods.

In the spatial domain, while the spectral constraints successfully preserve fine-grained local structures, they lack the spatial awareness necessary to preserve the global semantic topology. To preserve these global structures, we propose an Illumination-Aligned Perceptual (IAP) loss ($\mathcal{L}_{\text{iap}}$). Computing perceptual differences directly between the brightened reflectance $\hat{\mathbf{R}}$ and the dark input $\mathbf{I}_{\text{low}}$ is unstable due to the severe brightness gap. Thus, we introduce a spatial-agnostic scaling factor $k$ to align their global intensities. By matching the VGG features $\Phi_j(\cdot)$ between the brightness-aligned prediction $k\hat{\mathbf{R}}$ and $\mathbf{I}_{\text{low}}$, the network preserves the structural topology against significant illumination shifts:
\begin{equation}
    \mathcal{L}_{\text{iap}} = \sum_{j} \big\| \Phi_j(k\hat{\mathbf{R}}) - \Phi_j(\mathbf{I}_{\text{low}}) \big\|_1, \quad k = \frac{\langle \hat{\mathbf{R}}, \mathbf{I}_{\text{low}} \rangle}{\|\hat{\mathbf{R}}\|_2^2 + \epsilon}.
\end{equation}
The detailed derivation of $k$ is provided in the supplementary material.
Following these dual-domain structural references, we apply a self-reconstruction loss $\mathcal{L}_{\text{rec}} = \|\hat{\mathbf{R}} \odot \mathbf{L} - \mathbf{I}_{\text{low}}\|_1$ to enforce the Retinex physics, and an illumination guidance loss $\mathcal{L}_{\text{guide}} = \|\mathbf{L} - \mathbf{G}_{\text{target}}^{-1}\|_1$ constrained by the inverse of the pseudo-GT gain $\mathbf{G}_{\text{target}}$. 

Additionally, to prevent severe color degradation during brightness stretching, we formulate an aggregated color loss $\mathcal{L}_{\text{color}}$ guided by the internal physical reference $\mathbf{I}_{\text{pseudo}}$:
\begin{equation}
\begin{aligned}
    \mathcal{L}_{\text{color}} = &\lambda_{\text{loc}} \mathbb{E}\left[ 1 - \cos\big(\mathcal{P}(\hat{\mathbf{R}}), \mathcal{P}(\mathbf{I}_{\text{pseudo}})\big) \right] \\
    &+ \lambda_{\text{glo}} \sum_{c \in \{\text{r}, \text{g}, \text{b}\}} \big\| \mu_c(\hat{\mathbf{R}}) - \mu_{\text{gray}}(\hat{\mathbf{R}}) \big\|_1,
\end{aligned}
\end{equation}
where $\mathcal{P}(\cdot)$ denotes patch-wise average pooling, $\mu_c$ is the global spatial mean of channel $c$, and $\mu_{\text{gray}}$ is the average intensity across all channels. The first term effectively prevents local hue shifts by aligning with the estimated pseudo-GT, while the second mathematically enforces the global Gray-World color balance. Thus, the comprehensive spatial-domain objective is defined as $\mathcal{L}_{\text{spatial}} = \lambda_{\text{rec}} \mathcal{L}_{\text{rec}} + \lambda_{\text{guide}} \mathcal{L}_{\text{guide}} + \mathcal{L}_{\text{color}} + \lambda_{\text{iap}} \mathcal{L}_{\text{iap}}$.

Finally, combining the dual-domain constraints, the overall training objective for Stage 1 is optimized as:
\begin{equation}
    \mathcal{L}_{\text{Stage1}} = \mathcal{L}_{\text{spatial}} + \mathcal{L}_{\text{sisc}}.
\end{equation}

\begin{figure*}[t]
    \centering
    \begin{subfigure}[t]{0.19\textwidth}
        \centering
        \includegraphics[width=\textwidth]{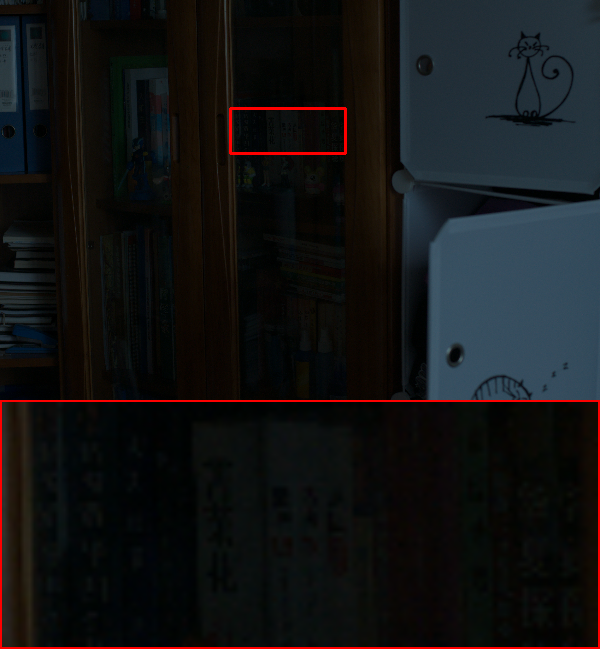}
        \caption{Input}
        \label{fig:comp_input}
    \end{subfigure}
    \hfill
    \begin{subfigure}[t]{0.19\textwidth}
        \centering
        \includegraphics[width=\textwidth]{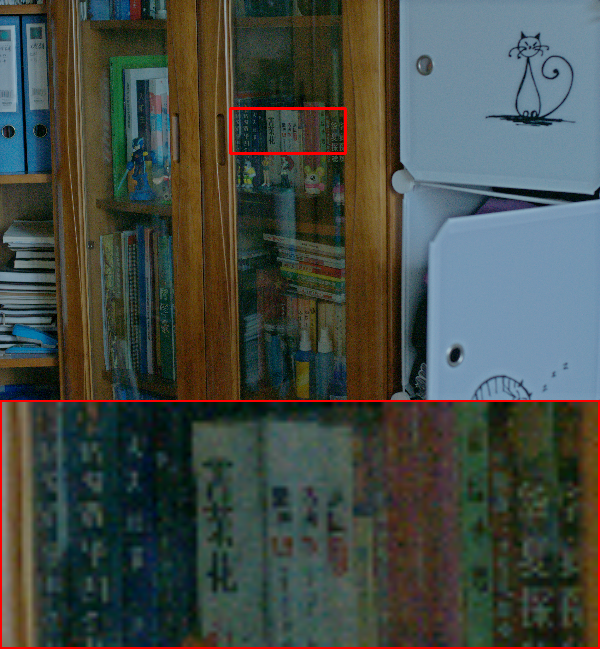}
        \caption{Zero-DCE~\cite{guo2020zero}}
        \label{fig:comp_zerodce}
    \end{subfigure}
    \hfill
    \begin{subfigure}[t]{0.19\textwidth}
        \centering
        \includegraphics[width=\textwidth]{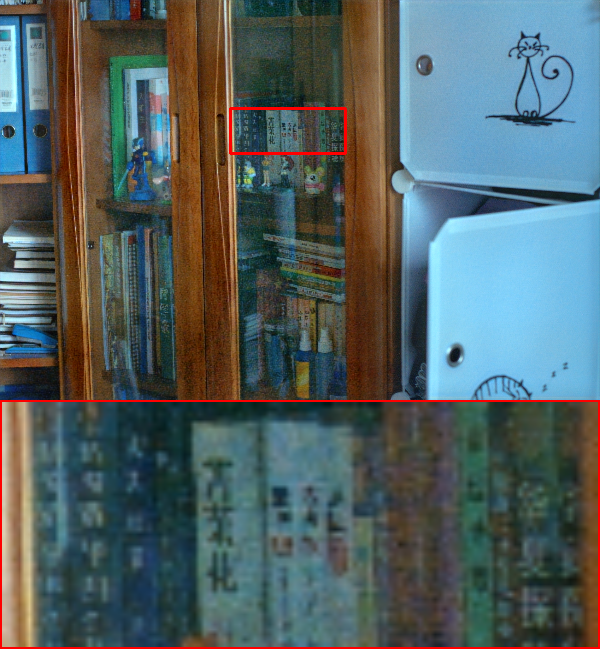}
        \caption{EnlightenGAN~\cite{jiang2021enlightengan}}
        \label{fig:comp_engan}
    \end{subfigure}
    \hfill
    \begin{subfigure}[t]{0.19\textwidth}
        \centering
        \includegraphics[width=\textwidth]{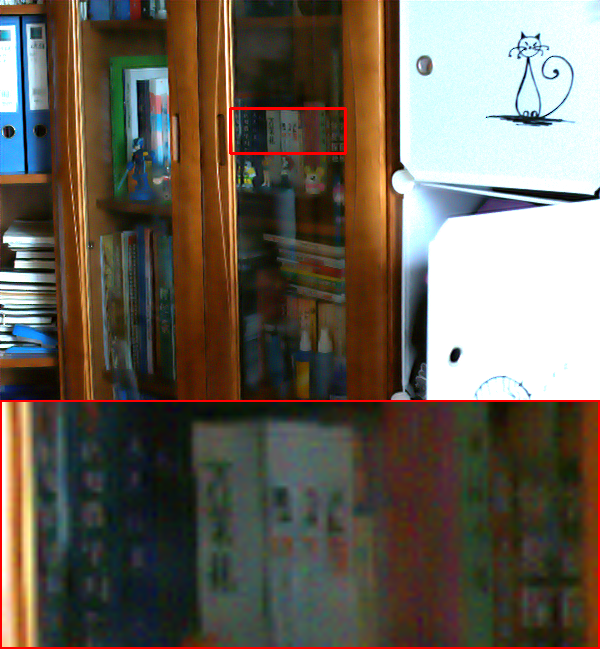}
        \caption{RUAS~\cite{liu2021retinex}}
        \label{fig:comp_ruas}
    \end{subfigure}
    \hfill
    \begin{subfigure}[t]{0.19\textwidth}
        \centering
        \includegraphics[width=\textwidth]{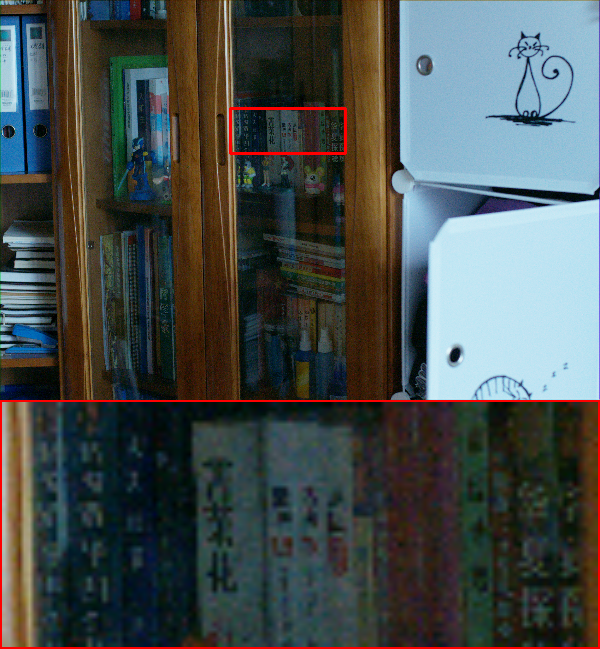}
        \caption{SCI~\cite{ma2022toward}}
        \label{fig:comp_sci}
    \end{subfigure}
    
    \medskip 
    
    \begin{subfigure}[t]{0.19\textwidth}
        \centering
        \includegraphics[width=\textwidth]{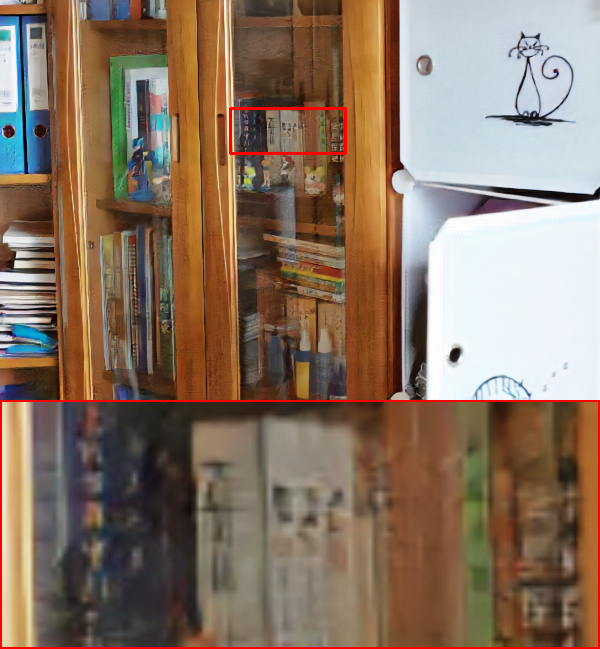}
        \caption{NeRCo~\cite{Yang_2023_ICCV}}
        \label{fig:comp_nerco}
    \end{subfigure}
    \hfill
    \begin{subfigure}[t]{0.19\textwidth}
        \centering
        \includegraphics[width=\textwidth]{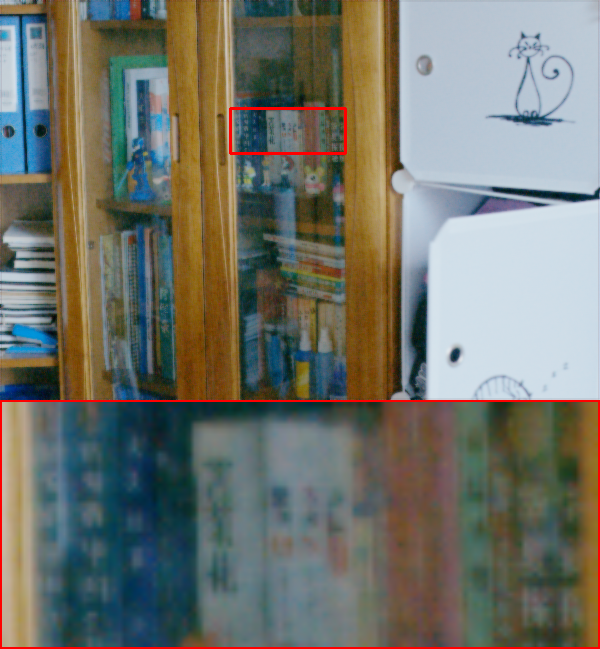}
        \caption{CLODE~\cite{jung2025continuous}}
        \label{fig:comp_clode}
    \end{subfigure}
    \hfill
    \begin{subfigure}[t]{0.19\textwidth}
        \centering
        \includegraphics[width=\textwidth]{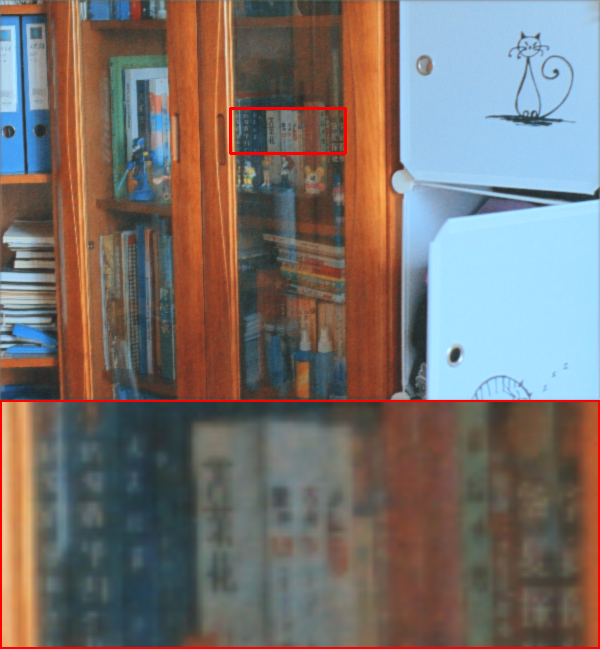}
        \caption{Li \textit{et al.}~\cite{li2025interpretable}}
        \label{fig:comp_li}
    \end{subfigure}
    \hfill
    \begin{subfigure}[t]{0.19\textwidth}
        \centering
        \includegraphics[width=\textwidth]{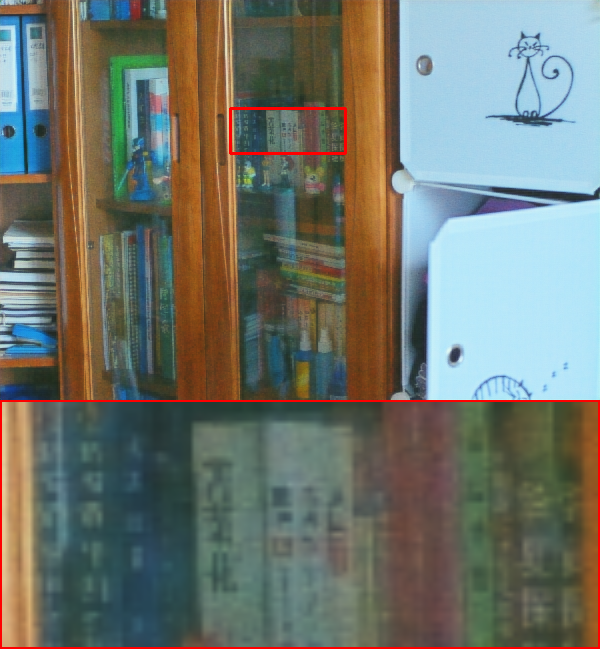}
        \caption{\textbf{Ours}}
        \label{fig:comp_ours}
    \end{subfigure}
    \hfill
    \begin{subfigure}[t]{0.19\textwidth}
        \centering
        \includegraphics[width=\textwidth]{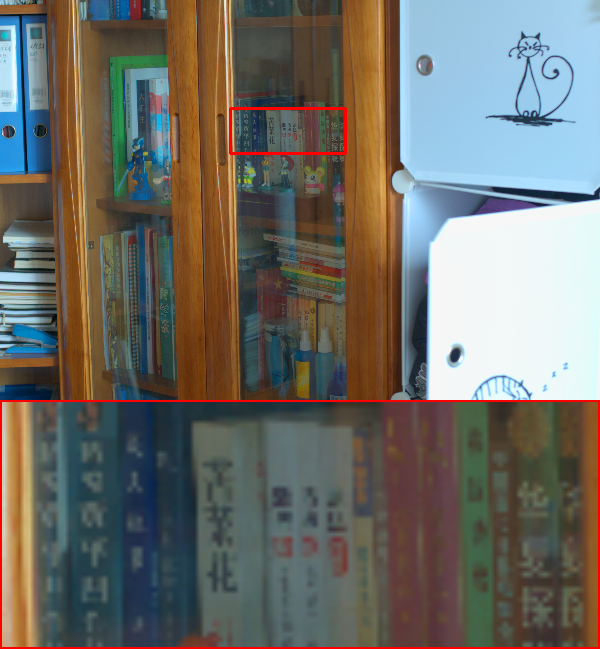}
        \caption{GT}
        \label{fig:comp_gt}
    \end{subfigure}

    \Description{A visual comparison consisting of ten subfigures arranged in two rows of five. The images show a dark indoor scene of a bookshelf. Each subfigure includes a red rectangular crop magnified below it to show the detailed texture of book spines. The 'Input' is extremely dark. Competitor methods (Zero-DCE, EnlightenGAN, RUAS, SCI, NeRCo, CLODE, Li \textit{et al.}) exhibit visible noise, color shifts, or blurred text in the magnified crops. In contrast, 'Ours' displays clear text and accurate colors, closely matching the Ground Truth (GT) image.}
    \vspace{-8pt}
    \caption{\textbf{Visual comparison on LOLv1~\cite{wei2018deep}.} Existing methods often exhibit color casts or lose delicate details due to over-smoothing. But our method maintains natural colors and effectively removes spatially-variant noise while preserving structures.}
    \vspace{-6pt}
    \label{fig:lolv1_comparison}
\end{figure*}

\subsection{Gain-Guided Blind-Spot Denoising}
\label{subsec:stage2}

Although Stage 1 successfully extracts a structurally preserved reflectance $\hat{\mathbf{R}}$, it inevitably retains the amplified signal-dependent noise. Following the Poisson-Gaussian model, brightening dark regions significantly amplifies this noise, while initially well-lit areas remain relatively clean. Consequently, the residual noise distribution observed in $\hat{\mathbf{R}}$ is highly spatially-variant. 

Standard self-supervised denoising networks, such as Blind-Spot Networks, typically apply spatially uniform smoothing. When faced with such spatially-variant degradation, they are forced into a difficult trade-off between under-denoising severely corrupted dark regions and blurring pristine bright textures. To overcome this limitation, we propose the Gain-Adaptive Feature Modulation (GAFM) mechanism, adopting PUCA~\cite{jang2023puca} as our baseline network. GAFM aims to transform any standard, spatially-invariant denoising network into a dynamic, spatially-aware denoiser by explicitly injecting an extracted internal spatial gain prior.

Recognizing that the severity of the residual noise is directly proportional to the illumination amplification, GAFM first translates the estimated illumination map $\hat{\mathbf{L}}$ from Stage 1 into this internal spatial gain prior, defined as $\mathbf{G}_{\text{prior}} = 1 / (\hat{\mathbf{L}} + \epsilon)$. This prior indicates the required amplification level for each spatial location, serving as foundational guidance for the spatially-variant denoising process.

To seamlessly inject this prior into the network hierarchy, the GAFM module (Fig.~\ref{fig:pipeline}(d)) projects the gain into the logarithmic domain to compress its extreme dynamic range, yielding $\mathbf{G}_{\text{log}} = \log(\mathbf{G}_{\text{prior}})$. Crucially, instead of relying on a detached sub-network, $\mathbf{G}_{\text{log}}$ undergoes the exact same downsampling operations (e.g., patch-unshuffling) as the noisy input. This design ensures strict spatial alignment with the intermediate features $\mathbf{F} \in \mathbb{R}^{C \times H \times W}$ at every hierarchical level. To prevent inherent checkerboard artifacts caused by such downsampling, a $3 \times 3$ convolutional smoother adapter $\mathrm{A}(\cdot)$ maps the aligned $\mathbf{G}_{\text{log}}$ into the latent space without breaking structural coherence. GAFM then dynamically generates affine parameters $[\mathbf{\Gamma}, \mathbf{\Delta}] = \mathrm{Conv}_{1 \times 1}\Big( \mathrm{A}(\mathbf{G}_{\text{log}}) \Big)$ to modulate the features via a Spatial Feature Transform:
\begin{equation}
    \tilde{\mathbf{F}} = \mathbf{F} \odot (1 + \mathbf{\Gamma}) + \mathbf{\Delta}.
\end{equation}

Optimized solely by the self-supervised blind-spot reconstruction loss $\mathcal{L}_{\text{denoise}} = \|\hat{\mathbf{R}} - \hat{\mathbf{I}}_{\text{clean}}\|_1$, this aligned internal guidance empowers the network to dynamically adjust its denoising intensity. It allows the model to apply stronger smoothing to high-gain dark regions while safely preserving delicate textures in well-illuminated areas, culminating in the final high-fidelity enhanced image $\hat{\mathbf{I}}_{\text{clean}}$.

\begin{table}[tb]
  \caption{Quantitative comparison on the LSRW-Huawei~\cite{hai2023r2rnet} dataset. The best and second-best results among unsupervised methods are highlighted in \textbf{bold} and \underline{underline}.}
  \vspace{-10pt}
  \label{tab:quantitative_huawei}
  \centering
  \setlength{\tabcolsep}{5pt}
  \begin{tabular}{@{}l cc cc@{}}
    \toprule
    \multicolumn{1}{@{}c@{}}{\multirow{2}{*}{Method}} & \multicolumn{2}{c}{Normal} & \multicolumn{2}{c}{GT-Mean} \\
    \cmidrule(lr){2-3} \cmidrule(lr){4-5}
    & PSNR & SSIM & PSNR & SSIM \\
    \midrule
    Zero-DCE~\cite{guo2020zero} (CVPR'20) & 16.40 & 0.475 & 19.98 & 0.465 \\
    RRDNet~\cite{zhu2020zero} (ICME'20) & 13.38 & 0.394 & 18.48 & 0.422 \\
    RUAS~\cite{liu2021retinex} (CVPR'21) & 15.71 & 0.500 & 17.60 & 0.544 \\
    EnlightenGAN~\cite{jiang2021enlightengan} (TIP'21) & 18.95 & 0.529 & 20.64 & 0.524 \\
    RetinexDIP~\cite{zhao2021retinexdip} (TCSVT'21) & 12.97 & 0.403 & 19.07 & 0.435 \\
    SCI~\cite{ma2022toward} (CVPR'22) & 15.70 & 0.439 & 18.28 & 0.454 \\
    PSENet~\cite{nguyen2023psenet} (WACV'23) & 18.20 & 0.474 & 19.69 & 0.472 \\
    PairLIE~\cite{fu2023learning} (CVPR'23) & \underline{18.98} & 0.562 & \underline{20.74} & 0.564 \\
    GDP~\cite{fei2023generative} (CVPR'23) & 13.81 & 0.403 & 17.61 & 0.481 \\
    CLIP-LIT~\cite{liang2023iterative} (ICCV'23) & 13.56 & 0.424 & 19.15 & 0.398 \\
    CoLIE~\cite{Chobola2024} (ECCV'24) & 14.76 & 0.420 & 19.30 & 0.444 \\
    CLODE~\cite{jung2025continuous} (ICLR'25) & 18.44 & \underline{0.564} & 20.48 & \underline{0.579} \\
    \midrule
    Ours & \textbf{19.43} & \textbf{0.601} & \textbf{21.57} & \textbf{0.613} \\
  \bottomrule
  \end{tabular}
  \vspace{-6pt}
\end{table}

\section{Experiments}
\label{sec:experiments}

\subsection{Experimental Settings}
\label{subsec:exp_settings}
\textbf{Datasets.} We evaluate our proposed internally referenced framework on four widely adopted low-light image enhancement benchmarks: LOLv1~\cite{wei2018deep}, LOLv2-Real~\cite{yang2021sparse}, LOLv2-Synthetic~\cite{yang2021sparse}, and the Huawei subset of the LSRW dataset~\cite{hai2023r2rnet}. The LOLv1 dataset contains 485 training pairs and 15 testing pairs. LOLv2-Real consists of 689 low/normal-light pairs for training and 100 images for testing. LOLv2-Synthetic provides 900 synthetic training pairs and 100 testing pairs. The LSRW-Huawei dataset includes 2450 training pairs and 30 testing pairs. Note that while these datasets provide paired ground truths, our method strictly utilizes only the low-light images during training. The paired normal-light images are exclusively used for quantitative evaluation.

\textbf{Evaluation Metrics.} To quantitatively assess the enhancement performance, we employ three standard full-reference metrics: Peak Signal-to-Noise Ratio to measure pixel-level fidelity, Structural Similarity Index to evaluate the preservation of structural details. Following recent standards, we report both standard metrics (Normal) and mean-aligned metrics (GT-Mean) to account for global brightness discrepancies across the dataset.

\textbf{Implementation Details.} Our framework is implemented in PyTorch and trained on a single NVIDIA RTX 3090 GPU. The network parameters are optimized using the Adam optimizer with hyperparameters set to $\beta_1 = 0.9$, $\beta_2 = 0.999$, and $\epsilon = 1 \times 10^{-8}$. The learning rate is initialized to $1 \times 10^{-4}$ and gradually decayed using a Cosine Annealing schedule to a minimum of $1 \times 10^{-6}$.

\subsection{Comparison with State-of-the-Art Methods}
\label{subsec:comparison}

To evaluate the effectiveness of our framework, we compare it against representative state-of-the-art unsupervised LLIE methods spanning diverse paradigms. These include curve estimation (Zero-DCE~\cite{guo2020zero}, SCI~\cite{ma2022toward}), unpaired GANs (EnlightenGAN~\cite{jiang2021enlightengan}), Retinex-based learning (RUAS~\cite{liu2021retinex}), implicit representations (NeRCo~\cite{Yang_2023_ICCV}), and recent advancements (CLODE~\cite{jung2025continuous}, Li \textit{et al.}~\cite{li2025interpretable}). 

\begin{figure*}[t]
    \centering
    \begin{subfigure}[t]{0.19\textwidth}
        \centering
        \includegraphics[width=\textwidth]{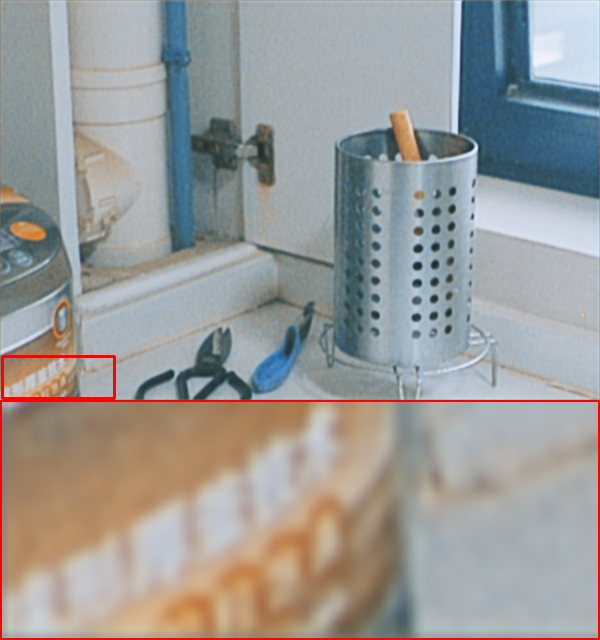}
        \caption{w/ Spatial L1}
        \label{fig:abl_exp1}
    \end{subfigure}
    \hfill
    \begin{subfigure}[t]{0.19\textwidth}
        \centering
        \includegraphics[width=\textwidth]{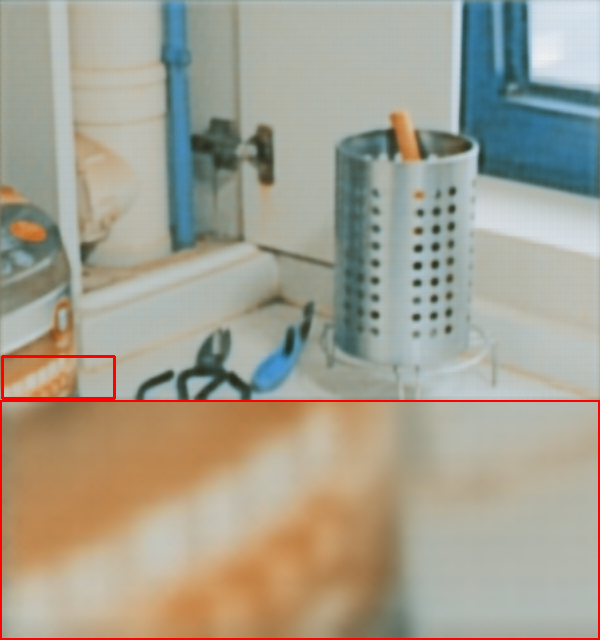}
        \caption{w/o Dual-D.}
        \label{fig:abl_exp2}
    \end{subfigure}
    \hfill
    \begin{subfigure}[t]{0.19\textwidth}
        \centering
        \includegraphics[width=\textwidth]{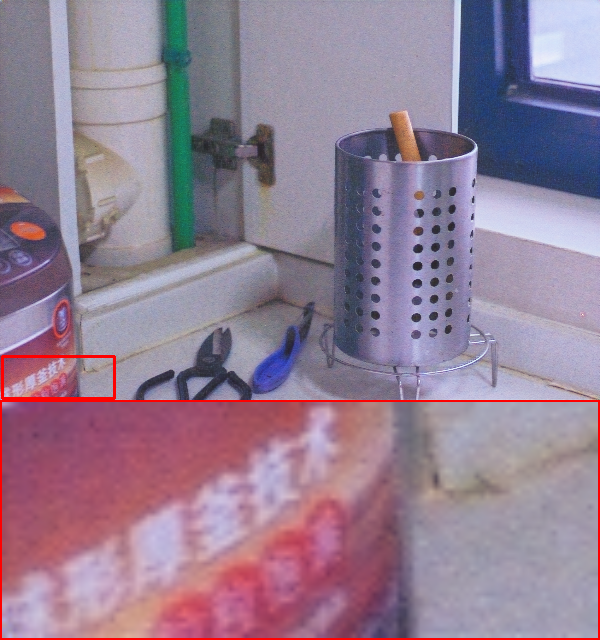}
        \caption{Max-RGB}
        \label{fig:abl_exp3}
    \end{subfigure}
    \hfill
    \begin{subfigure}[t]{0.19\textwidth}
        \centering
        \includegraphics[width=\textwidth]{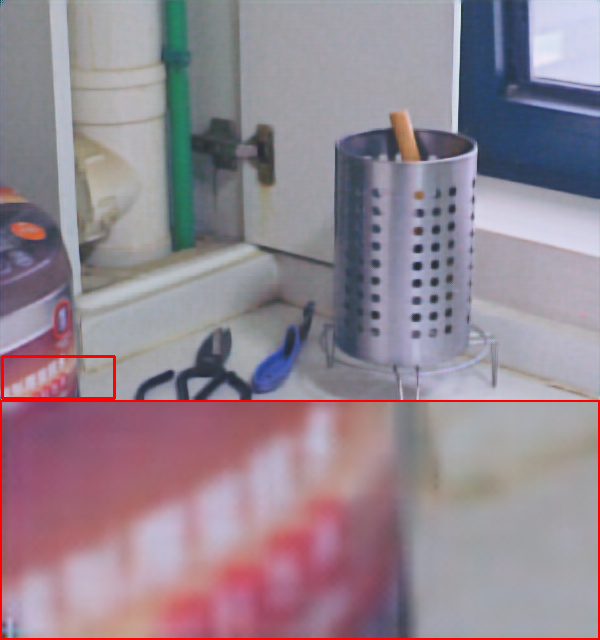}
        \caption{w/o GAFM}
        \label{fig:abl_exp4}
    \end{subfigure}
    \hfill
    \begin{subfigure}[t]{0.19\textwidth}
        \centering
        \includegraphics[width=\textwidth]{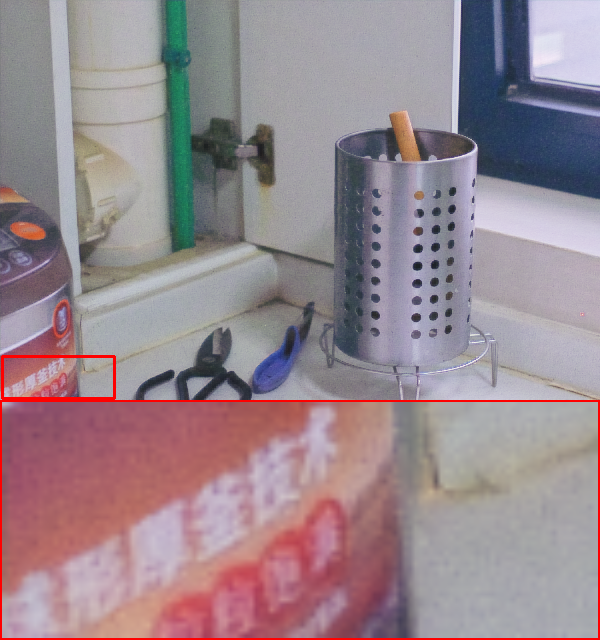}
        \caption{\textbf{Ours (Full)}}
        \label{fig:abl_ours}
    \end{subfigure}
    
    \Description{A visual ablation study comprising five subfigures in a single row. The images display a windowsill with a metal utensil holder, along with a magnified crop of text on a round container. The variants shown are: 'w/ Spatial L1' exhibiting blurry and ghosting text, 'w/o Dual-D.' showing structurally degraded text, 'Max-RGB' displaying a noticeable color cast, and 'w/o GAFM' presenting overly smoothed, smudged textures. The final subfigure, 'Ours (Full)', demonstrates sharp, clear text and natural colors, distinctly outperforming the ablated versions.}
    \caption{\textbf{Visual ablation study.} We compare IRLE (e) against 4 variants (a-d) lacking specific components. Our full model effectively suppresses spatially-variant noise while recovering sharp micro-textures without color casts or blurring artifacts.}
    \label{fig:ablation_visual}
\end{figure*}

\textbf{Quantitative Evaluation.} 
Tables \ref{tab:quantitative_main} and \ref{tab:quantitative_huawei} present the comprehensive quantitative comparison results. As shown, our method consistently achieves competitive PSNR performance across all four datasets. Specifically, on the LOLv1 and LOLv2-Real datasets, our method reaches 20.60 dB and 20.72 dB, respectively. On the LOLv2-Synthetic dataset, it achieves 19.62 dB. Furthermore, on the LSRW-Huawei dataset, our method achieves a PSNR of 19.43 dB, demonstrating favorable performance compared to other unsupervised methods.

\textbf{Qualitative Evaluation.}
To comprehensively evaluate the visual quality, we provide detailed visual comparisons of local details and global illumination on the LOLv1 dataset in Fig.~\ref{fig:lolv1_comparison}. In the comparison of local details, it can be observed that our enhanced results exhibit natural colors, clear texture details, and minimal residual noise. In contrast, CLODE and Li \textit{et al.} tend to exhibit color shifts, while NeRCo struggles to preserve the fine structural details of the text in the zoomed-in crops. Furthermore, in the visual comparison of global illumination and color, our method accurately restores global brightness and natural colors, achieving an overall visual quality consistent with the ground truth. Other methods, however, produce blurry results accompanied by inaccurate color tones.

\begin{table}[tb]
  \caption{Ablation study on LOLv1~\cite{wei2018deep}.}
  \vspace{-10pt}
  \label{tab:ablation}
  \centering
  \begin{tabular*}{\columnwidth}{@{\extracolsep{\fill}}l cccc@{}}
    \toprule
    \multirow{2}{*}{Variant} & \multicolumn{2}{c}{Normal} & \multicolumn{2}{c}{GT-Mean} \\
    \cmidrule(lr){2-3} \cmidrule(lr){4-5}
    & PSNR & SSIM & PSNR & SSIM \\
    \midrule
    (a) w/ Spatial L1 & 19.24 & 0.749 & 23.68 & 0.780 \\
    (b) w/o Dual-Domain & 17.59 & 0.665 & 21.17 & 0.693 \\
    (c) Basic Max-RGB & 19.93 & 0.704 & 24.06 & 0.728 \\
    (d) w/o GAFM & 20.07 & 0.738 & 24.11 & 0.765 \\
    \midrule
    \textbf{Ours (Full)} & \textbf{20.60} & \textbf{0.760} & \textbf{24.65} & \textbf{0.788} \\
  \bottomrule
  \end{tabular*}
  \vspace{-6pt}
\end{table}

\subsection{Ablation Study}
\label{subsec:ablation}

To validate the effectiveness of the core components in our framework, we conduct comprehensive ablation studies on the LOLv1 dataset. We design four variant models, and the visual impacts of each module are illustrated in Fig.~\ref{fig:ablation_visual}, with quantitative results summarized in Table~\ref{tab:ablation} for detailed comparison.

\textbf{Spatial Consistency vs. SISC Loss.} 
To demonstrate the limitations of spatial-domain sub-sampling, Variant (a) replaces our SISC loss with a traditional spatial L1 consistency loss. As shown in Fig.~\ref{fig:ablation_visual}(a), enforcing strict spatial alignment on physically shifted sub-images introduces noticeable blurring artifacts to the output.

\textbf{Necessity of Dual-Domain Collaborative Preservation.}
In Variant (b), we remove the proposed dual-domain constraints ($\mathcal{L}_{\text{iap}}$ and $\mathcal{L}_{\text{sisc}}$). Without these texture-preserving constraints, the PSNR drops severely to $17.59$ dB. Visually, as shown in Fig.~\ref{fig:ablation_visual}(b), the enhanced results become blurry and lose fine details, indicating that our dual-domain design is effective and essential for robust texture preservation during restoration.

\textbf{Effectiveness of High-Quality pseudo-GT.}
Variant (c) replaces our proposed local exposure-simulated pseudo-GT with a Max-RGB illumination prior. Without our pseudo-GT generation strategy, the network is more susceptible to amplified noise in extremely dark areas. Consequently, the visual results (Fig.~\ref{fig:ablation_visual}(c)) exhibit some residual noise and unnatural color shifts.

\textbf{Impact of GAFM.} 
Variant (d) removes the GAFM mechanism, depriving the blind-spot network of its spatial gain modulation and reducing it to a uniform BSN. Without the guidance of the internal spatial gain prior, the network applies uniform smoothing across the entire image. While this effectively removes noise, it leads to global over-smoothing (Fig.~\ref{fig:ablation_visual}(d)), causing a loss of pristine high-frequency details in bright regions.

Ultimately, as shown in the visual results (Fig.~\ref{fig:ablation_visual}(e)), by synergizing the shift-invariant SISC loss, dual-domain constraints, high-quality pseudo-GT, and spatially-aware GAFM, our internally referenced scheme successfully mitigates the aforementioned degradations, yielding final restored results with rich micro-textures, minimal residual noise, and accurate natural colors.

\section{Conclusion}
\label{sec:conclusion}

In this paper, we proposed an Internally Referenced Low-Light Image Enhancement (IRLE) framework to resolve the decoupling ambiguity among illumination, textures, and noise in zero-reference learning. Instead of seeking external targets, IRLE extracts reliable references directly from the degraded input. First, a local exposure-simulated pseudo-GT provides an internal physical reference for robust illumination and color correction. Second, a dual-domain collaborative strategy, combining Illumination-Aligned Perceptual and Shift-Invariant Spectral Correlation losses, constructs structural references to preserve fine textures without introducing spatial blurring. Finally, we introduced the Gain-Adaptive Feature Modulation (GAFM) mechanism, which translates the estimated illumination into an internal spatial gain prior for dynamic, spatially-aware denoising. Extensive experiments demonstrate that IRLE achieves state-of-the-art performance, delivering superior noise suppression and textural fidelity.

\section{Acknowledgments}
This work was partially supported by computational resources from TPU Research Cloud (TRC).

\bibliographystyle{ACM-Reference-Format}
\bibliography{main}


\begin{thebibliography}{54}


\ifx \showCODEN    \undefined \def \showCODEN     #1{\unskip}     \fi
\ifx \showISBNx    \undefined \def \showISBNx     #1{\unskip}     \fi
\ifx \showISBNxiii \undefined \def \showISBNxiii  #1{\unskip}     \fi
\ifx \showISSN     \undefined \def \showISSN      #1{\unskip}     \fi
\ifx \showLCCN     \undefined \def \showLCCN      #1{\unskip}     \fi
\ifx \shownote     \undefined \def \shownote      #1{#1}          \fi
\ifx \showarticletitle \undefined \def \showarticletitle #1{#1}   \fi
\ifx \showURL      \undefined \def \showURL       {\relax}        \fi
\providecommand\bibfield[2]{#2}
\providecommand\bibinfo[2]{#2}
\providecommand\natexlab[1]{#1}
\providecommand\showeprint[2][]{arXiv:#2}

\bibitem[Agustsson and Timofte(2017)]%
        {Agustsson2017}
\bibfield{author}{\bibinfo{person}{Eirikur Agustsson} {and} \bibinfo{person}{Radu Timofte}.} \bibinfo{year}{2017}\natexlab{}.
\newblock \showarticletitle{NTIRE 2017 Challenge on Single Image Super-Resolution: Dataset and Study}. In \bibinfo{booktitle}{\emph{IEEE Conference on Computer Vision and Pattern Recognition Workshops}}.
\newblock


\bibitem[Batson and Royer(2019)]%
        {batson2019noise2self}
\bibfield{author}{\bibinfo{person}{Joshua Batson} {and} \bibinfo{person}{Loic Royer}.} \bibinfo{year}{2019}\natexlab{}.
\newblock \showarticletitle{Noise2self: Blind denoising by self-supervision}. In \bibinfo{booktitle}{\emph{International conference on machine learning}}. PMLR, \bibinfo{pages}{524--533}.
\newblock


\bibitem[Cai et~al\mbox{.}(2023)]%
        {cai2023retinexformer}
\bibfield{author}{\bibinfo{person}{Yuanhao Cai}, \bibinfo{person}{Hao Bian}, \bibinfo{person}{Jing Lin}, \bibinfo{person}{Haoqian Wang}, \bibinfo{person}{Radu Timofte}, {and} \bibinfo{person}{Yulun Zhang}.} \bibinfo{year}{2023}\natexlab{}.
\newblock \showarticletitle{Retinexformer: One-stage retinex-based transformer for low-light image enhancement}. In \bibinfo{booktitle}{\emph{Proceedings of the IEEE/CVF international conference on computer vision}}. \bibinfo{pages}{12504--12513}.
\newblock


\bibitem[Chen et~al\mbox{.}(2021)]%
        {chen2021amplitude}
\bibfield{author}{\bibinfo{person}{Guangyao Chen}, \bibinfo{person}{Peixi Peng}, \bibinfo{person}{Li Ma}, \bibinfo{person}{Jia Li}, \bibinfo{person}{Lin Du}, {and} \bibinfo{person}{Yonghong Tian}.} \bibinfo{year}{2021}\natexlab{}.
\newblock \showarticletitle{Amplitude-phase recombination: Rethinking robustness of convolutional neural networks in frequency domain}. In \bibinfo{booktitle}{\emph{Proceedings of the IEEE/CVF international conference on computer vision}}. \bibinfo{pages}{458--467}.
\newblock


\bibitem[Chobola et~al\mbox{.}(2024)]%
        {Chobola2024}
\bibfield{author}{\bibinfo{person}{Tomáš Chobola}, \bibinfo{person}{Yu Liu}, \bibinfo{person}{Hanyi Zhang}, \bibinfo{person}{Julia~A. Schnabel}, {and} \bibinfo{person}{Tingying Peng}.} \bibinfo{year}{2024}\natexlab{}.
\newblock \bibinfo{booktitle}{\emph{Fast Context-Based Low-Light Image Enhancement via Neural Implicit Representations}}.
\newblock \bibinfo{publisher}{Springer Nature Switzerland}, \bibinfo{pages}{413–430}.
\newblock
\showISBNx{9783031730160}
\showISSN{1611-3349}
\href{https://doi.org/10.1007/978-3-031-73016-0_24}{doi:\nolinkurl{10.1007/978-3-031-73016-0_24}}


\bibitem[Fei et~al\mbox{.}(2023)]%
        {fei2023generative}
\bibfield{author}{\bibinfo{person}{Ben Fei}, \bibinfo{person}{Zhaoyang Lyu}, \bibinfo{person}{Liang Pan}, \bibinfo{person}{Junzhe Zhang}, \bibinfo{person}{Weidong Yang}, \bibinfo{person}{Tianyue Luo}, \bibinfo{person}{Bo Zhang}, {and} \bibinfo{person}{Bo Dai}.} \bibinfo{year}{2023}\natexlab{}.
\newblock \showarticletitle{Generative diffusion prior for unified image restoration and enhancement}. In \bibinfo{booktitle}{\emph{Proceedings of the IEEE/CVF conference on computer vision and pattern recognition}}. \bibinfo{pages}{9935--9946}.
\newblock


\bibitem[Foi et~al\mbox{.}(2008)]%
        {foi2008practical}
\bibfield{author}{\bibinfo{person}{Alessandro Foi}, \bibinfo{person}{Mejdi Trimeche}, \bibinfo{person}{Vladimir Katkovnik}, {and} \bibinfo{person}{Karen Egiazarian}.} \bibinfo{year}{2008}\natexlab{}.
\newblock \showarticletitle{Practical Poissonian-Gaussian noise modeling and fitting for single-image raw-data}.
\newblock \bibinfo{journal}{\emph{IEEE transactions on image processing}} \bibinfo{volume}{17}, \bibinfo{number}{10} (\bibinfo{year}{2008}), \bibinfo{pages}{1737--1754}.
\newblock


\bibitem[Fu et~al\mbox{.}(2023)]%
        {fu2023learning}
\bibfield{author}{\bibinfo{person}{Zhenqi Fu}, \bibinfo{person}{Yan Yang}, \bibinfo{person}{Xiaotong Tu}, \bibinfo{person}{Yue Huang}, \bibinfo{person}{Xinghao Ding}, {and} \bibinfo{person}{Kai-Kuang Ma}.} \bibinfo{year}{2023}\natexlab{}.
\newblock \showarticletitle{Learning a simple low-light image enhancer from paired low-light instances}. In \bibinfo{booktitle}{\emph{Proceedings of the IEEE/CVF conference on computer vision and pattern recognition}}. \bibinfo{pages}{22252--22261}.
\newblock


\bibitem[Guo et~al\mbox{.}(2020)]%
        {guo2020zero}
\bibfield{author}{\bibinfo{person}{Chunle Guo}, \bibinfo{person}{Chongyi Li}, \bibinfo{person}{Jichang Guo}, \bibinfo{person}{Chen~Change Loy}, \bibinfo{person}{Junhui Hou}, \bibinfo{person}{Sam Kwong}, {and} \bibinfo{person}{Runmin Cong}.} \bibinfo{year}{2020}\natexlab{}.
\newblock \showarticletitle{Zero-reference deep curve estimation for low-light image enhancement}. In \bibinfo{booktitle}{\emph{Proceedings of the IEEE/CVF conference on computer vision and pattern recognition}}. \bibinfo{pages}{1780--1789}.
\newblock


\bibitem[Guo et~al\mbox{.}(2016)]%
        {guo2016lime}
\bibfield{author}{\bibinfo{person}{Xiaojie Guo}, \bibinfo{person}{Yu Li}, {and} \bibinfo{person}{Haibin Ling}.} \bibinfo{year}{2016}\natexlab{}.
\newblock \showarticletitle{LIME: Low-light image enhancement via illumination map estimation}.
\newblock \bibinfo{journal}{\emph{IEEE Transactions on image processing}} \bibinfo{volume}{26}, \bibinfo{number}{2} (\bibinfo{year}{2016}), \bibinfo{pages}{982--993}.
\newblock


\bibitem[Hai et~al\mbox{.}(2023)]%
        {hai2023r2rnet}
\bibfield{author}{\bibinfo{person}{Jiang Hai}, \bibinfo{person}{Zhu Xuan}, \bibinfo{person}{Ren Yang}, \bibinfo{person}{Yutong Hao}, \bibinfo{person}{Fengzhu Zou}, \bibinfo{person}{Fang Lin}, {and} \bibinfo{person}{Songchen Han}.} \bibinfo{year}{2023}\natexlab{}.
\newblock \showarticletitle{R2rnet: Low-light image enhancement via real-low to real-normal network}.
\newblock \bibinfo{journal}{\emph{Journal of Visual Communication and Image Representation}}  \bibinfo{volume}{90} (\bibinfo{year}{2023}), \bibinfo{pages}{103712}.
\newblock


\bibitem[Han et~al\mbox{.}(2026)]%
        {han2026towards}
\bibfield{author}{\bibinfo{person}{Hongru Han}, \bibinfo{person}{Tingrui Guo}, \bibinfo{person}{Liming Zhang}, \bibinfo{person}{Yan Su}, \bibinfo{person}{Qiwen Xu}, {and} \bibinfo{person}{Zhuohua Ye}.} \bibinfo{year}{2026}\natexlab{}.
\newblock \showarticletitle{Towards Controllable Low-Light Image Enhancement: A Continuous Multi-illumination Dataset and Efficient State Space Framework}.
\newblock \bibinfo{journal}{\emph{arXiv preprint arXiv:2603.25296}} (\bibinfo{year}{2026}).
\newblock


\bibitem[Huang et~al\mbox{.}(2022)]%
        {huang2022deep}
\bibfield{author}{\bibinfo{person}{Jie Huang}, \bibinfo{person}{Yajing Liu}, \bibinfo{person}{Feng Zhao}, \bibinfo{person}{Keyu Yan}, \bibinfo{person}{Jinghao Zhang}, \bibinfo{person}{Yukun Huang}, \bibinfo{person}{Man Zhou}, {and} \bibinfo{person}{Zhiwei Xiong}.} \bibinfo{year}{2022}\natexlab{}.
\newblock \showarticletitle{Deep fourier-based exposure correction network with spatial-frequency interaction}. In \bibinfo{booktitle}{\emph{European conference on computer vision}}. Springer, \bibinfo{pages}{163--180}.
\newblock


\bibitem[Huang et~al\mbox{.}(2021)]%
        {huang2021neighbor2neighbor}
\bibfield{author}{\bibinfo{person}{Tao Huang}, \bibinfo{person}{Songjiang Li}, \bibinfo{person}{Xu Jia}, \bibinfo{person}{Huchuan Lu}, {and} \bibinfo{person}{Jianzhuang Liu}.} \bibinfo{year}{2021}\natexlab{}.
\newblock \showarticletitle{Neighbor2Neighbor: Self-Supervised Denoising From Single Noisy Images}. In \bibinfo{booktitle}{\emph{Proceedings of the IEEE/CVF Conference on Computer Vision and Pattern Recognition (CVPR)}}. \bibinfo{pages}{14781--14790}.
\newblock


\bibitem[Jang et~al\mbox{.}(2023)]%
        {jang2023puca}
\bibfield{author}{\bibinfo{person}{Hyemi Jang}, \bibinfo{person}{Junsung Park}, \bibinfo{person}{Dahuin Jung}, \bibinfo{person}{Jaihyun Lew}, \bibinfo{person}{Ho Bae}, {and} \bibinfo{person}{Sungroh Yoon}.} \bibinfo{year}{2023}\natexlab{}.
\newblock \showarticletitle{PUCA: Patch-unshuffle and channel attention for enhanced self-supervised image denoising}.
\newblock \bibinfo{journal}{\emph{Advances in Neural Information Processing Systems}}  \bibinfo{volume}{36} (\bibinfo{year}{2023}), \bibinfo{pages}{19217--19229}.
\newblock


\bibitem[Jiang et~al\mbox{.}(2021a)]%
        {jiang2021focal}
\bibfield{author}{\bibinfo{person}{Liming Jiang}, \bibinfo{person}{Bo Dai}, \bibinfo{person}{Wayne Wu}, {and} \bibinfo{person}{Chen~Change Loy}.} \bibinfo{year}{2021}\natexlab{a}.
\newblock \showarticletitle{Focal frequency loss for image reconstruction and synthesis}. In \bibinfo{booktitle}{\emph{Proceedings of the IEEE/CVF international conference on computer vision}}. \bibinfo{pages}{13919--13929}.
\newblock


\bibitem[Jiang et~al\mbox{.}(2021b)]%
        {jiang2021enlightengan}
\bibfield{author}{\bibinfo{person}{Yifan Jiang}, \bibinfo{person}{Xinyu Gong}, \bibinfo{person}{Ding Liu}, \bibinfo{person}{Yu Cheng}, \bibinfo{person}{Chen Fang}, \bibinfo{person}{Xiaohui Shen}, \bibinfo{person}{Jianchao Yang}, \bibinfo{person}{Pan Zhou}, {and} \bibinfo{person}{Zhangyang Wang}.} \bibinfo{year}{2021}\natexlab{b}.
\newblock \showarticletitle{Enlightengan: Deep light enhancement without paired supervision}.
\newblock \bibinfo{journal}{\emph{IEEE transactions on image processing}}  \bibinfo{volume}{30} (\bibinfo{year}{2021}), \bibinfo{pages}{2340--2349}.
\newblock


\bibitem[Jung et~al\mbox{.}(2025)]%
        {jung2025continuous}
\bibfield{author}{\bibinfo{person}{Donggoo Jung}, \bibinfo{person}{Daehyun Kim}, {and} \bibinfo{person}{Tae~Hyun Kim}.} \bibinfo{year}{2025}\natexlab{}.
\newblock \showarticletitle{Continuous exposure learning for low-light image enhancement using Neural ODEs}. In \bibinfo{booktitle}{\emph{The Thirteenth International Conference on Learning Representations}}.
\newblock


\bibitem[Krull et~al\mbox{.}(2019)]%
        {krull2019noise2void}
\bibfield{author}{\bibinfo{person}{Alexander Krull}, \bibinfo{person}{Tim-Oliver Buchholz}, {and} \bibinfo{person}{Florian Jug}.} \bibinfo{year}{2019}\natexlab{}.
\newblock \showarticletitle{Noise2void-learning denoising from single noisy images}. In \bibinfo{booktitle}{\emph{Proceedings of the IEEE/CVF conference on computer vision and pattern recognition}}. \bibinfo{pages}{2129--2137}.
\newblock


\bibitem[Laine et~al\mbox{.}(2019)]%
        {laine2019high}
\bibfield{author}{\bibinfo{person}{Samuli Laine}, \bibinfo{person}{Tero Karras}, \bibinfo{person}{Jaakko Lehtinen}, {and} \bibinfo{person}{Timo Aila}.} \bibinfo{year}{2019}\natexlab{}.
\newblock \showarticletitle{High-quality self-supervised deep image denoising}.
\newblock \bibinfo{journal}{\emph{Advances in neural information processing systems}}  \bibinfo{volume}{32} (\bibinfo{year}{2019}).
\newblock


\bibitem[Lee et~al\mbox{.}(2022)]%
        {lee2022ap}
\bibfield{author}{\bibinfo{person}{Wooseok Lee}, \bibinfo{person}{Sanghyun Son}, {and} \bibinfo{person}{Kyoung~Mu Lee}.} \bibinfo{year}{2022}\natexlab{}.
\newblock \showarticletitle{Ap-bsn: Self-supervised denoising for real-world images via asymmetric pd and blind-spot network}. In \bibinfo{booktitle}{\emph{Proceedings of the IEEE/CVF conference on computer vision and pattern recognition}}. \bibinfo{pages}{17725--17734}.
\newblock


\bibitem[Lehtinen et~al\mbox{.}(2018)]%
        {lehtinen2018noise2noise}
\bibfield{author}{\bibinfo{person}{Jaakko Lehtinen}, \bibinfo{person}{Jacob Munkberg}, \bibinfo{person}{Jon Hasselgren}, \bibinfo{person}{Samuli Laine}, \bibinfo{person}{Tero Karras}, \bibinfo{person}{Miika Aittala}, {and} \bibinfo{person}{Timo Aila}.} \bibinfo{year}{2018}\natexlab{}.
\newblock \showarticletitle{Noise2Noise: Learning image restoration without clean data}.
\newblock \bibinfo{journal}{\emph{arXiv preprint arXiv:1803.04189}} (\bibinfo{year}{2018}).
\newblock


\bibitem[Li et~al\mbox{.}(2021)]%
        {li2021learning}
\bibfield{author}{\bibinfo{person}{Chongyi Li}, \bibinfo{person}{Chunle Guo}, {and} \bibinfo{person}{Chen~Change Loy}.} \bibinfo{year}{2021}\natexlab{}.
\newblock \showarticletitle{Learning to enhance low-light image via zero-reference deep curve estimation}.
\newblock \bibinfo{journal}{\emph{IEEE transactions on pattern analysis and machine intelligence}} \bibinfo{volume}{44}, \bibinfo{number}{8} (\bibinfo{year}{2021}), \bibinfo{pages}{4225--4238}.
\newblock


\bibitem[Li et~al\mbox{.}({[n.\,d.]})]%
        {li2302embedding}
\bibfield{author}{\bibinfo{person}{C Li}, \bibinfo{person}{CL Guo}, \bibinfo{person}{M Zhou}, \bibinfo{person}{Z Liang}, \bibinfo{person}{S Zhou}, \bibinfo{person}{R Feng}, {and} \bibinfo{person}{CC Loy}.} \bibinfo{year}{[n.\,d.]}\natexlab{}.
\newblock \showarticletitle{Embedding fourier for ultra-high-definition low-light image enhancement. arXiv 2023}.
\newblock \bibinfo{journal}{\emph{arXiv preprint arXiv:2302.11831}} (\bibinfo{year}{[n.\,d.]}).
\newblock


\bibitem[Li et~al\mbox{.}(2023)]%
        {li2023embedding}
\bibfield{author}{\bibinfo{person}{Chongyi Li}, \bibinfo{person}{Chun-Le Guo}, \bibinfo{person}{Man Zhou}, \bibinfo{person}{Zhexin Liang}, \bibinfo{person}{Shangchen Zhou}, \bibinfo{person}{Ruicheng Feng}, {and} \bibinfo{person}{Chen~Change Loy}.} \bibinfo{year}{2023}\natexlab{}.
\newblock \showarticletitle{Embedding fourier for ultra-high-definition low-light image enhancement}.
\newblock \bibinfo{journal}{\emph{arXiv preprint arXiv:2302.11831}} (\bibinfo{year}{2023}).
\newblock


\bibitem[Li et~al\mbox{.}(2025)]%
        {li2025interpretable}
\bibfield{author}{\bibinfo{person}{Huaqiu Li}, \bibinfo{person}{Xiaowan Hu}, {and} \bibinfo{person}{Haoqian Wang}.} \bibinfo{year}{2025}\natexlab{}.
\newblock \showarticletitle{Interpretable unsupervised joint denoising and enhancement for real-world low-light scenarios}.
\newblock \bibinfo{journal}{\emph{arXiv preprint arXiv:2503.14535}} (\bibinfo{year}{2025}).
\newblock


\bibitem[Liang et~al\mbox{.}(2023)]%
        {liang2023iterative}
\bibfield{author}{\bibinfo{person}{Zhexin Liang}, \bibinfo{person}{Chongyi Li}, \bibinfo{person}{Shangchen Zhou}, \bibinfo{person}{Ruicheng Feng}, {and} \bibinfo{person}{Chen~Change Loy}.} \bibinfo{year}{2023}\natexlab{}.
\newblock \showarticletitle{Iterative prompt learning for unsupervised backlit image enhancement}. In \bibinfo{booktitle}{\emph{Proceedings of the IEEE/CVF International Conference on Computer Vision}}. \bibinfo{pages}{8094--8103}.
\newblock


\bibitem[Lim et~al\mbox{.}(2017)]%
        {lim2017enhanced}
\bibfield{author}{\bibinfo{person}{Bee Lim}, \bibinfo{person}{Sanghyun Son}, \bibinfo{person}{Heewon Kim}, \bibinfo{person}{Seungjun Nah}, {and} \bibinfo{person}{Kyoung Mu~Lee}.} \bibinfo{year}{2017}\natexlab{}.
\newblock \showarticletitle{Enhanced deep residual networks for single image super-resolution}. In \bibinfo{booktitle}{\emph{Proceedings of the IEEE conference on computer vision and pattern recognition workshops}}. \bibinfo{pages}{136--144}.
\newblock


\bibitem[Liu et~al\mbox{.}(2021)]%
        {liu2021retinex}
\bibfield{author}{\bibinfo{person}{Risheng Liu}, \bibinfo{person}{Long Ma}, \bibinfo{person}{Jiaao Zhang}, \bibinfo{person}{Xin Fan}, {and} \bibinfo{person}{Zhongxuan Luo}.} \bibinfo{year}{2021}\natexlab{}.
\newblock \showarticletitle{Retinex-inspired unrolling with cooperative prior architecture search for low-light image enhancement}. In \bibinfo{booktitle}{\emph{Proceedings of the IEEE/CVF conference on computer vision and pattern recognition}}. \bibinfo{pages}{10561--10570}.
\newblock


\bibitem[Lore et~al\mbox{.}(2017)]%
        {lore2017llnet}
\bibfield{author}{\bibinfo{person}{Kin~Gwn Lore}, \bibinfo{person}{Adedotun Akintayo}, {and} \bibinfo{person}{Soumik Sarkar}.} \bibinfo{year}{2017}\natexlab{}.
\newblock \showarticletitle{LLNet: A deep autoencoder approach to natural low-light image enhancement}.
\newblock \bibinfo{journal}{\emph{Pattern Recognition}}  \bibinfo{volume}{61} (\bibinfo{year}{2017}), \bibinfo{pages}{650--662}.
\newblock


\bibitem[Lv et~al\mbox{.}(2018)]%
        {lv2018mbllen}
\bibfield{author}{\bibinfo{person}{Feifan Lv}, \bibinfo{person}{Feng Lu}, \bibinfo{person}{Jianhua Wu}, {and} \bibinfo{person}{Chongsoon Lim}.} \bibinfo{year}{2018}\natexlab{}.
\newblock \showarticletitle{MBLLEN: Low-light image/video enhancement using cnns.}. In \bibinfo{booktitle}{\emph{Bmvc}}, Vol.~\bibinfo{volume}{220}. Northumbria University, \bibinfo{pages}{4}.
\newblock


\bibitem[Lv et~al\mbox{.}(2024)]%
        {lv2024fourier}
\bibfield{author}{\bibinfo{person}{Xiaoqian Lv}, \bibinfo{person}{Shengping Zhang}, \bibinfo{person}{Chenyang Wang}, \bibinfo{person}{Yichen Zheng}, \bibinfo{person}{Bineng Zhong}, \bibinfo{person}{Chongyi Li}, {and} \bibinfo{person}{Liqiang Nie}.} \bibinfo{year}{2024}\natexlab{}.
\newblock \showarticletitle{Fourier priors-guided diffusion for zero-shot joint low-light enhancement and deblurring}. In \bibinfo{booktitle}{\emph{Proceedings of the IEEE/CVF conference on computer vision and pattern recognition}}. \bibinfo{pages}{25378--25388}.
\newblock


\bibitem[Ma et~al\mbox{.}(2022)]%
        {ma2022toward}
\bibfield{author}{\bibinfo{person}{Long Ma}, \bibinfo{person}{Tengyu Ma}, \bibinfo{person}{Risheng Liu}, \bibinfo{person}{Xin Fan}, {and} \bibinfo{person}{Zhongxuan Luo}.} \bibinfo{year}{2022}\natexlab{}.
\newblock \showarticletitle{Toward fast, flexible, and robust low-light image enhancement}. In \bibinfo{booktitle}{\emph{Proceedings of the IEEE/CVF conference on computer vision and pattern recognition}}. \bibinfo{pages}{5637--5646}.
\newblock


\bibitem[Nguyen et~al\mbox{.}(2023)]%
        {nguyen2023psenet}
\bibfield{author}{\bibinfo{person}{Hue Nguyen}, \bibinfo{person}{Diep Tran}, \bibinfo{person}{Khoi Nguyen}, {and} \bibinfo{person}{Rang Nguyen}.} \bibinfo{year}{2023}\natexlab{}.
\newblock \showarticletitle{Psenet: Progressive self-enhancement network for unsupervised extreme-light image enhancement}. In \bibinfo{booktitle}{\emph{Proceedings of the IEEE/CVF winter conference on applications of computer vision}}. \bibinfo{pages}{1756--1765}.
\newblock


\bibitem[Parzen(1962)]%
        {parzen1962estimation}
\bibfield{author}{\bibinfo{person}{Emanuel Parzen}.} \bibinfo{year}{1962}\natexlab{}.
\newblock \showarticletitle{On estimation of a probability density function and mode}.
\newblock \bibinfo{journal}{\emph{The annals of mathematical statistics}} \bibinfo{volume}{33}, \bibinfo{number}{3} (\bibinfo{year}{1962}), \bibinfo{pages}{1065--1076}.
\newblock


\bibitem[Rubner et~al\mbox{.}(2000)]%
        {rubner2000earth}
\bibfield{author}{\bibinfo{person}{Yossi Rubner}, \bibinfo{person}{Carlo Tomasi}, {and} \bibinfo{person}{Leonidas~J Guibas}.} \bibinfo{year}{2000}\natexlab{}.
\newblock \showarticletitle{The earth mover's distance as a metric for image retrieval}.
\newblock \bibinfo{journal}{\emph{International journal of computer vision}} \bibinfo{volume}{40}, \bibinfo{number}{2} (\bibinfo{year}{2000}), \bibinfo{pages}{99--121}.
\newblock


\bibitem[Ryou et~al\mbox{.}(2024)]%
        {ryou2024robust}
\bibfield{author}{\bibinfo{person}{Donghun Ryou}, \bibinfo{person}{Inju Ha}, \bibinfo{person}{Hyewon Yoo}, \bibinfo{person}{Dongwan Kim}, {and} \bibinfo{person}{Bohyung Han}.} \bibinfo{year}{2024}\natexlab{}.
\newblock \showarticletitle{Robust image denoising through adversarial frequency mixup}. In \bibinfo{booktitle}{\emph{Proceedings of the IEEE/CVF Conference on Computer Vision and Pattern Recognition}}. \bibinfo{pages}{2723--2732}.
\newblock


\bibitem[Shi et~al\mbox{.}(2024)]%
        {shi2024zero}
\bibfield{author}{\bibinfo{person}{Yiqi Shi}, \bibinfo{person}{Duo Liu}, \bibinfo{person}{Liguo Zhang}, \bibinfo{person}{Ye Tian}, \bibinfo{person}{Xuezhi Xia}, {and} \bibinfo{person}{Xiaojing Fu}.} \bibinfo{year}{2024}\natexlab{}.
\newblock \showarticletitle{ZERO-IG: Zero-shot illumination-guided joint denoising and adaptive enhancement for low-light images}. In \bibinfo{booktitle}{\emph{Proceedings of the IEEE/CVF conference on computer vision and pattern recognition}}. \bibinfo{pages}{3015--3024}.
\newblock


\bibitem[Wang et~al\mbox{.}(2023)]%
        {wang2023fourllie}
\bibfield{author}{\bibinfo{person}{Chenxi Wang}, \bibinfo{person}{Hongjun Wu}, {and} \bibinfo{person}{Zhi Jin}.} \bibinfo{year}{2023}\natexlab{}.
\newblock \showarticletitle{Fourllie: Boosting low-light image enhancement by fourier frequency information}. In \bibinfo{booktitle}{\emph{Proceedings of the 31st ACM international conference on multimedia}}. \bibinfo{pages}{7459--7469}.
\newblock


\bibitem[Wang et~al\mbox{.}(2022b)]%
        {wang2022low}
\bibfield{author}{\bibinfo{person}{Yufei Wang}, \bibinfo{person}{Renjie Wan}, \bibinfo{person}{Wenhan Yang}, \bibinfo{person}{Haoliang Li}, \bibinfo{person}{Lap-Pui Chau}, {and} \bibinfo{person}{Alex Kot}.} \bibinfo{year}{2022}\natexlab{b}.
\newblock \showarticletitle{Low-light image enhancement with normalizing flow}. In \bibinfo{booktitle}{\emph{Proceedings of the AAAI conference on artificial intelligence}}, Vol.~\bibinfo{volume}{36}. \bibinfo{pages}{2604--2612}.
\newblock


\bibitem[Wang et~al\mbox{.}(2022a)]%
        {wang2022blind2unblind}
\bibfield{author}{\bibinfo{person}{Zejin Wang}, \bibinfo{person}{Jiazheng Liu}, \bibinfo{person}{Guoqing Li}, {and} \bibinfo{person}{Hua Han}.} \bibinfo{year}{2022}\natexlab{a}.
\newblock \showarticletitle{Blind2unblind: Self-supervised image denoising with visible blind spots}. In \bibinfo{booktitle}{\emph{Proceedings of the IEEE/CVF conference on computer vision and pattern recognition}}. \bibinfo{pages}{2027--2036}.
\newblock


\bibitem[Wei et~al\mbox{.}(2018)]%
        {wei2018deep}
\bibfield{author}{\bibinfo{person}{Chen Wei}, \bibinfo{person}{Wenjing Wang}, \bibinfo{person}{Wenhan Yang}, {and} \bibinfo{person}{Jiaying Liu}.} \bibinfo{year}{2018}\natexlab{}.
\newblock \showarticletitle{Deep retinex decomposition for low-light enhancement}.
\newblock \bibinfo{journal}{\emph{arXiv preprint arXiv:1808.04560}} (\bibinfo{year}{2018}).
\newblock


\bibitem[Wu et~al\mbox{.}(2022)]%
        {wu2022uretinex}
\bibfield{author}{\bibinfo{person}{Wenhui Wu}, \bibinfo{person}{Jian Weng}, \bibinfo{person}{Pingping Zhang}, \bibinfo{person}{Xu Wang}, \bibinfo{person}{Wenhan Yang}, {and} \bibinfo{person}{Jianmin Jiang}.} \bibinfo{year}{2022}\natexlab{}.
\newblock \showarticletitle{Uretinex-net: Retinex-based deep unfolding network for low-light image enhancement}. In \bibinfo{booktitle}{\emph{Proceedings of the IEEE/CVF conference on computer vision and pattern recognition}}. \bibinfo{pages}{5901--5910}.
\newblock


\bibitem[Xu et~al\mbox{.}(2022)]%
        {xu2022snr}
\bibfield{author}{\bibinfo{person}{Xiaogang Xu}, \bibinfo{person}{Ruixing Wang}, \bibinfo{person}{Chi-Wing Fu}, {and} \bibinfo{person}{Jiaya Jia}.} \bibinfo{year}{2022}\natexlab{}.
\newblock \showarticletitle{Snr-aware low-light image enhancement}. In \bibinfo{booktitle}{\emph{Proceedings of the IEEE/CVF conference on computer vision and pattern recognition}}. \bibinfo{pages}{17714--17724}.
\newblock


\bibitem[Yang et~al\mbox{.}(2023)]%
        {Yang_2023_ICCV}
\bibfield{author}{\bibinfo{person}{Shuzhou Yang}, \bibinfo{person}{Moxuan Ding}, \bibinfo{person}{Yanmin Wu}, \bibinfo{person}{Zihan Li}, {and} \bibinfo{person}{Jian Zhang}.} \bibinfo{year}{2023}\natexlab{}.
\newblock \showarticletitle{Implicit Neural Representation for Cooperative Low-light Image Enhancement}. In \bibinfo{booktitle}{\emph{Proceedings of the IEEE/CVF International Conference on Computer Vision (ICCV)}}. \bibinfo{pages}{12918--12927}.
\newblock


\bibitem[Yang et~al\mbox{.}(2021)]%
        {yang2021sparse}
\bibfield{author}{\bibinfo{person}{Wenhan Yang}, \bibinfo{person}{Wenjing Wang}, \bibinfo{person}{Haofeng Huang}, \bibinfo{person}{Shiqi Wang}, {and} \bibinfo{person}{Jiaying Liu}.} \bibinfo{year}{2021}\natexlab{}.
\newblock \showarticletitle{Sparse gradient regularized deep retinex network for robust low-light image enhancement}.
\newblock \bibinfo{journal}{\emph{IEEE Transactions on Image Processing}}  \bibinfo{volume}{30} (\bibinfo{year}{2021}), \bibinfo{pages}{2072--2086}.
\newblock


\bibitem[Yang and Soatto(2020)]%
        {yang2020fda}
\bibfield{author}{\bibinfo{person}{Yanchao Yang} {and} \bibinfo{person}{Stefano Soatto}.} \bibinfo{year}{2020}\natexlab{}.
\newblock \showarticletitle{Fda: Fourier domain adaptation for semantic segmentation}. In \bibinfo{booktitle}{\emph{Proceedings of the IEEE/CVF conference on computer vision and pattern recognition}}. \bibinfo{pages}{4085--4095}.
\newblock


\bibitem[Yi et~al\mbox{.}(2023)]%
        {yi2023diff}
\bibfield{author}{\bibinfo{person}{Xunpeng Yi}, \bibinfo{person}{Han Xu}, \bibinfo{person}{Hao Zhang}, \bibinfo{person}{Linfeng Tang}, {and} \bibinfo{person}{Jiayi Ma}.} \bibinfo{year}{2023}\natexlab{}.
\newblock \showarticletitle{Diff-retinex: Rethinking low-light image enhancement with a generative diffusion model}. In \bibinfo{booktitle}{\emph{Proceedings of the IEEE/CVF international conference on computer vision}}. \bibinfo{pages}{12302--12311}.
\newblock


\bibitem[Zamir et~al\mbox{.}(2022)]%
        {zamir2022restormer}
\bibfield{author}{\bibinfo{person}{Syed~Waqas Zamir}, \bibinfo{person}{Aditya Arora}, \bibinfo{person}{Salman Khan}, \bibinfo{person}{Munawar Hayat}, \bibinfo{person}{Fahad~Shahbaz Khan}, {and} \bibinfo{person}{Ming-Hsuan Yang}.} \bibinfo{year}{2022}\natexlab{}.
\newblock \showarticletitle{Restormer: Efficient transformer for high-resolution image restoration}. In \bibinfo{booktitle}{\emph{Proceedings of the IEEE/CVF conference on computer vision and pattern recognition}}. \bibinfo{pages}{5728--5739}.
\newblock


\bibitem[Zhang et~al\mbox{.}(2019)]%
        {zhang2019kindling}
\bibfield{author}{\bibinfo{person}{Yonghua Zhang}, \bibinfo{person}{Jiawan Zhang}, {and} \bibinfo{person}{Xiaojie Guo}.} \bibinfo{year}{2019}\natexlab{}.
\newblock \showarticletitle{Kindling the darkness: A practical low-light image enhancer}. In \bibinfo{booktitle}{\emph{Proceedings of the 27th ACM international conference on multimedia}}. \bibinfo{pages}{1632--1640}.
\newblock


\bibitem[Zhao et~al\mbox{.}(2021)]%
        {zhao2021retinexdip}
\bibfield{author}{\bibinfo{person}{Zunjin Zhao}, \bibinfo{person}{Bangshu Xiong}, \bibinfo{person}{Lei Wang}, \bibinfo{person}{Qiaofeng Ou}, \bibinfo{person}{Lei Yu}, {and} \bibinfo{person}{Fa Kuang}.} \bibinfo{year}{2021}\natexlab{}.
\newblock \showarticletitle{RetinexDIP: A unified deep framework for low-light image enhancement}.
\newblock \bibinfo{journal}{\emph{IEEE Transactions on Circuits and Systems for Video Technology}} \bibinfo{volume}{32}, \bibinfo{number}{3} (\bibinfo{year}{2021}), \bibinfo{pages}{1076--1088}.
\newblock


\bibitem[Zheng and Gupta(2022)]%
        {zheng2022semantic}
\bibfield{author}{\bibinfo{person}{Shen Zheng} {and} \bibinfo{person}{Gaurav Gupta}.} \bibinfo{year}{2022}\natexlab{}.
\newblock \showarticletitle{Semantic-guided zero-shot learning for low-light image/video enhancement}. In \bibinfo{booktitle}{\emph{Proceedings of the IEEE/CVF Winter conference on applications of computer vision}}. \bibinfo{pages}{581--590}.
\newblock


\bibitem[Zhou et~al\mbox{.}(2024)]%
        {zhou2024glare}
\bibfield{author}{\bibinfo{person}{Han Zhou}, \bibinfo{person}{Wei Dong}, \bibinfo{person}{Xiaohong Liu}, \bibinfo{person}{Shuaicheng Liu}, \bibinfo{person}{Xiongkuo Min}, \bibinfo{person}{Guangtao Zhai}, {and} \bibinfo{person}{Jun Chen}.} \bibinfo{year}{2024}\natexlab{}.
\newblock \showarticletitle{Glare: Low light image enhancement via generative latent feature based codebook retrieval}. In \bibinfo{booktitle}{\emph{European Conference on Computer Vision}}. Springer, \bibinfo{pages}{36--54}.
\newblock


\bibitem[Zhu et~al\mbox{.}(2020)]%
        {zhu2020zero}
\bibfield{author}{\bibinfo{person}{Anqi Zhu}, \bibinfo{person}{Lin Zhang}, \bibinfo{person}{Ying Shen}, \bibinfo{person}{Yong Ma}, \bibinfo{person}{Shengjie Zhao}, {and} \bibinfo{person}{Yicong Zhou}.} \bibinfo{year}{2020}\natexlab{}.
\newblock \showarticletitle{Zero-shot restoration of underexposed images via robust retinex decomposition}. In \bibinfo{booktitle}{\emph{2020 IEEE international conference on multimedia and expo (ICME)}}. IEEE, \bibinfo{pages}{1--6}.
\newblock


\end{thebibliography}

\clearpage 
\appendix  

\setcounter{section}{0}
\setcounter{equation}{0}
\setcounter{figure}{0}
\setcounter{table}{0}

\renewcommand{\thesection}{A\arabic{section}}
\renewcommand{\thesubsection}{A\arabic{section}.\arabic{subsection}}
\renewcommand{\theequation}{A\arabic{equation}}
\renewcommand{\thefigure}{A\arabic{figure}}
\renewcommand{\thetable}{A\arabic{table}}

\section*{Appendix}

\section{Detailed Formulation of the Cross-Frequency Correlation (CFC)}
\label{sec:supp_cfc}

In the main manuscript, we introduce the Cross-Frequency Correlation (CFC) matrix within the Shift-Invariant Spectral Correlation (SISC) loss. The motivation for designing this metric lies in the internal structural properties of natural textures. Natural images often exhibit nontrivial dependencies across frequency bands due to edges, contours, and structured textures, whereas amplified sensor noise is typically less organized in this regard. Therefore, by calculating the correlation-like cross-band energy dependencies, we can explicitly capture and preserve these underlying structural details while effectively isolating them from the sensor noise.

To formulate this, given the predicted reflectance map $\hat{\mathbf{R}}$, we first compute its 2D Fast Fourier Transform (FFT) and shift the zero-frequency (DC) component to the center of the spectrum. Let $\mathbf{A} = |\mathcal{F}(\hat{\mathbf{R}})|$ denote the amplitude spectrum. We partition the 2D frequency domain into $K$ disjoint concentric radial bands, denoted as $\{B_1, B_2, \dots, B_K\}$. The mask for the $k$-th band is defined based on the radial distance $r(u, v) = \sqrt{u^2 + v^2}$ from the center origin $(0, 0)$:
\begin{equation}
    B_k = \{(u, v) \mid r_{k-1} \le r(u, v) < r_k\},
\end{equation}
where $0 = r_0 < r_1 < \dots < r_K = R_{\text{max}}$ are the predefined radial thresholds evenly dividing the spectrum, and $R_{\text{max}}$ is the maximum frequency radius of the image.

To compute the structural dependencies across frequencies, we extract the global amplitude spectrum $\mathbf{A}$ and compute the average log-energy for each frequency band $k$:
\begin{equation}
    \mathbf{E}_{k} = \log \left( \frac{1}{|B_k|} \sum_{(u, v) \in B_k} \big(\mathbf{A}(u, v)\big)^2 + \epsilon \right),
\end{equation}
where $|B_k|$ is the total number of frequency coordinates contained in the $k$-th band, and $\epsilon = 1 \times 10^{-8}$ is a small constant to prevent numerical instability. By collecting these energies across all $K$ bands, we construct the global log-energy feature vector $\mathbf{E}(\hat{\mathbf{R}}) \in \mathbb{R}^{K \times 1}$.

Next, we compute the normalized cross-band dependency matrix. Let $\mu_E$ and $\sigma_E$ be the scalar mean and standard deviation of the elements within the vector $\mathbf{E}(\hat{\mathbf{R}})$. To align with the matrix operations in our framework, we define $\boldsymbol{\mu}_E \in \mathbb{R}^{K \times 1}$ and $\boldsymbol{\sigma}_E \in \mathbb{R}^{K \times 1}$ as vectors uniformly populated with these scalar values. The CFC matrix $\mathbf{C}(\hat{\mathbf{R}}) \in \mathbb{R}^{K \times K}$ is computed as a normalized outer product:
\begin{equation}
    \mathbf{C}(\hat{\mathbf{R}}) = \frac{\big(\mathbf{E}(\hat{\mathbf{R}}) - \boldsymbol{\mu}_E\big)\big(\mathbf{E}(\hat{\mathbf{R}}) - \boldsymbol{\mu}_E\big)^\top}{\boldsymbol{\sigma}_E \boldsymbol{\sigma}_E^\top + \epsilon}.
\end{equation}

The resulting matrix $\mathbf{C}(\hat{\mathbf{R}})$ is symmetric, with values approximately bounded in $[-1, 1]$ as a normalized correlation measure. From a signal perspective, this normalized outer product effectively captures the \textit{energy co-fluctuation} (or co-activation) between different frequency bands. Specifically, the term $(\mathbf{E} - \boldsymbol{\mu}_E)$ represents the relative energy deviation of each band from the global spectrum average. When two distinct bands $m$ and $n$ simultaneously exhibit energies higher or lower than the average—a characteristic typical for the harmonics of sharp structural edges—their outer product yields a large positive value, indicating strong structural co-activation. Conversely, amplified sensor noise scatters energy unpredictably without such organized co-fluctuation. 

Therefore, the element at the $m$-th row and $n$-th column measures the explicit underlying structural dependency between the $m$-th and $n$-th frequency bands. As discussed in the main text, comparing these matrices between low-light and normal-light conditions guides the design of our binary symmetric mask $\mathbf{M}_{\text{freq}}$, enabling the SISC loss to filter out extreme frequencies and focus exclusively on reliable textural dependencies.

\begin{figure*}[htbp]
  \centering
  \begin{subfigure}{0.24\textwidth}
    \includegraphics[width=\linewidth]{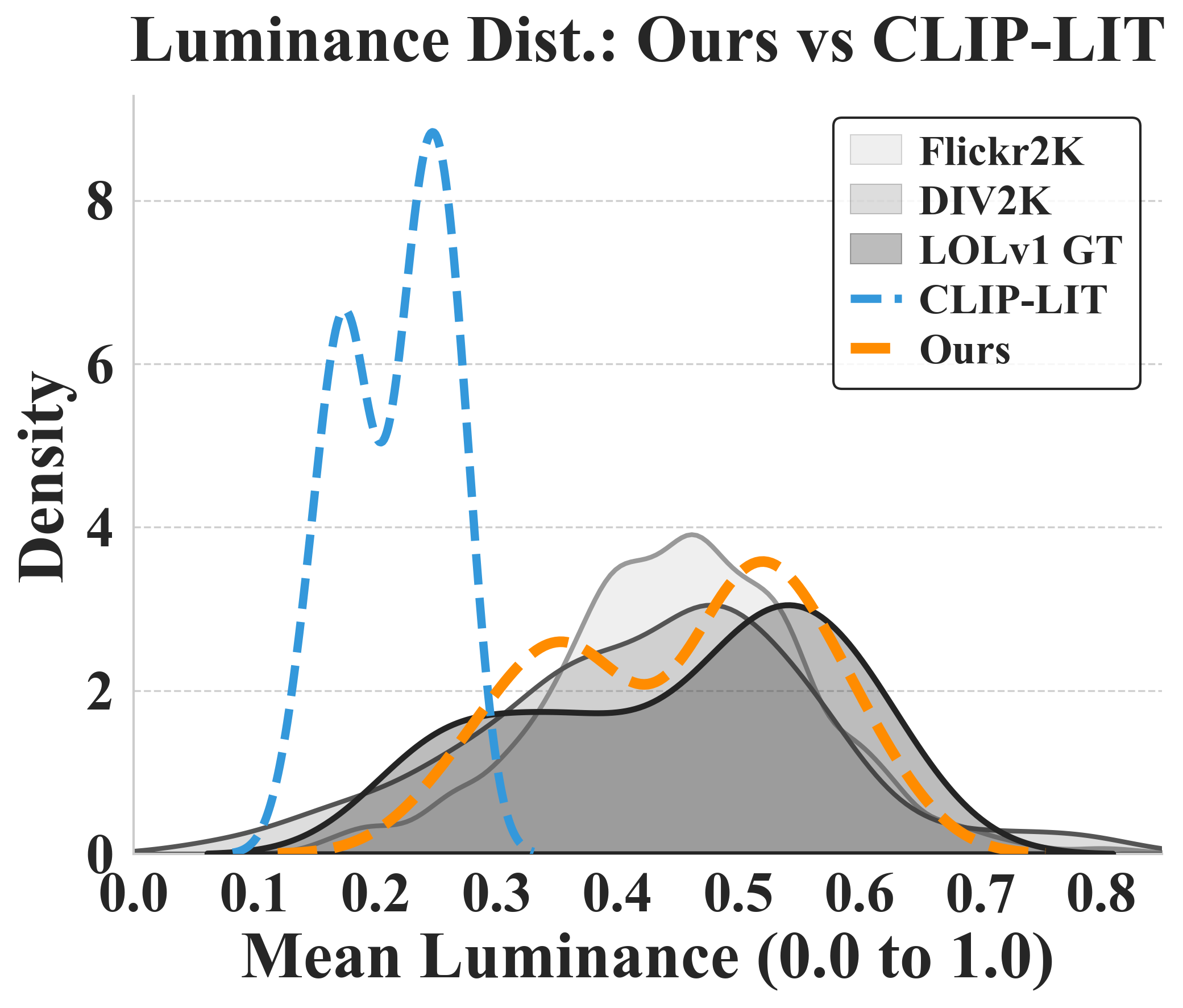}
    \caption{CLIP-LIT (LOLv1)}
  \end{subfigure}\hfill
  \begin{subfigure}{0.24\textwidth}
    \includegraphics[width=\linewidth]{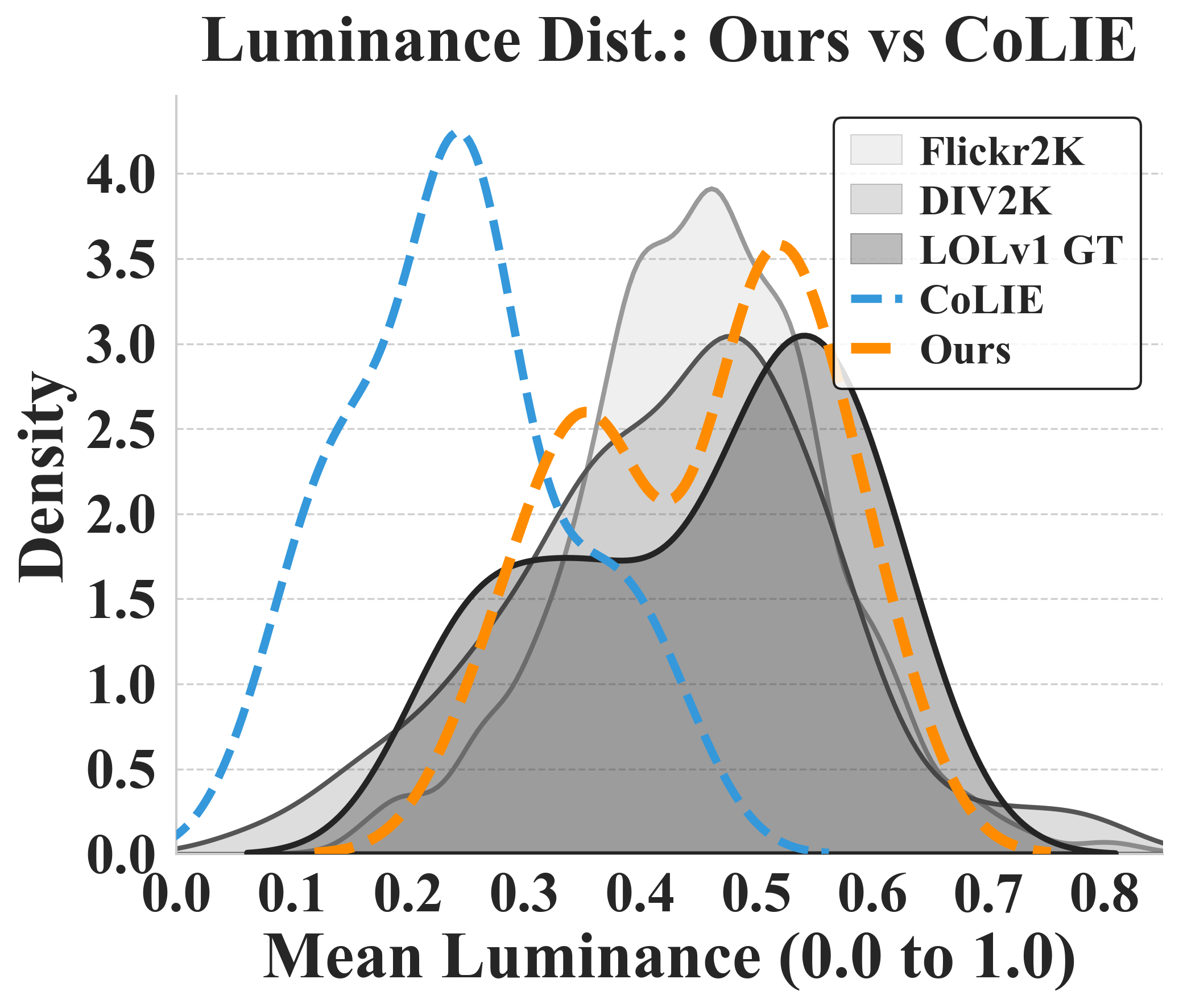}
    \caption{CoLIE (LOLv1)}
  \end{subfigure}\hfill
  \begin{subfigure}{0.24\textwidth}
    \includegraphics[width=\linewidth]{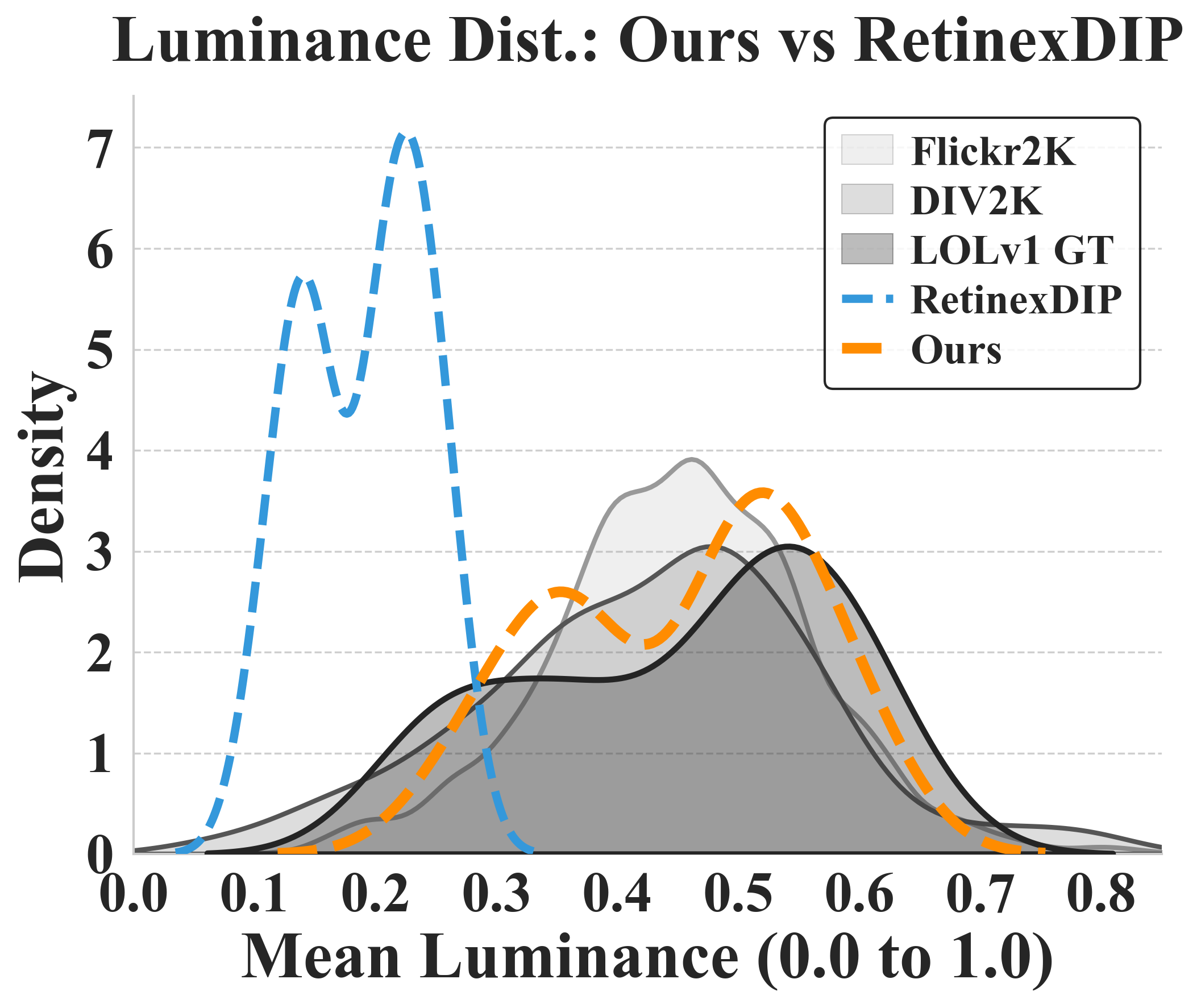}
    \caption{RetinexDIP (LOLv1)}
  \end{subfigure}\hfill
  \begin{subfigure}{0.24\textwidth}
    \includegraphics[width=\linewidth]{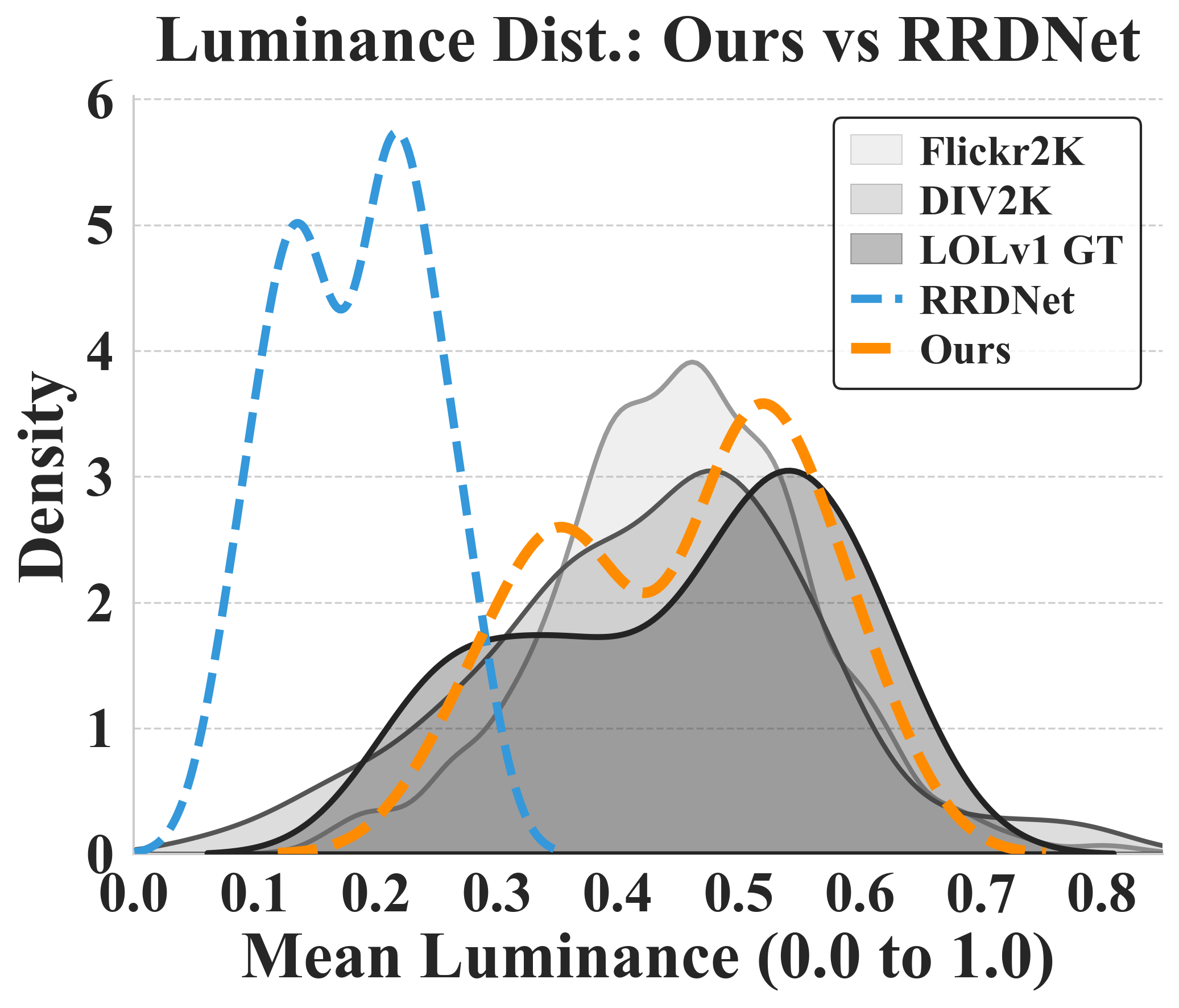}
    \caption{RRDNet (LOLv1)}
  \end{subfigure}
  
  \vspace{0.2cm}
  
  \begin{subfigure}{0.24\textwidth}
    \includegraphics[width=\linewidth]{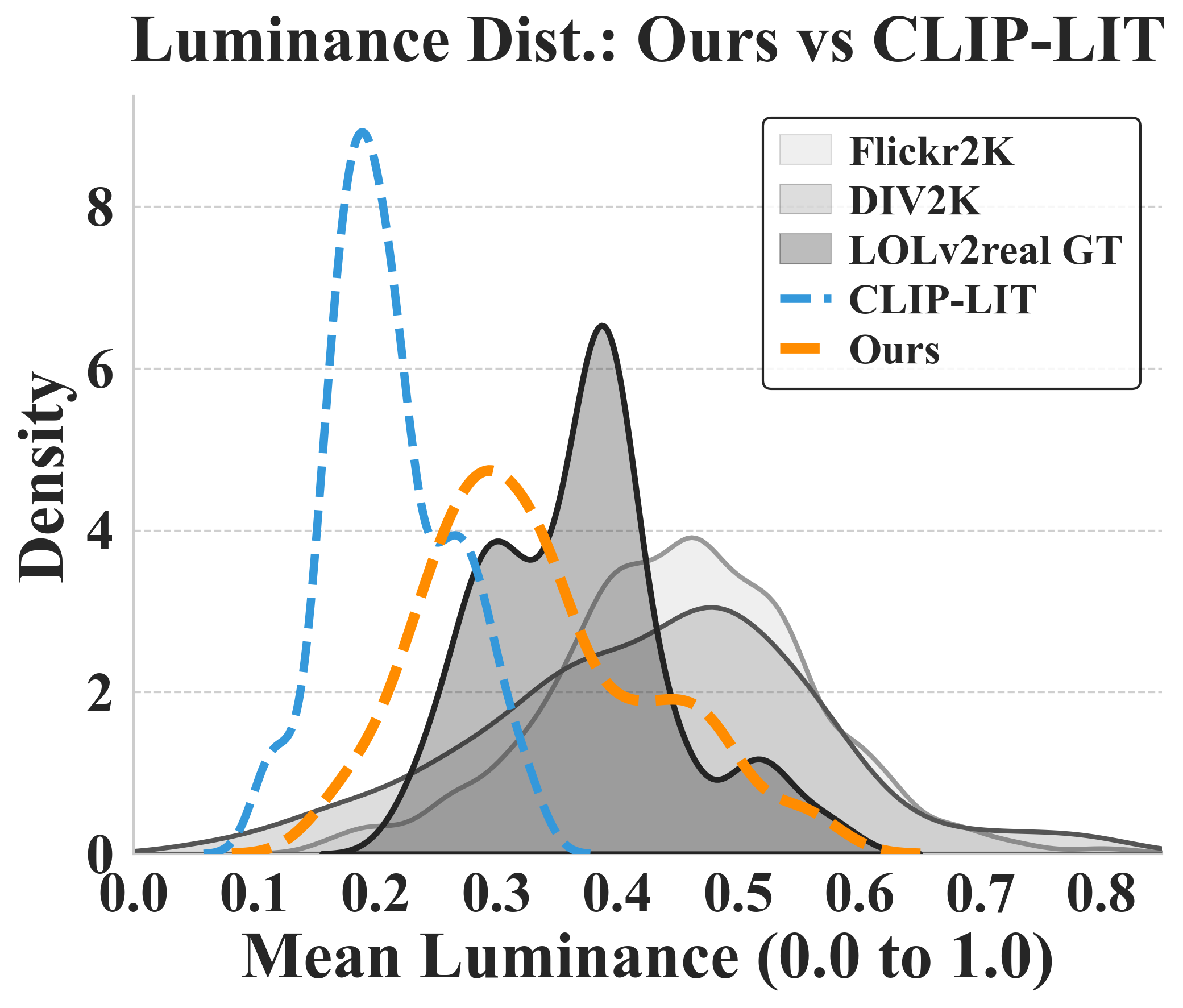}
    \caption{CLIP-LIT (LOLv2-Real)}
  \end{subfigure}\hfill
  \begin{subfigure}{0.24\textwidth}
    \includegraphics[width=\linewidth]{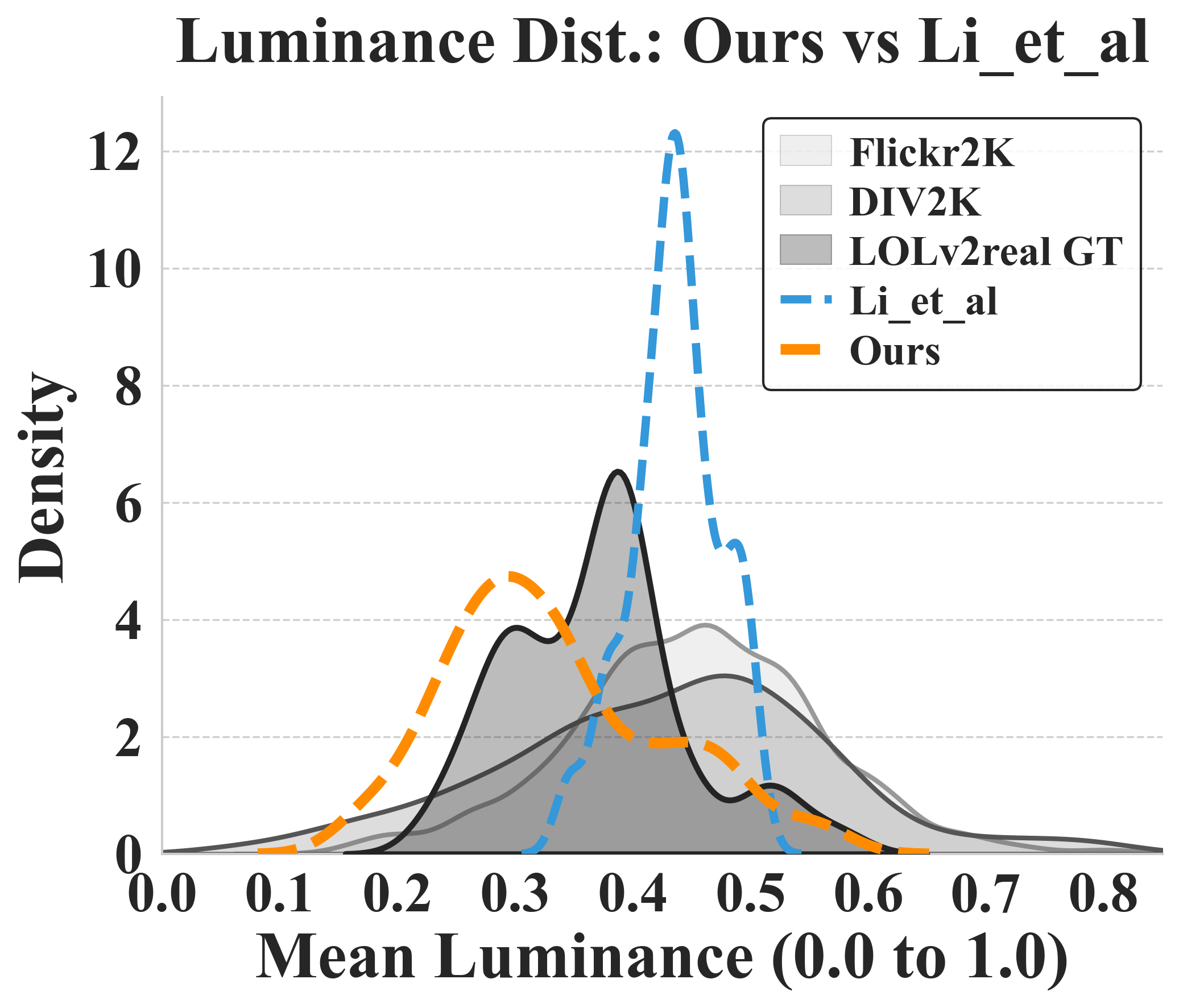}
    \caption{Li (LOLv2-Real)}
  \end{subfigure}\hfill
  \begin{subfigure}{0.24\textwidth}
    \includegraphics[width=\linewidth]{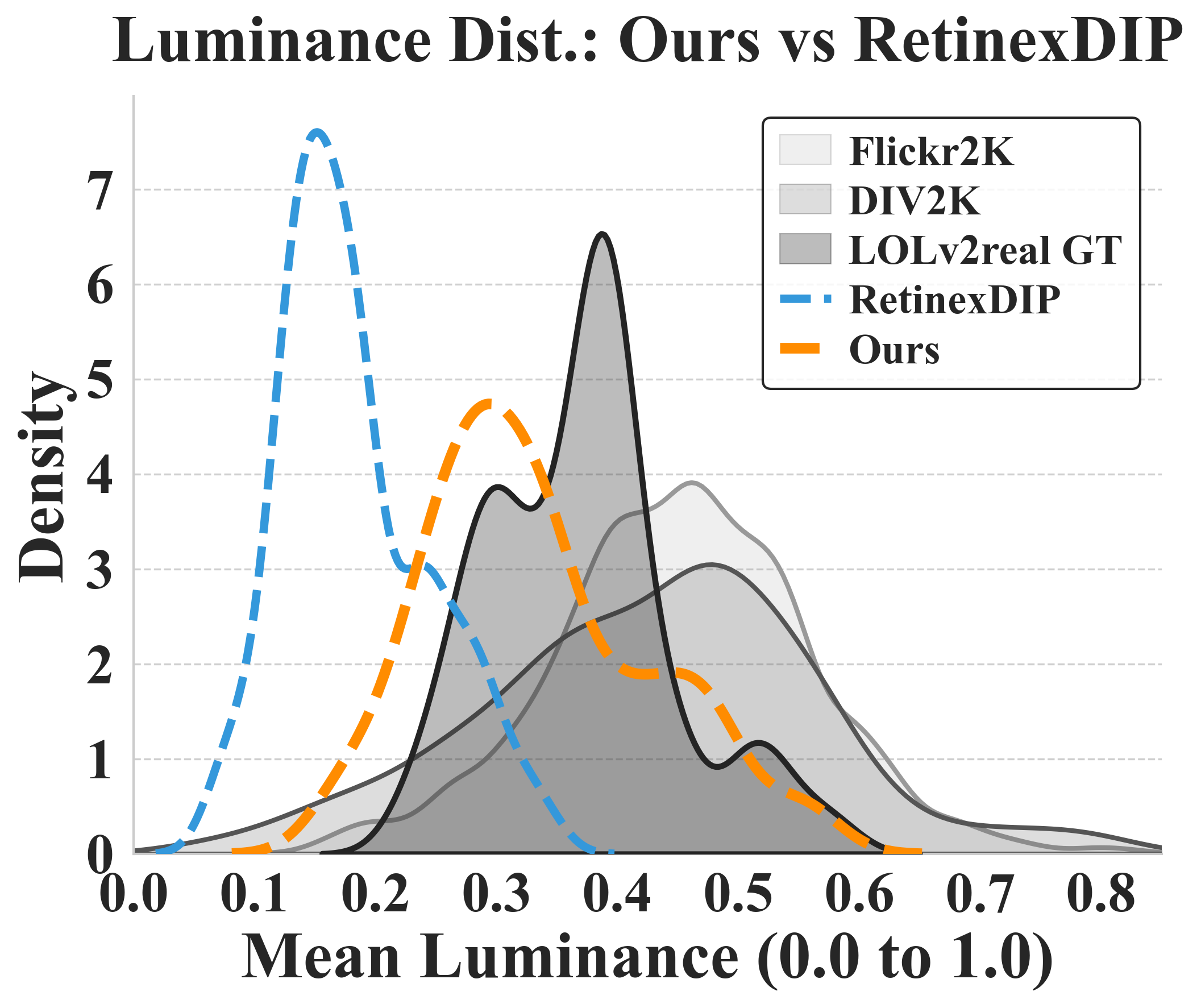}
    \caption{RetinexDIP (LOLv2-Real)}
  \end{subfigure}\hfill
  \begin{subfigure}{0.24\textwidth}
    \includegraphics[width=\linewidth]{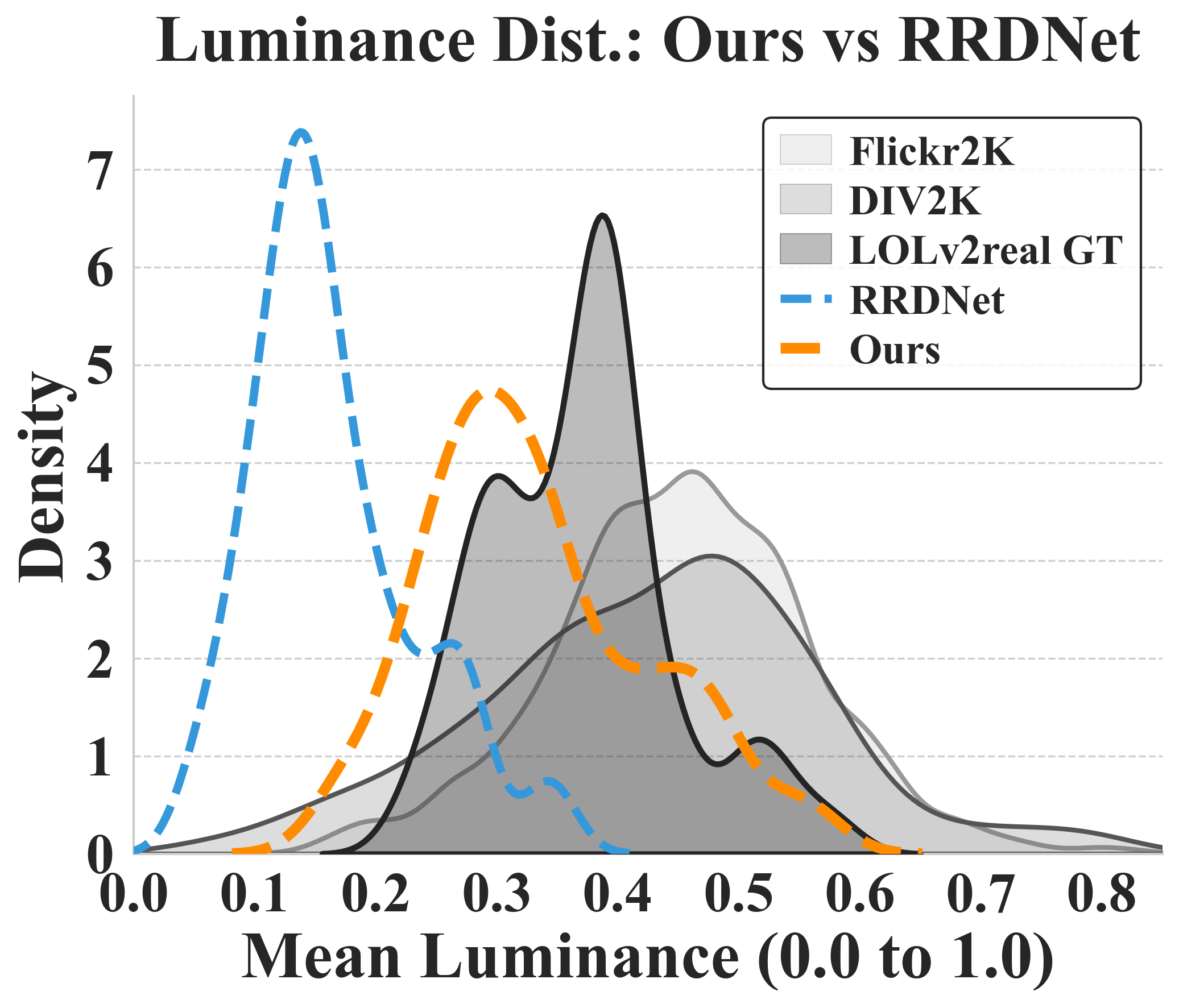}
    \caption{RRDNet (LOLv2-Real)}
  \end{subfigure}
  
  \caption{Luminance distribution comparisons via KDE~\cite{parzen1962estimation}. The orange dashed line represents our method, which consistently aligns with the Ground Truth (gray area) across different datasets, whereas competitors show significant deviations.}
  \label{fig:lum_kde}
\end{figure*}

\section{Derivation of the Optimal Scaling Factor $k$ in the IAP Loss}
\label{sec:supp_k_derivation}

In the main manuscript, we introduced the Illumination-Aligned Perceptual (IAP) loss to preserve global semantic structures. Because computing the perceptual differences directly between the brightened reflectance map $\hat{\mathbf{R}}$ and the dark input $\mathbf{I}_{\text{low}}$ is unstable due to the severe global intensity gap, we introduce a global scalar alignment factor $k$ to bridge this gap. The optimal $k$ has a closed-form solution derived by minimizing the global intensity difference between the scaled reflectance and the low-light input.

We formulate this objective as a standard least-squares optimization problem in the spatial domain. The goal is to find a scalar $k$ that minimizes the squared $L_2$ norm (which is equivalent to the squared Frobenius norm for the vectorized image tensors) of the residual error $\mathcal{E}(k)$:
\begin{equation}
    \min_k \mathcal{E}(k) = \frac{1}{2} \big\| k\hat{\mathbf{R}} - \mathbf{I}_{\text{low}} \big\|_2^2.
\end{equation}

Expanding the squared $L_2$ norm as an inner product, we obtain:
\begin{equation}
\begin{aligned}
    \mathcal{E}(k) &= \frac{1}{2} \langle k\hat{\mathbf{R}} - \mathbf{I}_{\text{low}}, k\hat{\mathbf{R}} - \mathbf{I}_{\text{low}} \rangle \\
    &= \frac{1}{2} \Big( k^2 \langle \hat{\mathbf{R}}, \hat{\mathbf{R}} \rangle - 2k \langle \hat{\mathbf{R}}, \mathbf{I}_{\text{low}} \rangle + \langle \mathbf{I}_{\text{low}}, \mathbf{I}_{\text{low}} \rangle \Big) \\
    &= \frac{1}{2} k^2 \|\hat{\mathbf{R}}\|_2^2 - k \langle \hat{\mathbf{R}}, \mathbf{I}_{\text{low}} \rangle + \frac{1}{2} \|\mathbf{I}_{\text{low}}\|_2^2.
\end{aligned}
\end{equation}

To find the minimum, we compute the partial derivative of $\mathcal{E}(k)$ with respect to $k$ and set it to zero:
\begin{equation}
    \frac{\partial \mathcal{E}(k)}{\partial k} = k \|\hat{\mathbf{R}}\|_2^2 - \langle \hat{\mathbf{R}}, \mathbf{I}_{\text{low}} \rangle = 0.
\end{equation}

Solving this linear equation directly yields the closed-form solution for the optimal scaling factor $k$. To prevent potential division-by-zero errors in completely dark or zero-padded regions during practical network training, we add a very small constant $\epsilon = 1 \times 10^{-8}$ to the denominator, yielding the final expression used in our framework:
\begin{equation}
    k = \frac{\langle \hat{\mathbf{R}}, \mathbf{I}_{\text{low}} \rangle}{\|\hat{\mathbf{R}}\|_2^2 + \epsilon}.
\end{equation}

This optimal scalar $k$ aligns the global energy of $\hat{\mathbf{R}}$ with $\mathbf{I}_{\text{low}}$, reducing the impact of the severe global intensity mismatch. This alignment allows the VGG network to focus more on structural consistency rather than responding to raw brightness discrepancies.

\begin{table}[htbp]
  \centering
  \caption{Quantitative comparison of luminance distributions. Values represent the Wasserstein Distance ($\downarrow$)~\cite{rubner2000earth} to the respective Ground Truth. The \textbf{best} and \underline{second-best} results are highlighted in bold and underline, respectively.}
  \label{tab:w_dist}
  \small
  \begin{tabular}{l c c c c}
    \toprule
    \multicolumn{1}{c}{Method} & LOLv1 & LOLv2-Real & LOLv2-Syn & Mean \\
    \midrule
    RRDNet       & 0.2713 & 0.2060 & 0.1711 & 0.2161 \\
    RetinexDIP   & 0.2631 & 0.1878 & 0.1428 & 0.1979 \\
    CoLIE        & 0.2087 & 0.1643 & 0.1961 & 0.1897 \\
    CLIP-LIT     & 0.2374 & 0.1587 & 0.1325 & 0.1762 \\
    Zero-DCE     & 0.1684 & 0.1070 & 0.0907 & 0.1220 \\
    SCI          & 0.1816 & 0.1179 & 0.0618 & 0.1204 \\
    RUAS         & 0.0792 & 0.0890 & 0.1022 & 0.0901 \\
    EnlightenGAN & 0.0974 & 0.0504 & \underline{0.0552} & 0.0677 \\
    Li    & 0.0652 & 0.0717 & 0.0644 & 0.0671 \\
    CLODE        & \underline{0.0266} & 0.0503 & 0.0636 & 0.0468 \\
    NeRCo        & 0.0325 & \underline{0.0425} & 0.0549 & \underline{0.0433} \\
    \midrule
    \textbf{Ours} & \textbf{0.0234} & \textbf{0.0406} & \textbf{0.0213} & \textbf{0.0285} \\
    \bottomrule
  \end{tabular}
\end{table}

\section{Luminance Distribution Analysis}
\label{sec:luminance_analysis}

In low-light image enhancement (LLIE), a critical challenge is recovering illumination accurately without introducing over-exposure or leaving the image under-enhanced. To comprehensively evaluate the illumination recovery capability of our method, we analyze the global luminance distribution of the enhanced images using both quantitative metrics and visual distribution plots.

Quantitatively, we employ the Wasserstein Distance (W-Dist)~\cite{rubner2000earth} to measure the discrepancy between the luminance distributions of the enhanced results and the corresponding Ground Truth (GT) images. As shown in Table \ref{tab:w_dist}, zero-shot or unsupervised methods (e.g., RRDNet, RetinexDIP, and CLIP-LIT) exhibit significantly high W-Dist scores (above 0.17 on average), indicating severe luminance shifts. While recent state-of-the-art methods narrow this gap, our method achieves the lowest W-Dist across all datasets, with an average W-Dist of 0.0285. This indicates that our method produces luminance distributions that are more closely aligned with the ground-truth distribution than those of competing methods.

Visually, we present the 1D Kernel Density Estimation (KDE)~\cite{parzen1962estimation} of the mean luminance in Figure \ref{fig:lum_kde}. In these plots, alongside the GT, we explicitly include the luminance distributions of high-quality natural image datasets (Flickr2K and DIV2K) to serve as normal-light distribution priors. By plotting the distributions of various methods across specific datasets, we can directly observe how closely each model approximates these natural priors and the target GT. The plots further support that our method aligns more consistently with the Ground Truth and natural image distributions, whereas competitors show noticeable deviations or tend to converge toward a narrow, unnatural luminance range.

\begin{table}[htbp]
  \centering
  \caption{Computational complexity and performance comparison. The restoration quality (Normal PSNR) is evaluated on the LOLv1 dataset. Inference time (RT) and Frames Per Second (FPS) are measured on input images of size $400 \times 600$ using a single NVIDIA RTX 3090 GPU.}
  \label{tab:complexity}
  \resizebox{1.0\columnwidth}{!}{
  \begin{tabular}{l c c c c c}
    \toprule
    Method & PSNR & Params (M) & MACs (G) $\downarrow$ & RT (ms) $\downarrow$ & FPS ($\uparrow$) \\
    \midrule
    GDP & 15.82 & 552.814 & 4872.000 & 1336965.00 & 0.0007 \\
    RRDNet & 11.46 & 0.128 & 31.063 & 21793.18 & 0.0459 \\
    RetinexDIP & 11.67 & 0.707 & 3.414 & 11325.17 & 0.0883 \\
    CoLIE & 13.76 & 0.133 & 8.657 & 1187.32 & 0.84 \\
    Li \textit{et al.} & 19.82 & 0.345 & 78.730 & 295.83 & 3.38 \\
    NeRCo & 19.74 & 23.046 & 1136.000 & 293.67 & 3.41 \\
    CLIP-LIT & 12.39 & 0.279 & 66.670 & 14.53 & 68.80 \\
    PairLIE & 19.51 & 0.342 & 82.929 & 13.69 & 73.06 \\
    EnlightenGAN & 17.56 & 54.410 & 108.878 & 5.63 & 177.76 \\
    PSENet & 17.50 & 0.015 & 0.557 & 3.76 & 265.65 \\
    RUAS & 16.41 & 0.003 & 0.795 & 3.50 & 285.91 \\
    Zero-DCE & 14.86 & 0.079 & 19.008 & 3.44 & 290.81 \\
    SCI & 14.78 & 0.00026 & 0.130 & 0.32 & 3113.86 \\
    \midrule
    \textbf{Ours (Stage 1)} & \textbf{20.39} & 0.887 & 32.479 & 5.09 & 196.40 \\
    \textbf{Ours (Total)} & \textbf{20.60} & 2.204 & 79.139 & 36.26 & 27.58 \\
    \bottomrule
  \end{tabular}
  }
\end{table}

\section{Computational Complexity Analysis}
\label{sec:supp_complexity}

To comprehensively evaluate the practical applicability of our Internally Referenced Low-Light Enhancement (IRLE) framework, we conduct a detailed computational complexity analysis. We compare our proposed method against representative state-of-the-art approaches in terms of restoration quality (Normal PSNR evaluated on the LOLv1 dataset), model size (Parameters), computational cost (MACs), inference time (RT), and Frames Per Second (FPS). All running times and FPS metrics are evaluated on input images of size $400 \times 600$ using a single NVIDIA RTX 3090 GPU.

As detailed in Table \ref{tab:complexity}, existing unsupervised methods can be broadly categorized by their inference types. Diffusion-based methods (e.g., GDP) and Optimization-based methods (e.g., RRDNet, RetinexDIP, CoLIE) require iterative refinement at inference time. Consequently, they suffer from prohibitive inference times (often exceeding thousands of milliseconds per image) and negligible FPS, making them less suitable for latency-sensitive applications.

Among Feed-forward methods, there is typically a trade-off between computational efficiency and restoration performance. Ultra-lightweight models such as Zero-DCE, RUAS, and SCI achieve high FPS but struggle to model complex spatially-variant degradation, resulting in suboptimal PSNR scores (all below 17 dB). Conversely, heavier feed-forward models demand significantly higher parameter counts (e.g., EnlightenGAN and NeRCo possess 54.41 M and 23.046 M parameters, respectively) or suffer from severe latency bottlenecks (e.g., NeRCo and Li \textit{et al.} process at under 4 FPS).

In contrast, our proposed framework offers a highly favorable trade-off between computational efficiency and enhancement performance. To clearly demonstrate this, we evaluate our method in two operational modes:
\begin{itemize}
    \item \textbf{Stage 1 Only:} Our Illumination and Structure Estimation stage alone is lightweight, requiring only 0.887 M parameters and 32.479 G MACs. It achieves a fast inference time of 5.09 ms (196.4 FPS) while delivering a highly competitive PSNR of 20.39 dB, outperforming most existing methods while maintaining a low computational footprint.
    \item \textbf{Total Framework:} When incorporating the Gain-Guided Blind-Spot Denoising (Stage 2) to comprehensively suppress spatially-variant noise, the total parameters and MACs logically increase. However, this full pipeline achieves the overall state-of-the-art PSNR of 20.60 dB. Importantly, with an inference time of 36.26 ms (27.58 FPS), the complete IRLE framework demonstrates practical inference efficiency, supporting near-real-time deployment under our test settings.
\end{itemize}

These results suggest that IRLE offers a favorable balance between restoration quality and efficiency, indicating its practical potential for deployment in real-world low-light scenarios.

\begin{table}[htbp]
  \centering
  \caption{Quantitative comparison using the non-reference metric NIQE ($\downarrow$). Lower scores indicate better perceptual quality and higher naturalness. The \textbf{best} and \underline{second-best} results are highlighted in bold and underline, respectively.}
  \label{tab:niqe}
  \resizebox{\columnwidth}{!}{
  \begin{tabular}{l c c c c}
    \toprule
    \multicolumn{1}{c}{\multirow{2}{*}{Method}} & \multicolumn{2}{c}{LOLv1} & \multicolumn{2}{c}{LOLv2-Real} \\
    \cmidrule(lr){2-3} \cmidrule(lr){4-5}
    & Normal & GT-Mean & Normal & GT-Mean \\
    \midrule
    Zero-DCE~\cite{guo2020zero} & 7.764 & 8.151 & 8.058 & 8.403 \\
    RRDNet~\cite{zhu2020zero} & 7.509 & 8.341 & 7.751 & 8.723 \\
    RUAS~\cite{liu2021retinex} & 6.341 & 6.453 & 6.533 & 6.630 \\
    EnlightenGAN~\cite{jiang2021enlightengan} & 4.582 & 4.564 & 4.874 & 4.848 \\
    RetinexDIP~\cite{zhao2021retinexdip} & 7.660 & 8.355 & 7.858 & 8.621 \\
    SCI~\cite{ma2022toward} & 7.874 & 8.369 & 8.047 & 8.569 \\
    PSENet~\cite{nguyen2023psenet} & 8.045 & 8.151 & 8.340 & 8.410 \\
    QuadPrior    & 5.080 & 4.902 & 4.987 & 4.935 \\
    GDP~\cite{fei2023generative} & 6.160 & 6.281 & 6.486 & 6.657 \\
    CLIP-LIT~\cite{liang2023iterative} & 8.287 & 8.980 & 8.411 & 9.126 \\
    CoLIE~\cite{Chobola2024} & 7.856 & 8.485 & 7.958 & 8.657 \\
    CLODE~\cite{jung2025continuous} & \textbf{4.240} & \underline{4.284} & \underline{4.521} & \underline{4.611} \\
    \midrule
    \textbf{Ours} & \underline{4.279} & \textbf{4.231} & \textbf{4.218} & \textbf{4.036} \\
    \bottomrule
  \end{tabular}
  }
\end{table}

\begin{figure*}[htbp]
    \centering
    \begin{subfigure}[t]{0.19\textwidth}
        \centering
        \includegraphics[width=\textwidth]{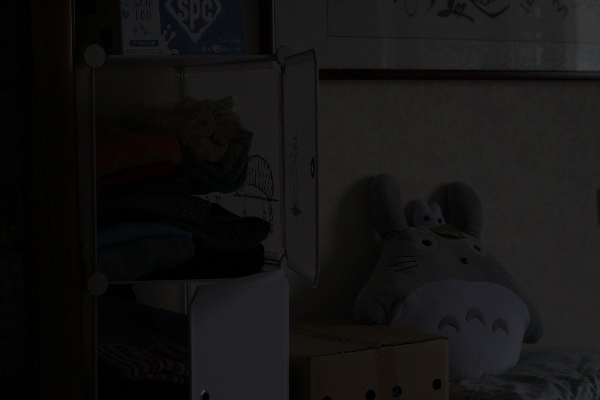}
        \caption{Input}
        \label{fig:lolv1_22_input}
    \end{subfigure}
    \hfill
    \begin{subfigure}[t]{0.19\textwidth}
        \centering
        \includegraphics[width=\textwidth]{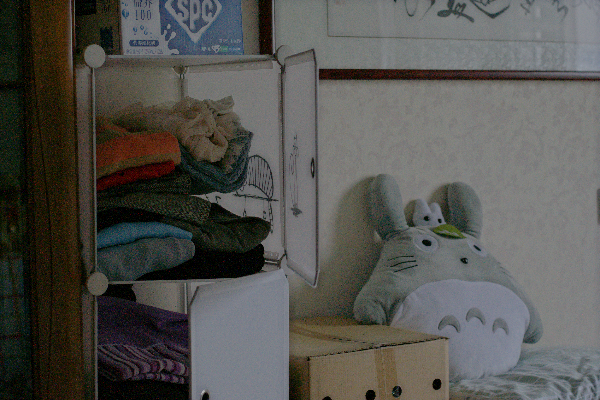}
        \caption{Zero-DCE~\cite{guo2020zero}}
        \label{fig:lolv1_22_zerodce}
    \end{subfigure}
    \hfill
    \begin{subfigure}[t]{0.19\textwidth}
        \centering
        \includegraphics[width=\textwidth]{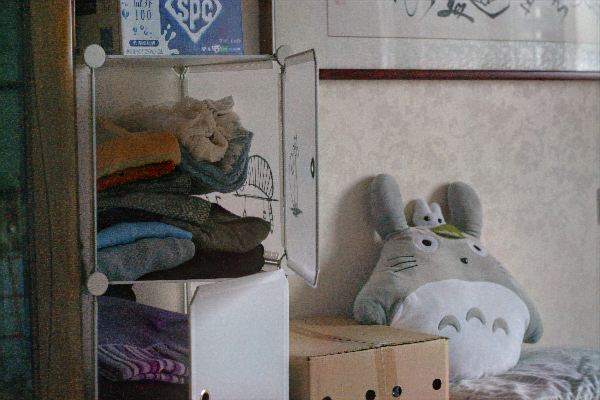}
        \caption{EnlightenGAN~\cite{jiang2021enlightengan}}
        \label{fig:lolv1_22_engan}
    \end{subfigure}
    \hfill
    \begin{subfigure}[t]{0.19\textwidth}
        \centering
        \includegraphics[width=\textwidth]{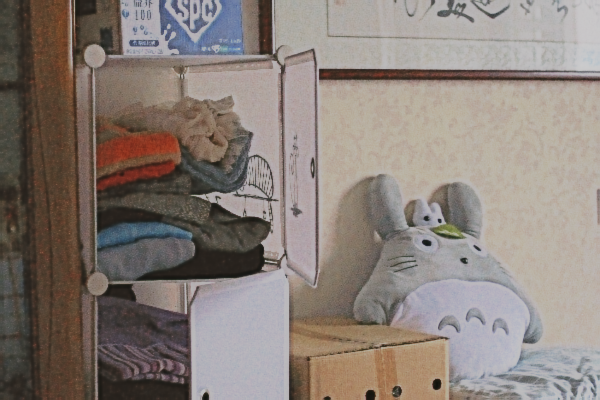}
        \caption{PairLIE~\cite{fu2023learning}}
        \label{fig:lolv1_22_pairlie}
    \end{subfigure}
    \hfill
    \begin{subfigure}[t]{0.19\textwidth}
        \centering
        \includegraphics[width=\textwidth]{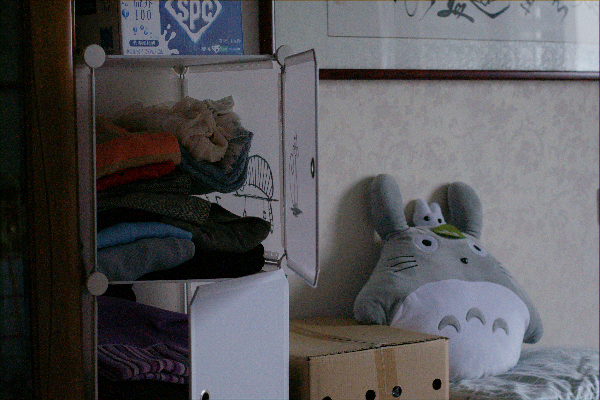}
        \caption{SCI~\cite{ma2022toward}}
        \label{fig:lolv1_22_sci}
    \end{subfigure}
    
    \medskip 
    
    \begin{subfigure}[t]{0.19\textwidth}
        \centering
        \includegraphics[width=\textwidth]{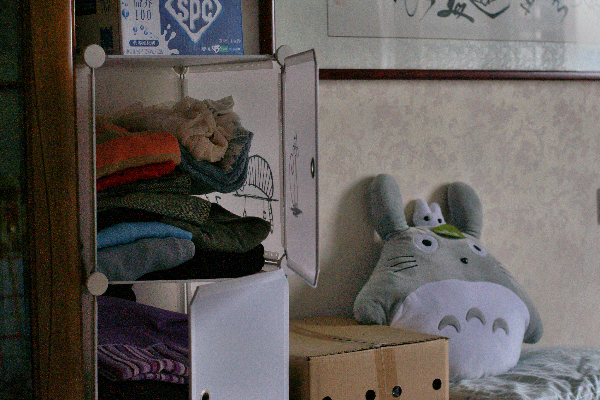}
        \caption{CoLIE~\cite{Chobola2024}}
        \label{fig:lolv1_22_colie}
    \end{subfigure}
    \hfill
    \begin{subfigure}[t]{0.19\textwidth}
        \centering
        \includegraphics[width=\textwidth]{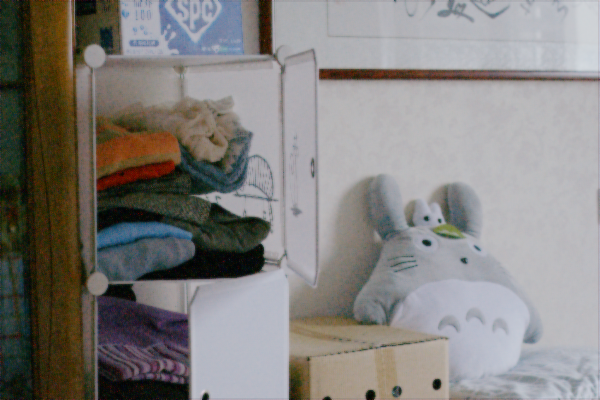}
        \caption{CLODE~\cite{jung2025continuous}}
        \label{fig:lolv1_22_clode}
    \end{subfigure}
    \hfill
    \begin{subfigure}[t]{0.19\textwidth}
        \centering
        \includegraphics[width=\textwidth]{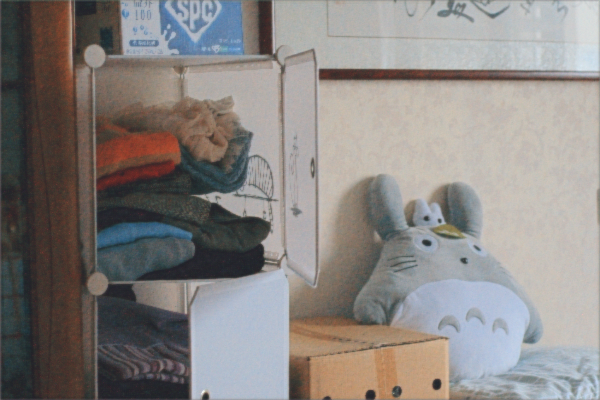}
        \caption{Li \textit{et al.}~\cite{li2025interpretable}}
        \label{fig:lolv1_22_li}
    \end{subfigure}
    \hfill
    \begin{subfigure}[t]{0.19\textwidth}
        \centering
        \includegraphics[width=\textwidth]{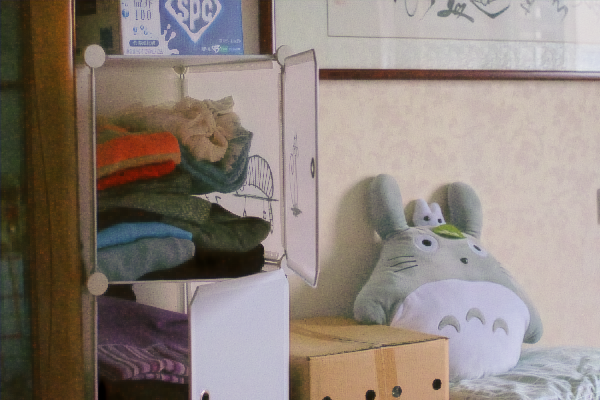}
        \caption{\textbf{Ours}}
        \label{fig:lolv1_22_ours}
    \end{subfigure}
    \hfill
    \begin{subfigure}[t]{0.19\textwidth}
        \centering
        \includegraphics[width=\textwidth]{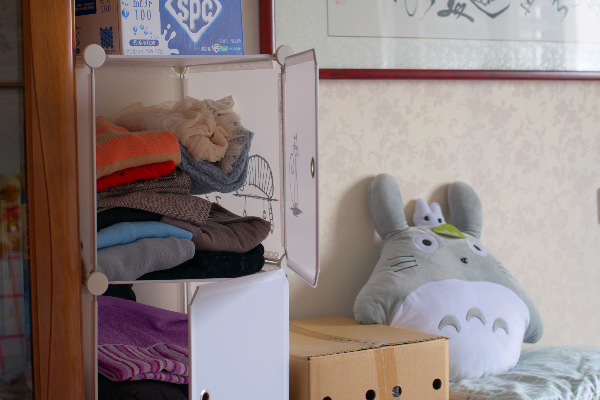}
        \caption{GT}
        \label{fig:lolv1_22_gt}
    \end{subfigure}

    \Description{A visual comparison on the LOLv1 dataset. Competitor methods exhibit visible noise or color shifts. In contrast, 'Ours' displays clear textures and accurate colors, closely matching the Ground Truth.}
    \caption{\textbf{Visual comparison on LOLv1~\cite{wei2018deep}.} Existing methods often exhibit color casts or lose delicate details due to over-smoothing. Our method maintains natural colors and effectively removes spatially-variant noise while preserving structures.}
    \label{fig:supp_lolv1_22}
\end{figure*}

\begin{figure*}[htbp]
    \centering
    \begin{subfigure}[t]{0.19\textwidth}
        \centering
        \includegraphics[width=\textwidth]{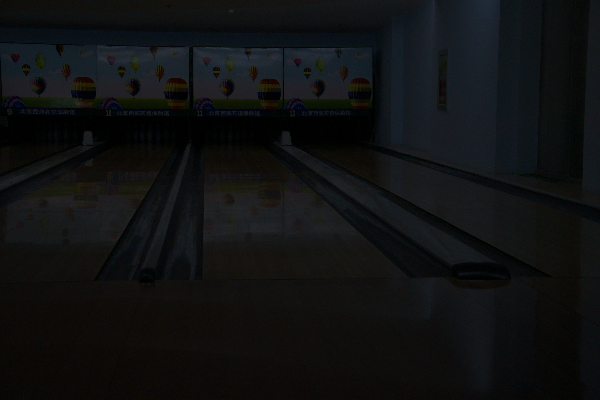}
        \caption{Input}
        \label{fig:lolv1_669_input}
    \end{subfigure}
    \hfill
    \begin{subfigure}[t]{0.19\textwidth}
        \centering
        \includegraphics[width=\textwidth]{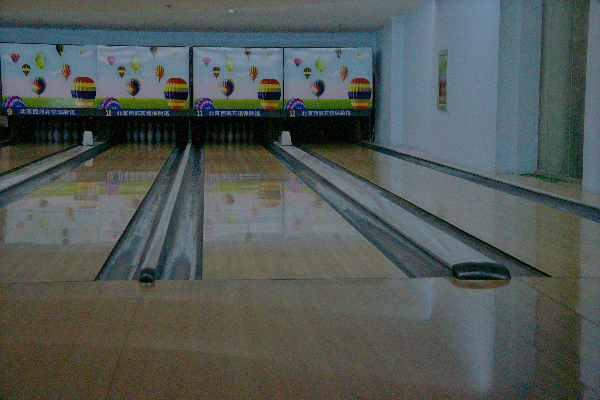}
        \caption{Zero-DCE~\cite{guo2020zero}}
        \label{fig:lolv1_669_zerodce}
    \end{subfigure}
    \hfill
    \begin{subfigure}[t]{0.19\textwidth}
        \centering
        \includegraphics[width=\textwidth]{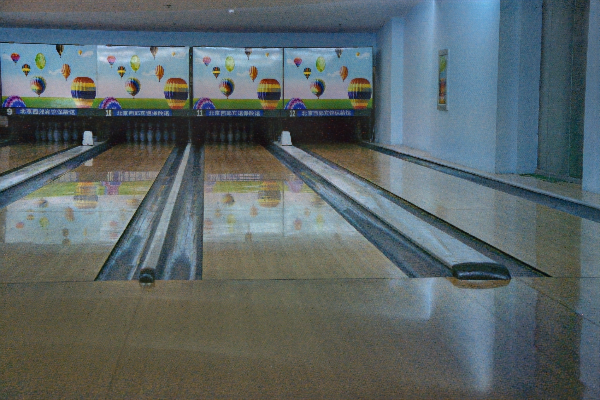}
        \caption{EnlightenGAN~\cite{jiang2021enlightengan}}
        \label{fig:lolv1_669_engan}
    \end{subfigure}
    \hfill
    \begin{subfigure}[t]{0.19\textwidth}
        \centering
        \includegraphics[width=\textwidth]{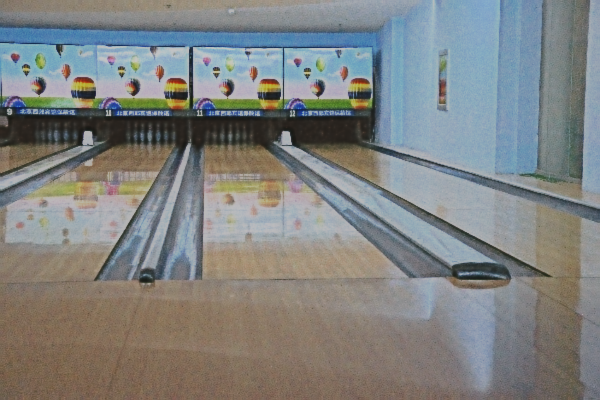}
        \caption{PairLIE~\cite{fu2023learning}}
        \label{fig:lolv1_669_pairlie}
    \end{subfigure}
    \hfill
    \begin{subfigure}[t]{0.19\textwidth}
        \centering
        \includegraphics[width=\textwidth]{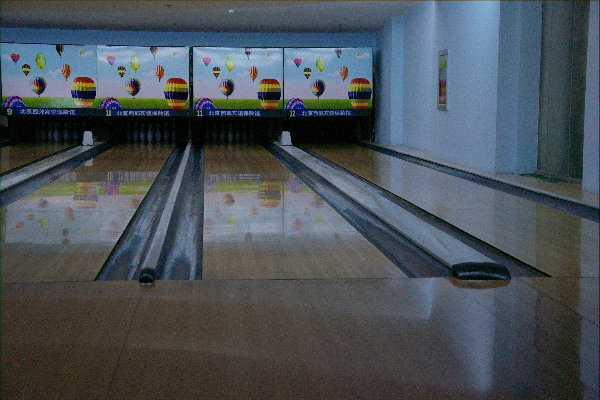}
        \caption{SCI~\cite{ma2022toward}}
        \label{fig:lolv1_669_sci}
    \end{subfigure}
    
    \medskip 
    
    \begin{subfigure}[t]{0.19\textwidth}
        \centering
        \includegraphics[width=\textwidth]{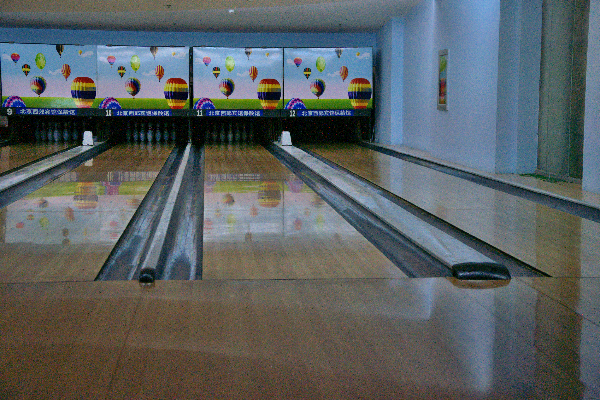}
        \caption{CoLIE~\cite{Chobola2024}}
        \label{fig:lolv1_669_colie}
    \end{subfigure}
    \hfill
    \begin{subfigure}[t]{0.19\textwidth}
        \centering
        \includegraphics[width=\textwidth]{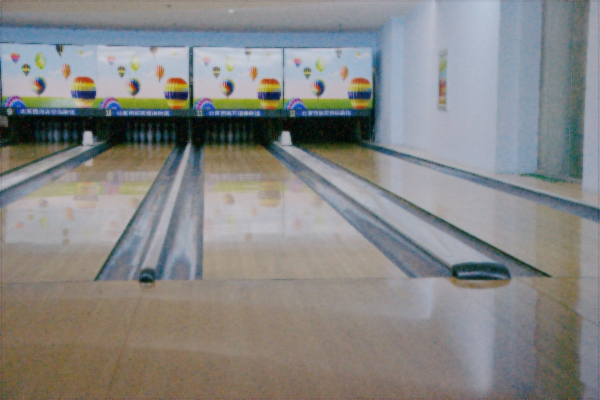}
        \caption{CLODE~\cite{jung2025continuous}}
        \label{fig:lolv1_669_clode}
    \end{subfigure}
    \hfill
    \begin{subfigure}[t]{0.19\textwidth}
        \centering
        \includegraphics[width=\textwidth]{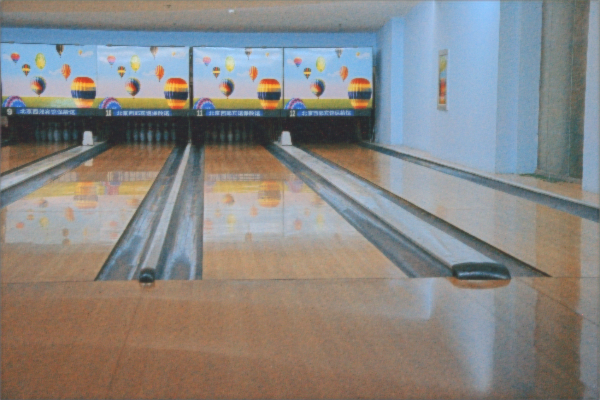}
        \caption{Li \textit{et al.}~\cite{li2025interpretable}}
        \label{fig:lolv1_669_li}
    \end{subfigure}
    \hfill
    \begin{subfigure}[t]{0.19\textwidth}
        \centering
        \includegraphics[width=\textwidth]{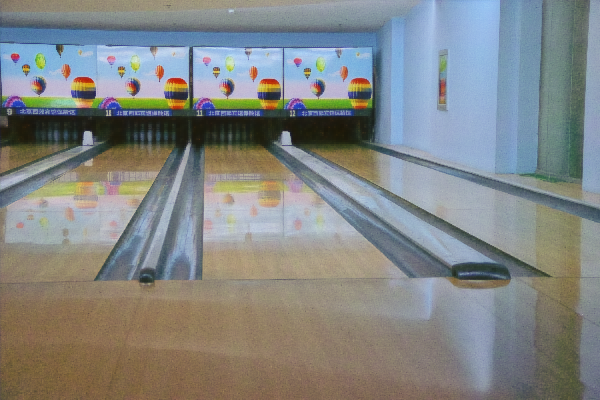}
        \caption{\textbf{Ours}}
        \label{fig:lolv1_669_ours}
    \end{subfigure}
    \hfill
    \begin{subfigure}[t]{0.19\textwidth}
        \centering
        \includegraphics[width=\textwidth]{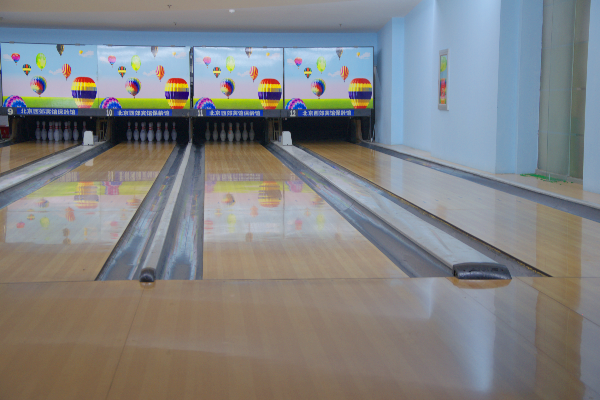}
        \caption{GT}
        \label{fig:lolv1_669_gt}
    \end{subfigure}

    \Description{A visual comparison on the LOLv1 dataset. Competitor methods exhibit visible noise or color shifts. In contrast, 'Ours' displays clear textures and accurate colors, closely matching the Ground Truth.}
    \caption{\textbf{Visual comparison on LOLv1~\cite{wei2018deep}.} Existing methods often exhibit color casts or lose delicate details due to over-smoothing. Our method maintains natural colors and effectively removes spatially-variant noise while preserving structures.}
    \label{fig:supp_lolv1_669}
\end{figure*}

\section{Non-Reference Perceptual Evaluation}
\label{sec:supp_niqe}

In addition to the full-reference metrics (PSNR and SSIM) evaluated in the main manuscript, we further assess the visual and perceptual quality of the enhanced images using the Natural Image Quality Evaluator (NIQE). As a no-reference image quality metric, NIQE evaluates the naturalness of an image based on deviations from statistical regularities observed in natural, pristine image datasets. A lower NIQE score indicates higher perceptual quality, presenting fewer unnatural artifacts such as amplified sensor noise, severe color casts, or over-smoothed textures. 

As shown in Table \ref{tab:niqe}, we report the NIQE scores on both the LOLv1 and LOLv2-Real datasets under standard (Normal) and GT-Mean evaluation settings. Unsupervised methods that struggle to decouple entangled noise and textures (e.g., SCI, PSENet, and CLIP-LIT) consistently yield high NIQE scores (around 8.0 to 9.0), reflecting noticeable perceptual degradation and unnatural visual artifacts. Even among strong competitors like EnlightenGAN and CLODE, our proposed IRLE framework demonstrates competitive perceptual performance. 

Specifically, our method achieves the best (lowest) NIQE scores across almost all settings, including LOLv1 (GT-Mean) and LOLv2-Real (both Normal and GT-Mean), while securing a highly competitive second-best score on LOLv1 (Normal). This non-reference evaluation solidifies our claim that by utilizing an internally referenced perspective and the GAFM dynamic denoising mechanism, our framework not only restores global illumination accurately but also produces results with exceptional visual naturalness and textural fidelity.

\section{More Visual Comparisons}
\label{sec:supp_more_visuals}

\begin{figure*}[htbp]
    \centering
    \begin{subfigure}[t]{0.19\textwidth}
        \centering
        \includegraphics[width=\textwidth]{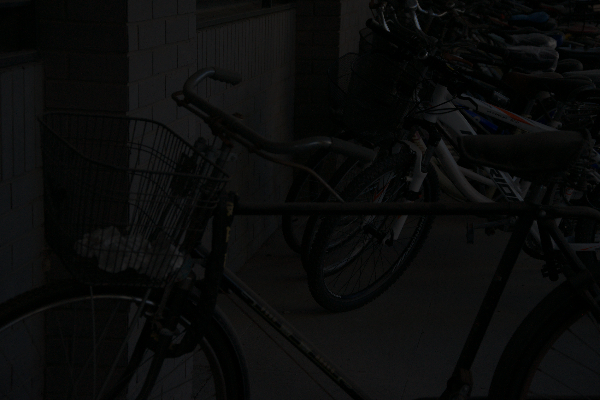}
        \caption{Input}
        \label{fig:lolv2real_721_input}
    \end{subfigure}
    \hfill
    \begin{subfigure}[t]{0.19\textwidth}
        \centering
        \includegraphics[width=\textwidth]{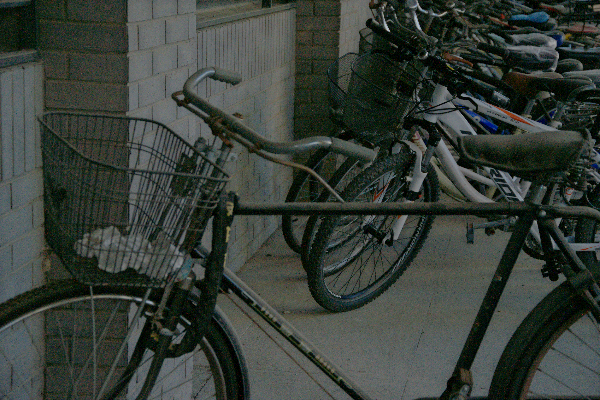}
        \caption{Zero-DCE~\cite{guo2020zero}}
        \label{fig:lolv2real_721_zerodce}
    \end{subfigure}
    \hfill
    \begin{subfigure}[t]{0.19\textwidth}
        \centering
        \includegraphics[width=\textwidth]{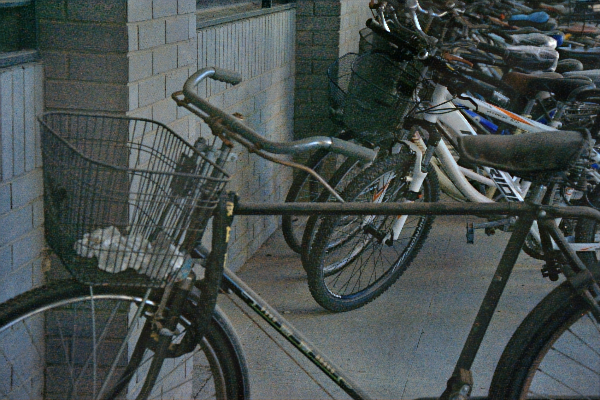}
        \caption{EnlightenGAN~\cite{jiang2021enlightengan}}
        \label{fig:lolv2real_721_engan}
    \end{subfigure}
    \hfill
    \begin{subfigure}[t]{0.19\textwidth}
        \centering
        \includegraphics[width=\textwidth]{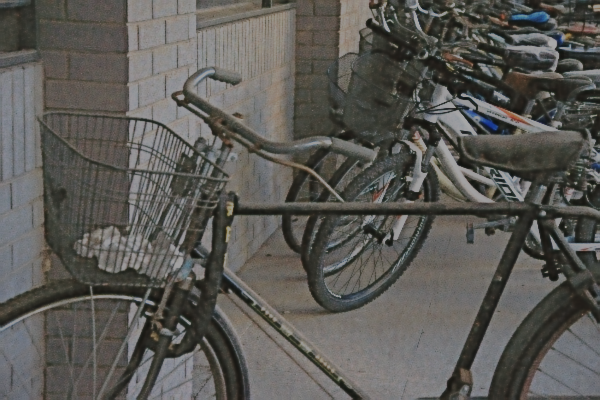}
        \caption{PairLIE~\cite{fu2023learning}}
        \label{fig:lolv2real_721_pairlie}
    \end{subfigure}
    \hfill
    \begin{subfigure}[t]{0.19\textwidth}
        \centering
        \includegraphics[width=\textwidth]{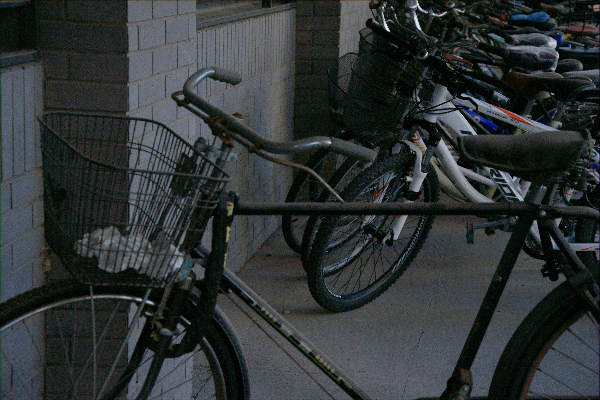}
        \caption{SCI~\cite{ma2022toward}}
        \label{fig:lolv2real_721_sci}
    \end{subfigure}
    
    \medskip 

    \begin{subfigure}[t]{0.19\textwidth}
        \centering
        \includegraphics[width=\textwidth]{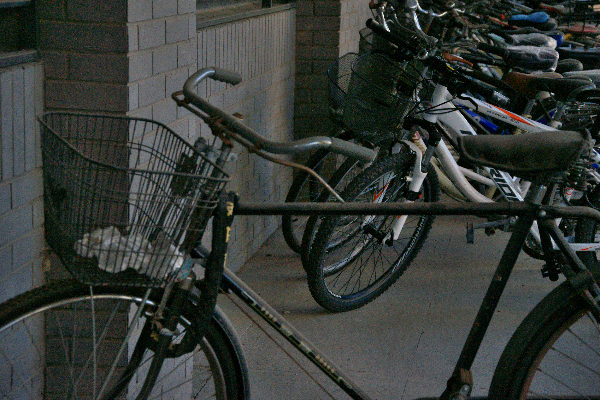}
        \caption{CoLIE~\cite{Chobola2024}}
        \label{fig:lolv2real_721_colie}
    \end{subfigure}
    \hfill
    \begin{subfigure}[t]{0.19\textwidth}
        \centering
        \includegraphics[width=\textwidth]{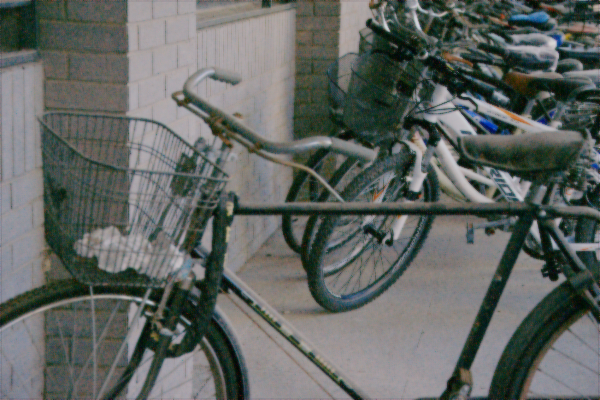}
        \caption{CLODE~\cite{jung2025continuous}}
        \label{fig:lolv2real_721_clode}
    \end{subfigure}
    \hfill
    \begin{subfigure}[t]{0.19\textwidth}
        \centering
        \includegraphics[width=\textwidth]{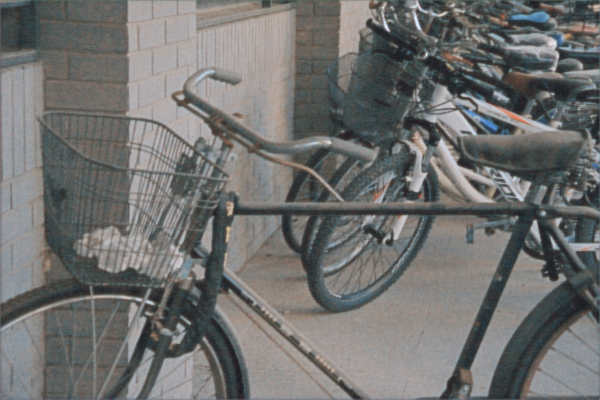}
        \caption{Li \textit{et al.}~\cite{li2025interpretable}}
        \label{fig:lolv2real_721_li}
    \end{subfigure}
    \hfill
    \begin{subfigure}[t]{0.19\textwidth}
        \centering
        \includegraphics[width=\textwidth]{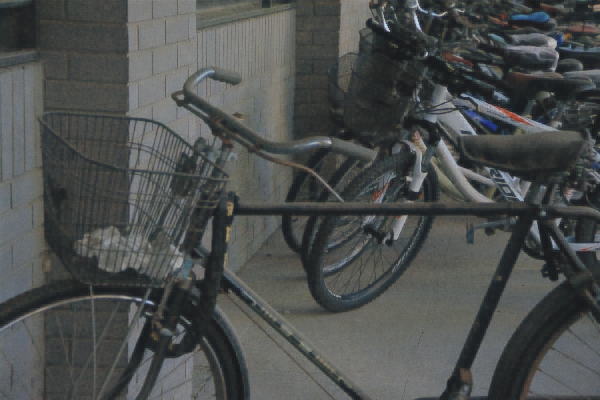}
        \caption{\textbf{Ours}}
        \label{fig:lolv2real_721_ours}
    \end{subfigure}
    \hfill
    \begin{subfigure}[t]{0.19\textwidth}
        \centering
        \includegraphics[width=\textwidth]{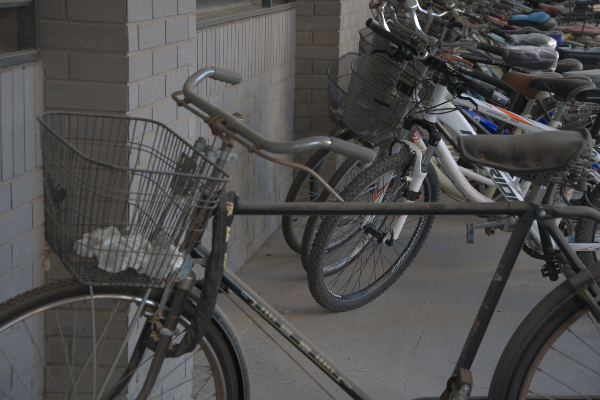}
        \caption{GT}
        \label{fig:lolv2real_721_gt}
    \end{subfigure}

    \Description{A visual comparison on the LOLv2-Real dataset. Competitor methods exhibit visible noise or color shifts. In contrast, 'Ours' displays clear textures and accurate colors, closely matching the Ground Truth.}
    \caption{\textbf{Visual comparison on LOLv2-Real~\cite{yang2021sparse}.} Our method restores accurate global illumination and fine textures, demonstrating robust performance on diverse real-world scenes.}
    \label{fig:supp_lolv2real_721}
\end{figure*}

\begin{figure*}[htbp]
    \centering
    \begin{subfigure}[t]{0.19\textwidth}
        \centering
        \includegraphics[width=\textwidth]{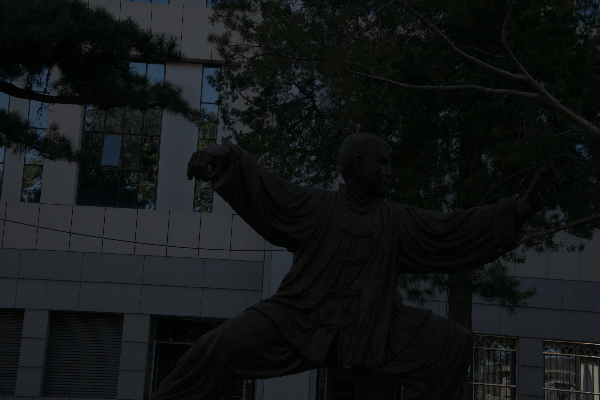}
        \caption{Input}
        \label{fig:lolv2real_736_input}
    \end{subfigure}
    \hfill
    \begin{subfigure}[t]{0.19\textwidth}
        \centering
        \includegraphics[width=\textwidth]{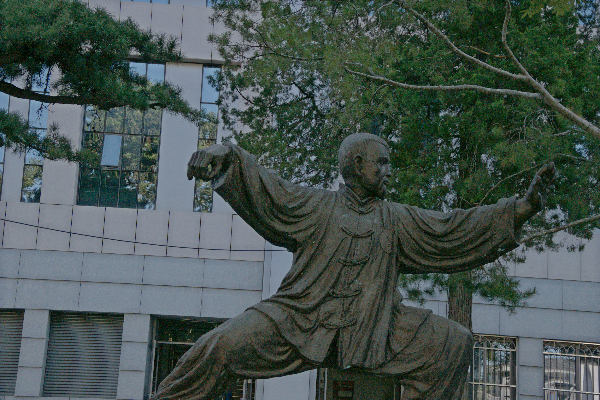}
        \caption{Zero-DCE~\cite{guo2020zero}}
        \label{fig:lolv2real_736_zerodce}
    \end{subfigure}
    \hfill
    \begin{subfigure}[t]{0.19\textwidth}
        \centering
        \includegraphics[width=\textwidth]{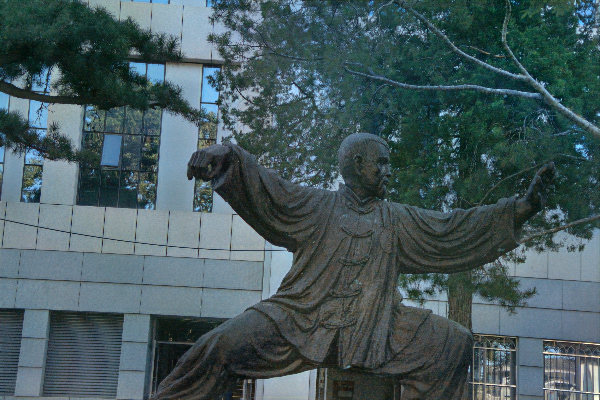}
        \caption{EnlightenGAN~\cite{jiang2021enlightengan}}
        \label{fig:lolv2real_736_engan}
    \end{subfigure}
    \hfill
    \begin{subfigure}[t]{0.19\textwidth}
        \centering
        \includegraphics[width=\textwidth]{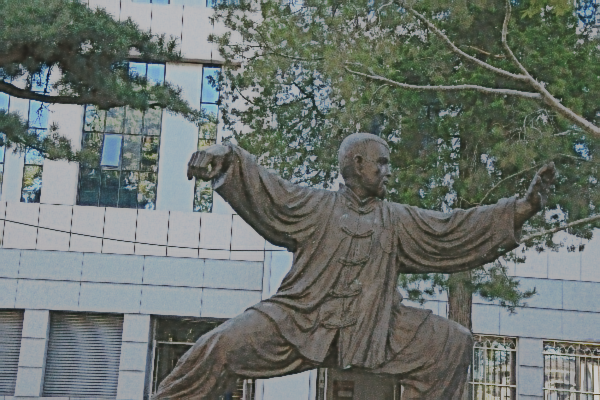}
        \caption{PairLIE~\cite{fu2023learning}}
        \label{fig:lolv2real_736_pairlie}
    \end{subfigure}
    \hfill
    \begin{subfigure}[t]{0.19\textwidth}
        \centering
        \includegraphics[width=\textwidth]{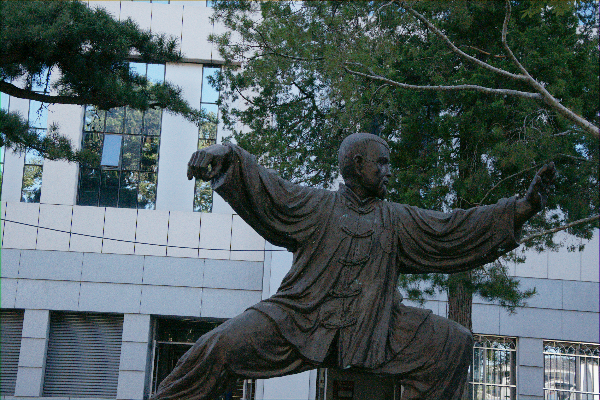}
        \caption{SCI~\cite{ma2022toward}}
        \label{fig:lolv2real_736_sci}
    \end{subfigure}
    
    \medskip 
    
    \begin{subfigure}[t]{0.19\textwidth}
        \centering
        \includegraphics[width=\textwidth]{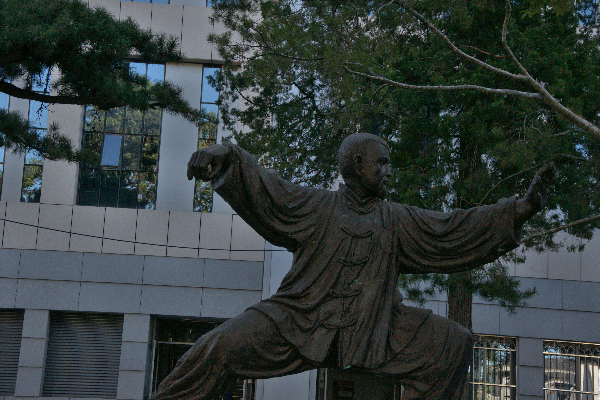}
        \caption{CoLIE~\cite{Chobola2024}}
        \label{fig:lolv2real_736_colie}
    \end{subfigure}
    \hfill
    \begin{subfigure}[t]{0.19\textwidth}
        \centering
        \includegraphics[width=\textwidth]{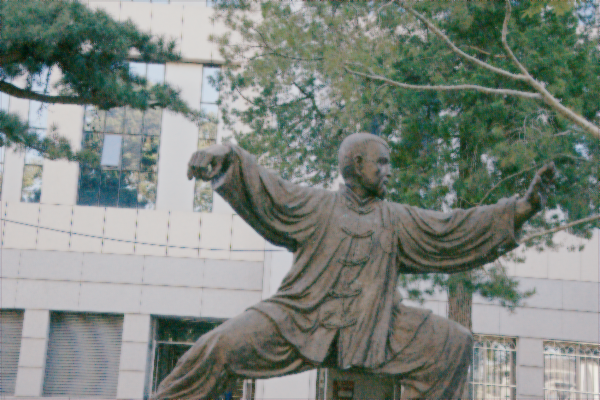}
        \caption{CLODE~\cite{jung2025continuous}}
        \label{fig:lolv2real_736_clode}
    \end{subfigure}
    \hfill
    \begin{subfigure}[t]{0.19\textwidth}
        \centering
        \includegraphics[width=\textwidth]{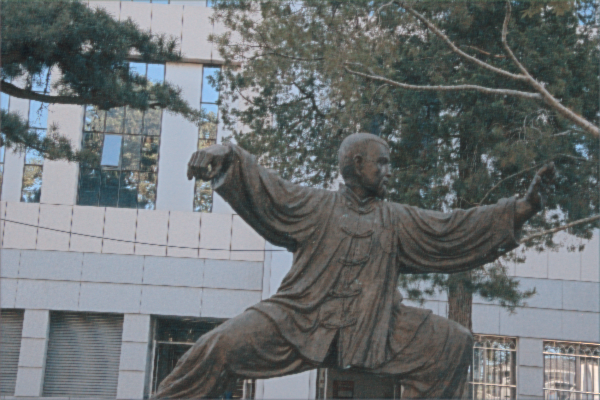}
        \caption{Li \textit{et al.}~\cite{li2025interpretable}}
        \label{fig:lolv2real_736_li}
    \end{subfigure}
    \hfill
    \begin{subfigure}[t]{0.19\textwidth}
        \centering
        \includegraphics[width=\textwidth]{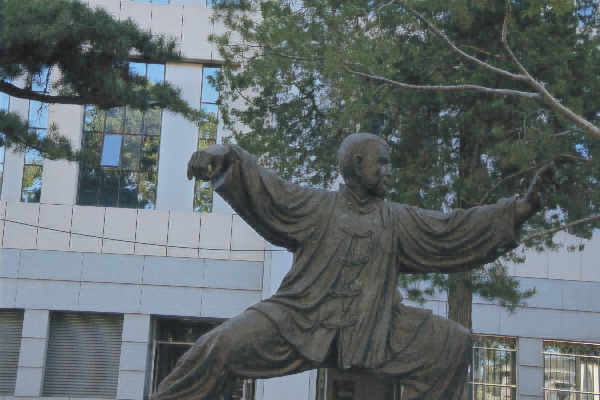}
        \caption{\textbf{Ours}}
        \label{fig:lolv2real_736_ours}
    \end{subfigure}
    \hfill
    \begin{subfigure}[t]{0.19\textwidth}
        \centering
        \includegraphics[width=\textwidth]{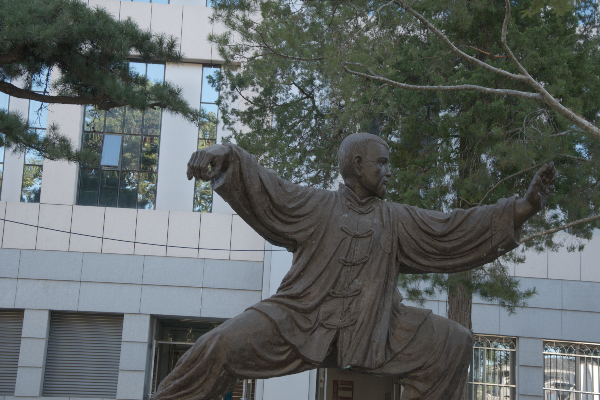}
        \caption{GT}
        \label{fig:lolv2real_736_gt}
    \end{subfigure}

    \Description{A visual comparison on the LOLv2-Real dataset. Competitor methods exhibit visible noise or color shifts. In contrast, 'Ours' displays clear textures and accurate colors, closely matching the Ground Truth.}
    \caption{\textbf{Visual comparison on LOLv2-Real~\cite{yang2021sparse}.} Our method restores accurate global illumination and fine textures, demonstrating robust performance on diverse real-world scenes.}
    \label{fig:supp_lolv2real_736}
\end{figure*}

\begin{figure*}[htbp]
    \centering
    \begin{subfigure}[t]{0.19\textwidth}
        \centering
        \includegraphics[width=\textwidth]{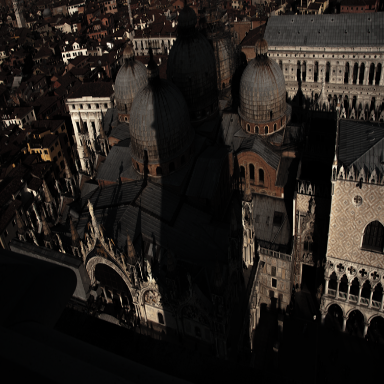}
        \caption{Input}
        \label{fig:lolv2syn_r108_input}
    \end{subfigure}
    \hfill
    \begin{subfigure}[t]{0.19\textwidth}
        \centering
        \includegraphics[width=\textwidth]{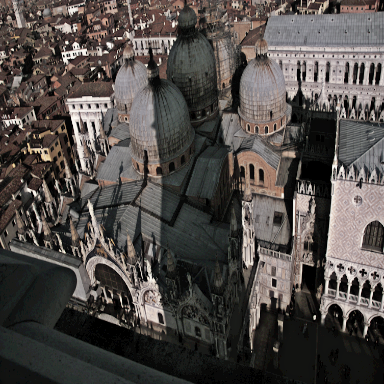}
        \caption{Zero-DCE~\cite{guo2020zero}}
        \label{fig:lolv2syn_r108_zerodce}
    \end{subfigure}
    \hfill
    \begin{subfigure}[t]{0.19\textwidth}
        \centering
        \includegraphics[width=\textwidth]{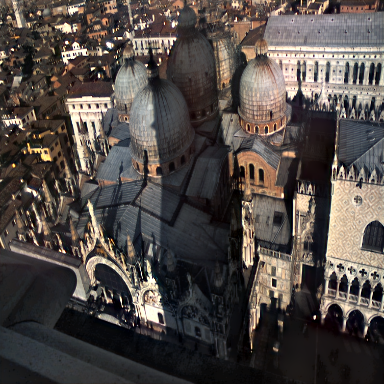}
        \caption{EnlightenGAN~\cite{jiang2021enlightengan}}
        \label{fig:lolv2syn_r108_engan}
    \end{subfigure}
    \hfill
    \begin{subfigure}[t]{0.19\textwidth}
        \centering
        \includegraphics[width=\textwidth]{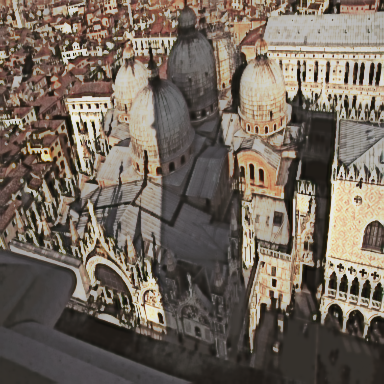}
        \caption{PairLIE~\cite{fu2023learning}}
        \label{fig:lolv2syn_r108_pairlie}
    \end{subfigure}
    \hfill
    \begin{subfigure}[t]{0.19\textwidth}
        \centering
        \includegraphics[width=\textwidth]{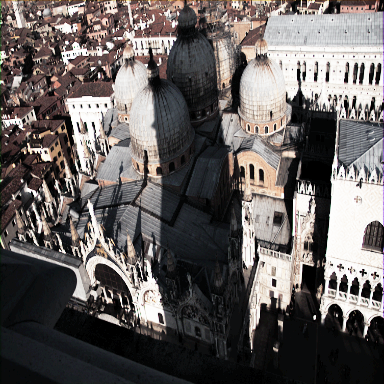}
        \caption{SCI~\cite{ma2022toward}}
        \label{fig:lolv2syn_r108_sci}
    \end{subfigure}
    
    \medskip 
    
    \begin{subfigure}[t]{0.19\textwidth}
        \centering
        \includegraphics[width=\textwidth]{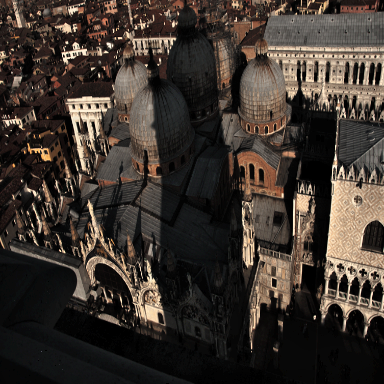}
        \caption{CoLIE~\cite{Chobola2024}}
        \label{fig:lolv2syn_r108_colie}
    \end{subfigure}
    \hfill
    \begin{subfigure}[t]{0.19\textwidth}
        \centering
        \includegraphics[width=\textwidth]{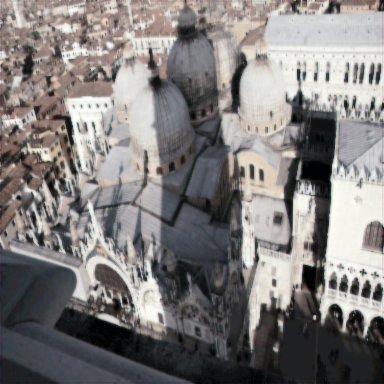}
        \caption{CLODE~\cite{jung2025continuous}}
        \label{fig:lolv2syn_r108_clode}
    \end{subfigure}
    \hfill
    \begin{subfigure}[t]{0.19\textwidth}
        \centering
        \includegraphics[width=\textwidth]{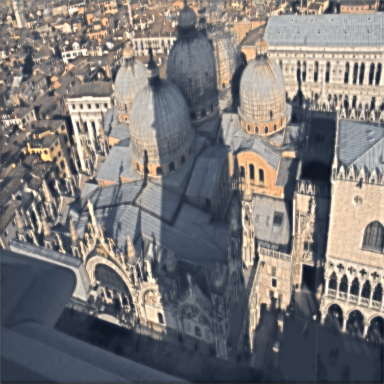}
        \caption{Li \textit{et al.}~\cite{li2025interpretable}}
        \label{fig:lolv2syn_r108_li}
    \end{subfigure}
    \hfill
    \begin{subfigure}[t]{0.19\textwidth}
        \centering
        \includegraphics[width=\textwidth]{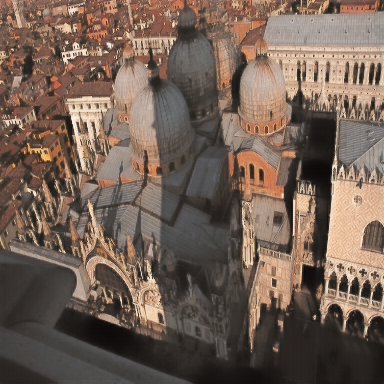}
        \caption{\textbf{Ours}}
        \label{fig:lolv2syn_r108_ours}
    \end{subfigure}
    \hfill
    \begin{subfigure}[t]{0.19\textwidth}
        \centering
        \includegraphics[width=\textwidth]{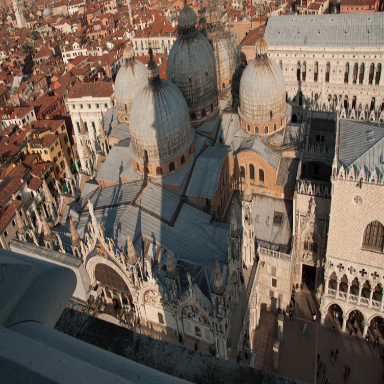}
        \caption{GT}
        \label{fig:lolv2syn_r108_gt}
    \end{subfigure}

    \Description{A visual comparison on the LOLv2-Synthetic dataset. Competitor methods exhibit visible noise or color shifts. In contrast, 'Ours' displays clear textures and accurate colors, closely matching the Ground Truth.}
    \caption{\textbf{Visual comparison on LOLv2-Synthetic~\cite{yang2021sparse}.} Our internally referenced enhancement effectively suppresses spatially-variant synthetic noise without compromising high-frequency edge details.}
    \label{fig:supp_lolv2syn_r108}
\end{figure*}

\begin{figure*}[htbp]
    \centering
    \begin{subfigure}[t]{0.19\textwidth}
        \centering
        \includegraphics[width=\textwidth]{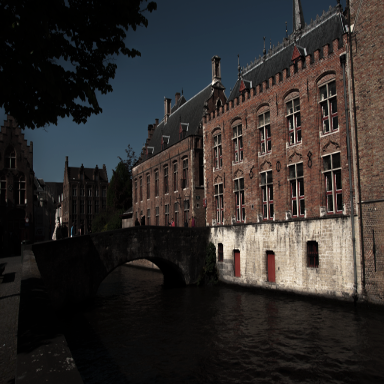}
        \caption{Input}
        \label{fig:lolv2syn_r130_input}
    \end{subfigure}
    \hfill
    \begin{subfigure}[t]{0.19\textwidth}
        \centering
        \includegraphics[width=\textwidth]{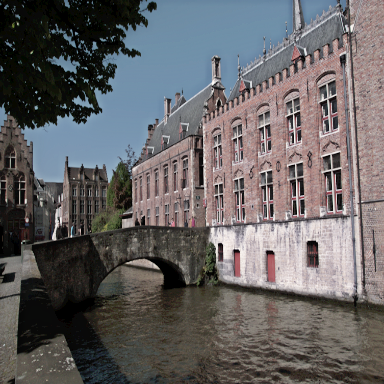}
        \caption{Zero-DCE~\cite{guo2020zero}}
        \label{fig:lolv2syn_r130_zerodce}
    \end{subfigure}
    \hfill
    \begin{subfigure}[t]{0.19\textwidth}
        \centering
        \includegraphics[width=\textwidth]{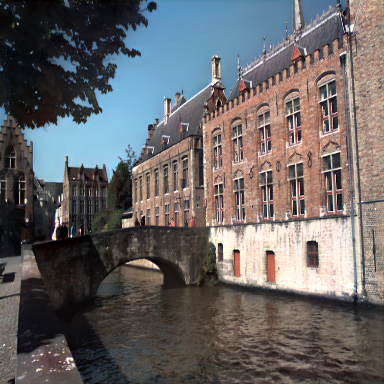}
        \caption{EnlightenGAN~\cite{jiang2021enlightengan}}
        \label{fig:lolv2syn_r130_engan}
    \end{subfigure}
    \hfill
    \begin{subfigure}[t]{0.19\textwidth}
        \centering
        \includegraphics[width=\textwidth]{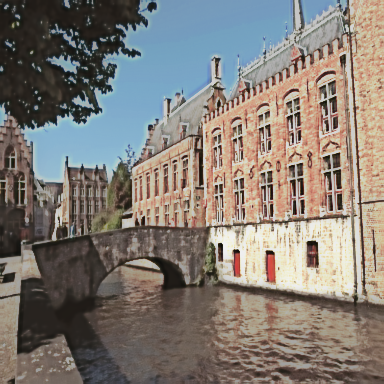}
        \caption{PairLIE~\cite{fu2023learning}}
        \label{fig:lolv2syn_r130_pairlie}
    \end{subfigure}
    \hfill
    \begin{subfigure}[t]{0.19\textwidth}
        \centering
        \includegraphics[width=\textwidth]{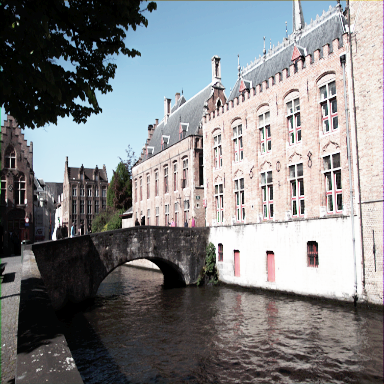}
        \caption{SCI~\cite{ma2022toward}}
        \label{fig:lolv2syn_r130_sci}
    \end{subfigure}
    
    \medskip 

    \begin{subfigure}[t]{0.19\textwidth}
        \centering
        \includegraphics[width=\textwidth]{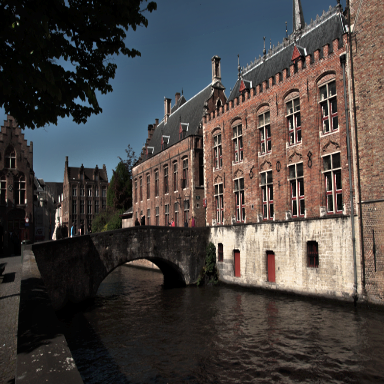}
        \caption{CoLIE~\cite{Chobola2024}}
        \label{fig:lolv2syn_r130_colie}
    \end{subfigure}
    \hfill
    \begin{subfigure}[t]{0.19\textwidth}
        \centering
        \includegraphics[width=\textwidth]{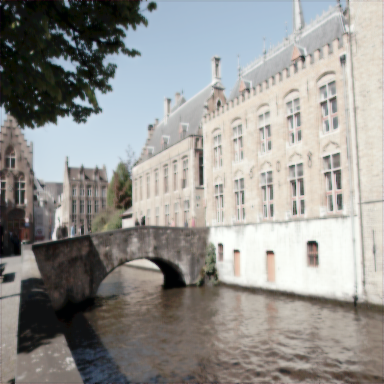}
        \caption{CLODE~\cite{jung2025continuous}}
        \label{fig:lolv2syn_r130_clode}
    \end{subfigure}
    \hfill
    \begin{subfigure}[t]{0.19\textwidth}
        \centering
        \includegraphics[width=\textwidth]{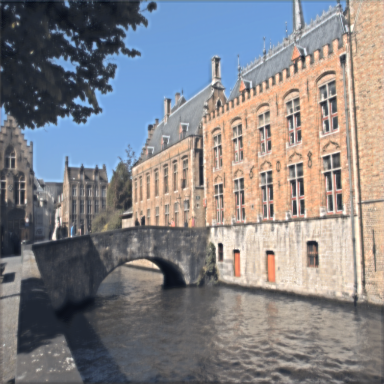}
        \caption{Li \textit{et al.}~\cite{li2025interpretable}}
        \label{fig:lolv2syn_r130_li}
    \end{subfigure}
    \hfill
    \begin{subfigure}[t]{0.19\textwidth}
        \centering
        \includegraphics[width=\textwidth]{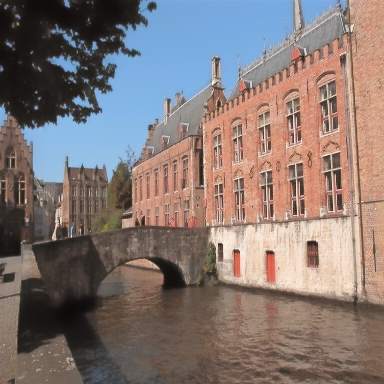}
        \caption{\textbf{Ours}}
        \label{fig:lolv2syn_r130_ours}
    \end{subfigure}
    \hfill
    \begin{subfigure}[t]{0.19\textwidth}
        \centering
        \includegraphics[width=\textwidth]{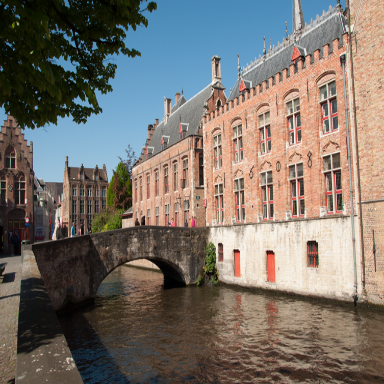}
        \caption{GT}
        \label{fig:lolv2syn_r130_gt}
    \end{subfigure}

    \Description{A visual comparison on the LOLv2-Synthetic dataset. Competitor methods exhibit visible noise or color shifts. In contrast, 'Ours' displays clear textures and accurate colors, closely matching the Ground Truth.}
    \caption{\textbf{Visual comparison on LOLv2-Synthetic~\cite{yang2021sparse}.} Our internally referenced enhancement effectively suppresses spatially-variant synthetic noise without compromising high-frequency edge details.}
    \label{fig:supp_lolv2syn_r130}
\end{figure*}

In this section, we provide extensive visual comparison results across the LOLv1, LOLv2-Real, and LOLv2-Synthetic datasets to further demonstrate the qualitative superiority of our proposed IRLE framework. As shown from Fig.~\ref{fig:supp_lolv1_22} to Fig.~\ref{fig:supp_lolv2syn_r130}, existing unsupervised and zero-reference methods frequently struggle with the highly ill-posed nature of low-light enhancement, often resulting in severe color casts, amplified sensory noise, or the over-smoothing of underlying textures.

\textbf{Performance on LOLv1.} In typical indoor low-light scenarios (e.g., the indoor desk in Fig.~\ref{fig:supp_lolv1_22} and the bowling alley in Fig.~\ref{fig:supp_lolv1_669}), the primary challenges are recovering uniform illumination and maintaining color constancy. Methods like Zero-DCE and PairLIE tend to under-enhance the severely dark regions, leaving the images visibly dim. Conversely, methods such as EnlightenGAN and CLODE aggressively boost brightness but introduce unnatural color shifts (e.g., noticeable yellowish or greenish tints on the wooden floor and walls). Thanks to our local exposure-simulated pseudo-GT, IRLE restores more natural global brightness and more faithfully preserves the color distribution of the scenes.

\textbf{Performance on LOLv2-Real.} Real-world outdoor scenes present complex overlapping structures and textures (e.g., the structural lines in Fig.~\ref{fig:supp_lolv2real_721} and the statue with background foliage in Fig.~\ref{fig:supp_lolv2real_736}). In these cases, baselines that rely on traditional spatial-domain consistency or uniform denoising (such as CoLIE and Li \textit{et al.}) exhibit a severe trade-off: they either leave blotchy residual noise or aggressively blur the high-frequency details. Our IRLE framework, guided by the shift-invariant SISC loss and the Gain-Adaptive Feature Modulation (GAFM) mechanism, helps distinguish structurally coherent edges from random noise. As a result, it better reconstructs complex details while keeping the background relatively clean.

\textbf{Performance on LOLv2-Synthetic.} The synthetic dataset introduces highly spatially-variant and heavy artificial noise patterns, which are particularly challenging when coupled with dense scene elements (e.g., the building miniature in Fig.~\ref{fig:supp_lolv2syn_r108} and the canal bridge in Fig.~\ref{fig:supp_lolv2syn_r130}). Competitors like SCI and RUAS fail to suppress this heavy noise, non-linearly amplifying it into severe visual artifacts. Meanwhile, other generative methods tend to wash out the sky or distort the contrast. IRLE effectively rejects these unshared random noise patterns in the spectral domain without severely compromising the underlying structural edges of the scenes, delivering a more balanced and visually pleasing restoration.

\begin{figure*}[htbp]
    \centering
    \begin{subfigure}[t]{0.19\textwidth}
        \centering
        \includegraphics[width=\textwidth]{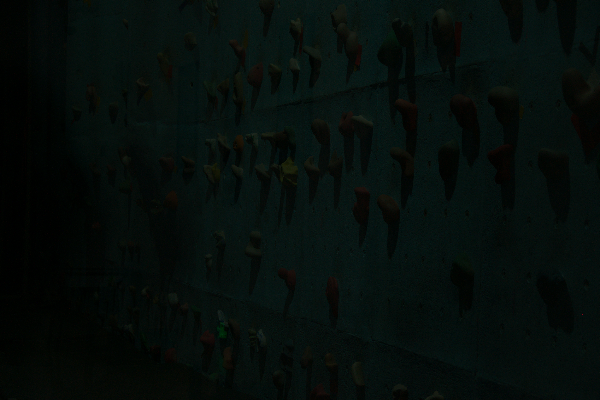}
        \caption{Input}
        \label{fig:lolv2real_754_input}
    \end{subfigure}
    \hfill
    \begin{subfigure}[t]{0.19\textwidth}
        \centering
        \includegraphics[width=\textwidth]{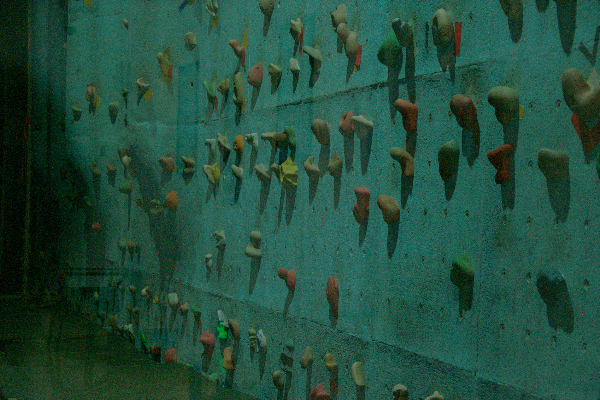}
        \caption{Zero-DCE~\cite{guo2020zero}}
        \label{fig:lolv2real_754_zerodce}
    \end{subfigure}
    \hfill
    \begin{subfigure}[t]{0.19\textwidth}
        \centering
        \includegraphics[width=\textwidth]{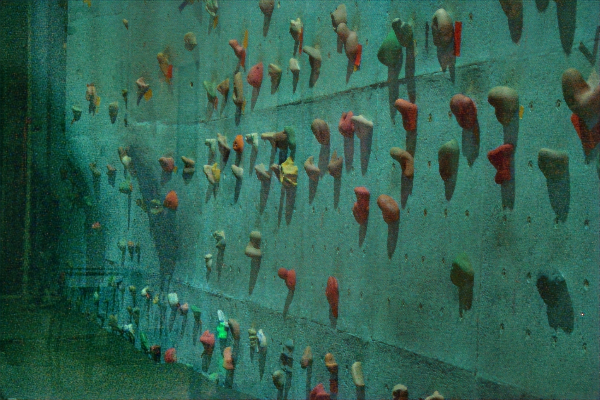}
        \caption{EnlightenGAN~\cite{jiang2021enlightengan}}
        \label{fig:lolv2real_754_engan}
    \end{subfigure}
    \hfill
    \begin{subfigure}[t]{0.19\textwidth}
        \centering
        \includegraphics[width=\textwidth]{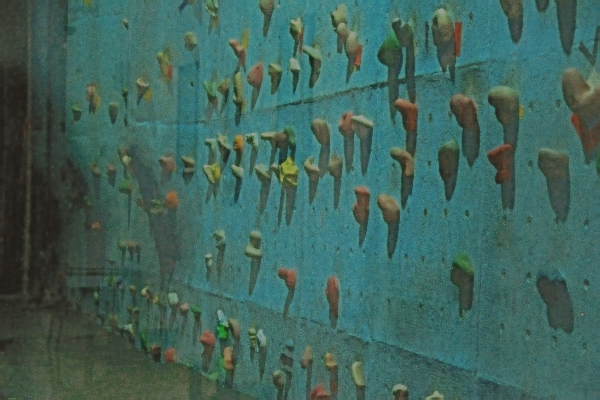}
        \caption{PairLIE~\cite{fu2023learning}}
        \label{fig:lolv2real_754_pairlie}
    \end{subfigure}
    \hfill
    \begin{subfigure}[t]{0.19\textwidth}
        \centering
        \includegraphics[width=\textwidth]{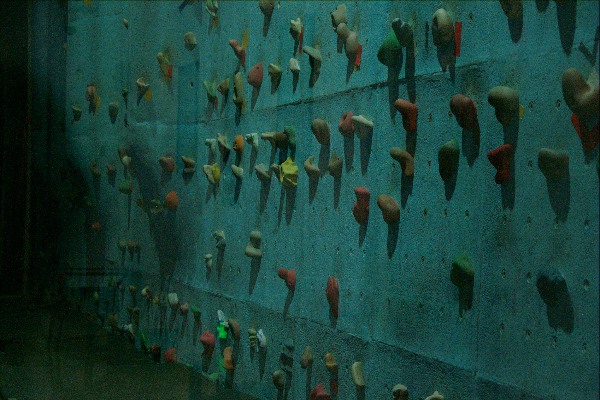}
        \caption{SCI~\cite{ma2022toward}}
        \label{fig:lolv2real_754_sci}
    \end{subfigure}
    
    \medskip 
    
    \begin{subfigure}[t]{0.19\textwidth}
        \centering
        \includegraphics[width=\textwidth]{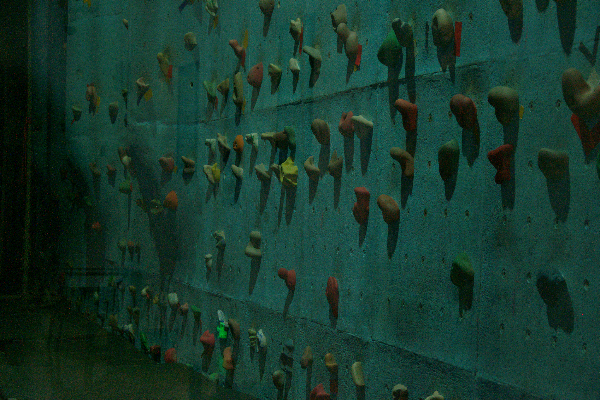}
        \caption{CoLIE~\cite{Chobola2024}}
        \label{fig:lolv2real_754_colie}
    \end{subfigure}
    \hfill
    \begin{subfigure}[t]{0.19\textwidth}
        \centering
        \includegraphics[width=\textwidth]{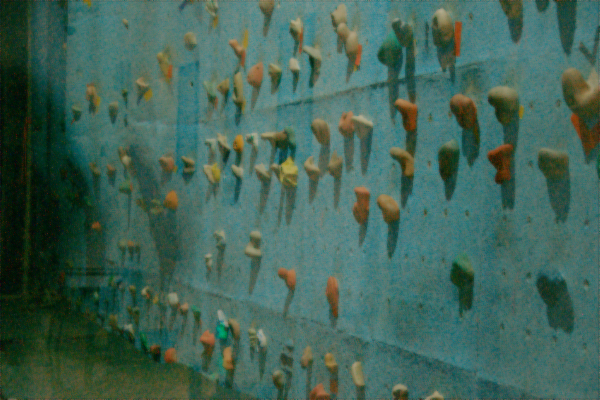}
        \caption{CLODE~\cite{jung2025continuous}}
        \label{fig:lolv2real_754_clode}
    \end{subfigure}
    \hfill
    \begin{subfigure}[t]{0.19\textwidth}
        \centering
        \includegraphics[width=\textwidth]{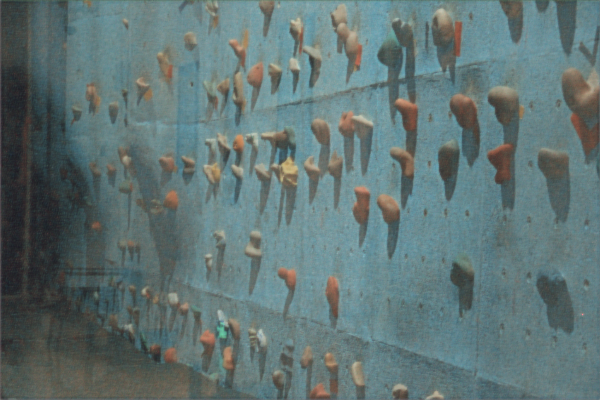}
        \caption{Li \textit{et al.}~\cite{li2025interpretable}}
        \label{fig:lolv2real_754_li}
    \end{subfigure}
    \hfill
    \begin{subfigure}[t]{0.19\textwidth}
        \centering
        \includegraphics[width=\textwidth]{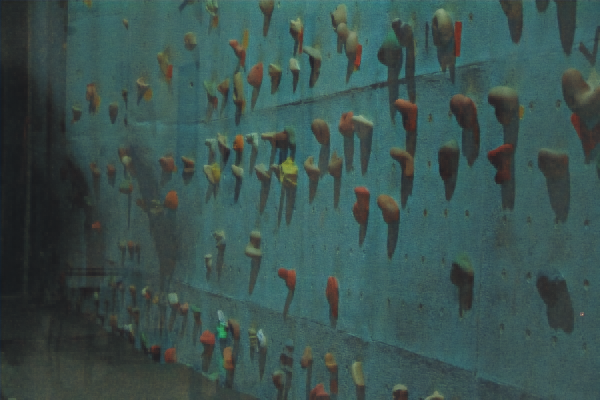}
        \caption{\textbf{Ours}}
        \label{fig:lolv2real_754_ours}
    \end{subfigure}
    \hfill
    \begin{subfigure}[t]{0.19\textwidth}
        \centering
        \includegraphics[width=\textwidth]{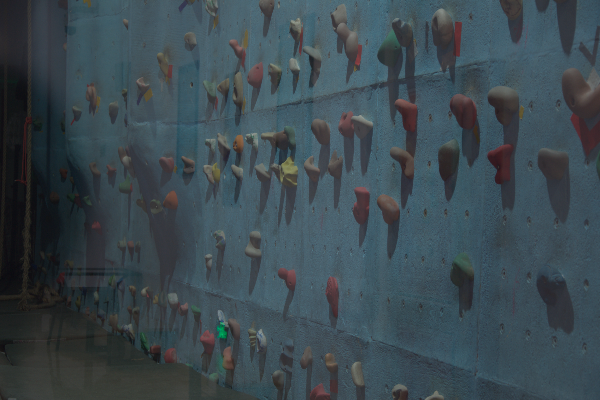}
        \caption{GT}
        \label{fig:lolv2real_754_gt}
    \end{subfigure}

    \Description{A visual comparison on a challenging case with severe color cast. While our method cannot perfectly restore the ground-truth colors from the highly degraded input, it still significantly outperforms competitors by suffering the least from unnatural color shifts.}
    \caption{\textbf{Visual comparison on a challenging case with severe color cast (LOLv2-Real~\cite{yang2021sparse}).} When the original input exhibits extreme color degradation, our internally referenced method can only partially mitigate the color cast. However, compared to competing baselines that severely distort the scene's hue, our method still yields the most natural visual results.}
    \label{fig:supp_lolv2real_754}
\end{figure*}

\section{Limitations and Future Work}
\label{sec:supp_limitations}

While our proposed Internally Referenced Low-Light Enhancement (IRLE) framework demonstrates state-of-the-art performance in decoupling illumination, textures, and noise without external targets, it still has a few limitations that warrant further investigation.

First, recovering accurate chromaticity in near-pitch-black regions and completely rectifying extreme inherent color casts remain fundamental physical challenges. Because our framework explicitly relies on extracting internal physical and structural references directly from the degraded input, it fundamentally requires a minimal level of underlying signal. When the signal-to-noise ratio (SNR) approaches zero in severely dark areas, or when the input image suffers from an extreme intrinsic color shift, the authentic color information is highly corrupted. Consequently, without external dataset priors to hallucinate the missing content, our internally referenced paradigm cannot generate colors out of nothing or perfectly rectify the initial bias. However, as illustrated in Fig.~\ref{fig:supp_lolv2real_754}, even in such extreme cases, our method can still partially mitigate the degradation and yields a significantly weaker color cast compared to existing baselines.

Second, the current framework is exclusively designed and optimized for single-image low-light enhancement. When applied directly to video sequences on a frame-by-frame basis, the independent generation of the local exposure-simulated pseudo-GT and the dynamic spatial gain prior for each frame may introduce slight temporal flickering. 

In our future work, we plan to address these limitations from two perspectives. To tackle the extreme color cast and chromaticity degradation, we will explore the integration of lightweight, zero-shot color constancy priors or language-driven color guidance. This could provide reliable external semantic color hints without sacrificing the generalizability of our unsupervised framework. Furthermore, to adapt our framework for robust and flicker-free low-light video enhancement, we aim to extend the internally referenced paradigm into the temporal domain by explicitly exploiting cross-frame structural dependencies and temporal consistency constraints.

\end{document}